%% file: main.tex
\documentclass{article} 
\usepackage{iclr2025_conference}
\usepackage{times}

\input{math_commands.tex}

\usepackage{hyperref}
\usepackage{url}

\usepackage{xcolor}
\usepackage{graphicx}
\usepackage[capitalize,noabbrev]{cleveref}
\usepackage{caption}
\usepackage{subcaption}
\usepackage{wrapfig}

\usepackage{algorithm}
\usepackage{algpseudocode}
\usepackage{amsmath}
\usepackage{booktabs}
\usepackage{multirow}
\usepackage{amsthm} 
\usepackage{soul}

\title{QERA: an Analytical Framework for Quantization Error Reconstruction}

\author{Cheng Zhang,  Jeffrey T. H. Wong, Can Xiao, George A. Constantinides \& Yiren Zhao \\
Department of Electrical and Electronic Engineering\\
Imperial College London\\
London, UK\\
\texttt{\{cheng.zhang122,tsz.wong20,can.xiao22,g.constantinides,a.zhao\}@imperial.ac.uk} \\
}

\newcommand{\LQER}{LQER}
\newcommand{\loqer}{QERA}
\newcommand{\loqerdiag}{QERA-approx}
\newcommand{\loqerrxx}{QERA-exact}

\newcommand{\roberta}{RoBERTa}
\newcommand{\llama}{LLaMA}
\newcommand{\loftq}{LoftQ}
\newcommand{\qlora}{QLoRA}

\newcommand{\lqlora}{LQ-LoRA}

\newcommand{\qllm}{QLLM}

\newcommand{\zeroquantii}{ZeroQuant-V2}
\newcommand{\ie}{\textit{i.e.}}
\newcommand{\eg}{\textit{e.g.}}

\newcommand{\good}[1]{\textbf{#1}}

\definecolor{Green}{HTML}{00A64F}

\newcommand{\isiclr}{iclr}

\newcommand{\caldera}{CALDERA}

\theoremstyle{plain}
\newtheorem{theorem}{Theorem}

\newtheorem{remark}{Remark}
\theoremstyle{definition}
\newtheorem{assumption}{Assumption}

\newtheorem{problem}{Problem} 

\iclrfinalcopy 
\begin{document}

\maketitle

\begin{abstract}
    The growing number of parameters and computational demands of large language models (LLMs) present significant challenges for their efficient deployment.
    Recently, there is an increasing interest in quantizing weights to extremely low precision while offsetting the resulting error with low-rank, high-precision error reconstruction terms.
    The combination of quantization and low-rank approximation is now popular in both adapter-based, parameter-efficient fine-tuning methods such as \loftq{}~\citep{li2023loftq} and low-precision inference techniques including \zeroquantii{}~\citep{yao2023zeroquant}.
    Usually, the low-rank terms are calculated via the singular value decomposition (SVD) of the weight quantization error,
    minimizing the Frobenius and spectral norms of the weight approximation error.
    Recent methods like \lqlora{}~\citep{guo2023lqlora} and \LQER{}~\citep{zhang2024lqer} introduced hand-crafted heuristics to minimize errors in layer outputs (activations) rather than weights, resulting improved quantization results.
    However, these heuristic-based methods lack an analytical solution to guide the design of quantization error reconstruction terms.
    In this paper, we revisit this problem and formulate an analytical framework, named Quantization Error Reconstruction Analysis (\loqer{}),
    and offer a closed-form solution to the problem.
    We show \loqer{} benefits both existing low-precision fine-tuning and inference methods --
    \loqer{} achieves a fine-tuned accuracy gain for $\Delta_{\text{acc}}$ = 6.05\% of 2-bit \roberta{}-base on GLUE compared to \loftq{};
    and obtains $\Delta_{\text{acc}}$ = 2.97\% higher post-training quantization accuracy of 4-bit Llama-3.1-70B compared to \zeroquantii{} and $\Delta_{\text{ppl}}$ = $-$ 0.28 lower perplexity on WikiText2 compared to \LQER{}.
    We open-source our code and models at \href{https://github.com/ChengZhang-98/QERA}{github.com/ChengZhang-98/QERA}.

\end{abstract}

\input{sections/1_introduction}
\input{sections/2_related_work}
\input{sections/3_method}

\input{sections/4_experiments}
\input{sections/5_discussion}

\input{sections/6_conclusion.tex}
\input{sections/7_acknowlegement}

\bibliography{ref}
\bibliographystyle{iclr2025_conference}

\newpage
\appendix
\input{sections/99_appendix}

\end{document}

%% file: math_commands.tex
\usepackage{amsmath,amsfonts,bm}

\def\eqref#1{equation~\ref{#1}}

\def\1{\bm{1}}

\def\ve{{\bm{e}}}

\def\vp{{\bm{p}}}

\def\vs{{\bm{s}}}

\def\vx{{\bm{x}}}
\def\vy{{\bm{y}}}

\def\mA{{\bm{A}}}
\def\mB{{\bm{B}}}
\def\mC{{\bm{C}}}
\def\mD{{\bm{D}}}

\def\mP{{\bm{P}}}
\def\mQ{{\bm{Q}}}
\def\mR{{\bm{R}}}
\def\mS{{\bm{S}}}

\def\mU{{\bm{U}}}
\def\mV{{\bm{V}}}
\def\mW{{\bm{W}}}
\def\mX{{\bm{X}}}
\def\mY{{\bm{Y}}}

\def\mSigma{{\bm{\Sigma}}}

\DeclareMathAlphabet{\mathsfit}{\encodingdefault}{\sfdefault}{m}{sl}
\SetMathAlphabet{\mathsfit}{bold}{\encodingdefault}{\sfdefault}{bx}{n}

\def\sR{{\mathbb{R}}}

\def\sX{{\mathbb{X}}}
\def\sY{{\mathbb{Y}}}

\newcommand{\E}{\mathbb{E}}

\newcommand{\R}{\mathbb{R}}

\DeclareMathOperator*{\argmin}{arg\,min}

\DeclareMathOperator{\Tr}{Tr}

%% file: sections/1_introduction.tex
\vspace{-1em}
\section{Introduction}
\label{sec:introduction}
\ifdefined\isiclr
      \vspace{-0.5em}
\fi

The demand for efficient deployment of large language models (LLMs) has been increasing~\citep{faiz2023llmcarbon}.
LLMs now typically
contain billions of parameters~\citep{kaplan2020scaling,dubey2024llama}, making their fine-tuning and inference computationally expensive and
resource-intensive~\citep{ding2023parameter}.
To address these challenges, there has been a surge of interest in building efficient fine-tuning and inference methods.
One popular formulation is to apply a low-rank term to reconstruct the error after quantization.
Given a linear layer $\vy=\vx\mW$,
the weight matrix $\mW \in \sR^{m\times n}$ is quantized to $\widetilde{\mW}$,
and we rewrite $\mW \approx \widetilde{\mW} + \mA_k \mB_k$ such that both $\mA_k$ and $\mB_k$ are low-rank yet high-precision terms with rank $k \ll \min(m, n)$.

We call the problem of finding the optimal $\mA_k$ and $\mB_k$ \textit{quantization error reconstruction}. Interestingly, this problem has, coincidentally, seen widespread application in two actively researched areas: quantized parameter-efficient fine-tuning (QPEFT) and post-training quantization (PTQ) for model inference.
QPEFT refers to fine-tuning techniques that adapt LLMs to specific tasks by quantizing pretrained weights and updating only a small number of extra parameters, hence significantly reducing memory requirements and training time, such as \qlora{}~\citep{guo2023lqlora}.
On the other side, PTQ is a training-free method that reduces the model size and may accelerate the forward pass if the underlying hardware supports it.
Recently, researchers combined PTQ with quantization error reconstruction~\citep{yao2023zeroquant,liu2023qllm,zhang2024lqer} to further reduce weight precision.
Works such as ZeroQuant-V2~\citep{yao2023zeroquant} and \LQER{}~\citep{zhang2024lqer} have shown that adding a high-precision low-rank component, as low as 8 or 32, can recover considerable model performance for 3- or 4-bit weight quantization.

Although both the QPEFT and PTQ methods have demonstrated substantial performance improvements in lowering the computational overhead of LLMs,
a theoretical analysis of quantization error reconstruction is lacking.
Usually, $\mA_k$ and $\mB_k$ are calculated by applying truncated singular value decomposition (SVD) to the weight quantization error $(\mW - \widetilde{\mW})$,
minimizing the Frobenius and spectral norms of the \textit{weight approximation error}.
However, recent work on activation-aware quantization and knowledge distillation implies that minimizing \textit{layer output error} may
lead to a greater performance gain than minimizing weight approximation error~\citep{lin2024awq,liu2023qllm,shao2023omniquant}.

Besides the unsettled minimization objective, it has remained unclear whether there exists a theoretically optimal solution for the values of $\mA_k$ and $\mB_k$, and if so, how one can solve for it.
A better initialization or theoretically grounded initialization of $\mA_k$ and $\mB_k$ brings direct benefits for both QPEFT and PTQ.
In QPEFT, the initialization of LoRA~\citep{hu2021lora}, which uses element-wise Gaussian random values for $\mA_k$ and zeros for $\mB_k$, struggles under aggressive quantization since the quantization error can derail fine-tuning.
In PTQ, the quantized model performance is based on the computation of the low-rank terms, given a specific quantization function $\mathrm{q}(\cdot)$ and rank $k$.

In this paper, we aim to provide an analytical framework for the \textit{quantization error reconstruction} problem.
To demonstrate the effectiveness of our theoretical framework, we further apply our analytical solutions to state-of-the-art QPEFT and PTQ methods and
show the significant performance improvements under the same computational budget. Specifically, our contributions are as follows:

\ifdefined\isiclr
      \vspace{-0.5em}
\fi
\begin{itemize}
      \item We show that the commonly used objective for solving the \textit{quantization error reconstruction} problem in prior work \cite{}, \ie{},
            minimizing the weight approximation error (\eg, $||\mW - \widetilde{\mW}||_p$),
            does not guarantee a reduced \textit{model output error}.
            Instead, we show that minimizing the layer output error (\eg, $||\vy - \widetilde{\vy}||_p$) is closely related to minimizing the model output error.
      \item We derive the analytical solution to the low-rank terms $\mA_k$ and $\mB_k$ by minimizing the layer output error.
            We demonstrate that under a statistical assumption, this solution can be found in a particularly computationally efficient manner, also explaining the success of \LQER{}.
      \item We empirically demonstrate the effectiveness of our solutions by applying them to state-of-the-art QPEFT and PTQ methods.
            Our analytical framework, \loqer{}, significantly improves the performance of these methods.
            For example, \loqer{} achieves $\Delta_{\text{acc}}$ = 6.05\% higher accuracy of 2-bit \roberta{}-base on GLUE compared to \loftq{}, improving the fine-tuning accuracy and efficiency. Moreover, \loqer{} obtains $\Delta_{\text{acc}}$ = 2.97\% higher accuracy than \zeroquantii{}, when quantizing \llama{}-3-70B to 4 bits, averaged across six tasks. This narrows the model performance gap between error-reconstruction-based post-training quantization and full-precision models.
\end{itemize}

%% file: sections/2_related_work.tex
\section{Related Work}
\label{sec:related_work}
\ifdefined\isiclr
    \vspace{-0.5em}
\fi

In this section, we review the existing methods that combine weight quantization and low-rank error reconstruction.
These methods can be roughly categorized into two groups based on their applications:
QPEFT for training and PTQ for inference.

\ifdefined\isiclr
    \vspace{-0.5em}
\fi
\paragraph{LoRA and QPEFT} LoRA~\cite{hu2021lora} is a representative PEFT method that
introduces trainable low-rank terms to adapt the model to a specific task.
Take a linear layer as an example,
\begin{equation}
    \vy = \vx (\mW + \mA_k \mB_k)
\end{equation}
where $\mW\in \sR^{m\times n}$ is the pretrained weight matrix,
row vector $\vx \in \sR^{m}$ and $\vy \in \sR^{n}$ are the input and output,
and $\mA_k\in \sR^{m\times k}$ and $\mB_k\in \sR^{k \times n}$ are trainable low-rank matrices (``adapter'') with rank $k \ll \min(m, n)$.
During fine-tuning, the pretrained $\mW$ is frozen and only the adapter $\mA_k$ and $\mB_k$ are updated.
To make the adapted layer's output match the original one at the start of fine-tuning,
LoRA initializes $\mA_k$ with Gaussian random values and $\mB_k$ with zeros. Once the fine-tuning is completed, the adapter
is merged into the pre-trained weights.

\qlora{}~\citep{guo2023lqlora} extends LoRA by quantizing the pretrained weights stored in GPU memory to reduce memory footprint.
\begin{equation}
    \mW_q = \mathrm{q}(\mW)
\end{equation}
One difference between \qlora{} and LoRA is that during fine-tuning, $\mW_q$ needs to be dequantized before involved into matrix multiplications:
\begin{equation}
    \begin{split}
        \widetilde{\mW} = \mathrm{dq}(\mW_q),\;
        \vy = \vx (\widetilde{\mW} + \mA_k \mB_k)
    \end{split}
\end{equation}
where $\mathrm{dq}(\cdot)$ is the dequantization function.
\qlora{} introduces weight quantization error $(\mW - \widetilde{\mW})$,
shifting the starting point of fine-tuning.
To address this problem, \loftq{}~\citep{li2023loftq} initializes the adapter
using the SVD-based low-rank approximation of $(\mW - \widetilde{\mW})$ to reduce the weight approximation error:
\begin{equation}
    \argmin_{\mA_k, \mB_k} ||\mW - \widetilde{\mW} - \mA_k \mB_k||_F
\end{equation}
Specifically, LoftQ uses a heuristic-based algorithm to iteratively update the quantized weights and the adapter (\Cref{alg:loftq} in the Appendix).
Their experiments show that a larger number of iterations leads to a smaller weight error.

\lqlora{}~\citep{guo2023lqlora} also adopts \loftq{}'s iterative method but keeps track of a scaled variant of the objective,
$\argmin_{\mA_k, \mB_k} ||\mD_{\text{row}}(\mW -\widetilde{\mW} - \mA_k \mB_k)\mD_{\text{col}}||_F$,
where $\mD_{\text{row}}$ and $\mD_{\text{col}}$ are heuristic homogenous row/column matrices from activation statistics.
\lqlora{} exits the iteration when the scaled objective function stops decreasing
due to the lack of a theoretical justification for~\loftq{}.

\ifdefined\isiclr
    \vspace{-0.5em}
\fi
\paragraph{Quantization Error Reconstruction for PTQ}

Similar to the forward pass of fine-tuning in \qlora{}, there are also PTQ methods that quantize the pretrained weights
to low-precision formats and recover the model performance with additional low-rank terms.
With a small enough rank $k$, the additional computation introduced is negligible.
Note that unlike QPEFT which can utilize fine-tuning to correct the quantization error,
PTQ methods aim to recover the model performance as much as possible without any training.

\zeroquantii{}~\citep{yao2023zeroquant} is the earliest weight-only quantization method introducing low-rank quantization error reconstruction to the PTQ problem.
They apply SVD to the weight quantization error $(\mW-\widetilde{\mW})$ to calculate $\mA_k$ and $\mB_k$
(equivalent to \loftq{} with one iteration). Combining low-rank terms and fine-grained quantization,
\zeroquantii{} recovers the performance of 4-bit LLMs to a level comparable to 8-bit.

Recent quantization works have shown that activation statistics play a crucial role in weight-only LLM quantization~\citep{liu2023llmqat,lin2024awq}.
\qllm{}~\citep{liu2023qllm} trains the low-rank terms using gradient descent with a loss function that minimizes the output error of the attention layer.
\LQER{}~\citep{zhang2024lqer} applies an activation-induced heuristic scale matrix $\mS$ to the quantization error before calculating SVD,
$\mU\mSigma \mV^T = \mathrm{SVD}(\mS (\mW - \widetilde{\mW}))$, and assigns $\mA_k := \mS^{-1}\mU_{:,:k},\; \mB_k := \mSigma_{:k,:k} \mV^T_{:k,:}$ (Refer to \Cref{alg:lqer} in the Appendix).
\LQER{} achieves significant improvement over \zeroquantii{} and
observes that in some layers singular values are shaped toward a more desirable distribution where singular values decay faster.
Note that \zeroquantii{} can also be considered as a special case where $\mS$ in \LQER{} is an identity matrix.
To our knowledge, \caldera{}~\citep{saha2024caldera} is the concurrent work close to ours.
\caldera{} focuses on a different problem setup to find optimal $\widetilde{W}$, $A_k$, $B_k$ all in low-precision formats that minimizes output error,
with a lemma agreeing with our exact solution.
We elaborate the connection and difference between \caldera{} and QERA in~\Cref{sec:appendix:caldera_vs_qera}.

In summary, to solve the quantization error reconstruction (QER) problem,
most of existing methods target the minimization of the weight approximation error.
Several recent works such as \lqlora{}, \qllm{}, and \LQER{} introduce activation-induced heuristics
to the calculation of adapters/low-rank terms,
but \textbf{a justification for the optimization objective and the corresponding analytical framework are still missing}.

%% file: sections/3_method.tex
\ifdefined\isiclr
    \vspace{-0.5em}
\fi
\section{Our Analytical Framework}
\label{sec:method}
\ifdefined\isiclr
    \vspace{-0.5em}
\fi

In this section, we formulate the optimization objective of quantization error reconstruction
and derive the analytical solution to the low-rank term $\mC_k := \mA_k\mB_k$.

\subsection{Problem Statement}

Given a pretrained linear layer $\vy = \vx\mW$ with input vector $\vx\in\sR^m$,
output vector $\vy\in\sR^n$, and weight matrix $\mW\in\sR^{m\times n}$, our aim is to
approximate it with a high-rank low-precision $\widetilde{\mW}$
and a low-rank high-precision term $\mC_k \in \sR^{m\times n}$ with rank $k\ll \min(m,n)$.
\begin{equation}
    \widetilde{\vy} = \vx(\widetilde{\mW} + \mC_k)
\end{equation}

This raises the question of the actual optimization target:
Should we minimize the weight reconstruction error $||\mW - \widetilde{\mW}||_F$ or the output reconstruction error $||\vy - \widetilde{\vy}||_2$?
We separate these two problems and introduce them formally below.

\begin{problem}[Minimization of weight error] \label{problem:minimize-Eq}
For a pretrained linear layer $\vy=\vx\mW$ and its approximated form $\widetilde{\vy}=\vx(\widetilde{\mW}+\mC_k)$,
reconstructing the quantization error by minimizing weight approximation error has the following objective:
\begin{equation}
    \argmin_{\mC_k} ||\mW - \widetilde{\mW} - \mC_k||_F
\end{equation}
where $\|\cdot\|_F$ denotes the Frobenius norm.
\end{problem}

\textbf{Solution to Problem~\ref{problem:minimize-Eq}}. From the Eckart-Young-Mirsky theorem~\citep{eckart1936approximation},
the optimal solution to Problem~\ref{problem:minimize-Eq} with respect to rank $k$ is the truncated SVD of the weight error matrix:
\begin{equation}
    \mC_k = \mU_{:,:k} \mSigma_{:k,:k} \mV^T_{:k,:}
\end{equation}
where $\mU$, $\mSigma$, and $\mV^T$ form the SVD of the weight quantization error,
$\mU\mSigma\mV^T = \mathrm{SVD}(\mW - \widetilde{\mW})$.

As noted in~\Cref{sec:related_work},
most existing works~\citep{li2023loftq,yao2023zeroquant,guo2023lqlora} in QPEFT and PTQ adopt this solution.
However, we know that minimizing the weight approximation error is not equivalent to minimizing the layer output error.
Furthermore, does minimizing the weight approximation error for each layer in a network effectively reduce the final model output error?
We will show that the answer is negative in~\Cref{sec:experiments:peft}.

\begin{problem}[Minimization of layer output error]\label{problem:minimize-output-error}
For a pretrained linear layer $\vy=\vx\mW$ and its approximated form $\widetilde{\vy}=\vx(\widetilde{\mW}+\mC_k)$,
approximating the layer by minimizing the error between $\vy$ and $\widetilde{\vy}$ is to minimize the following expectation.
\begin{equation}\label{eq:problem2_objective}
    \argmin_{\mC_k}\E_{\vy \sim \sY} \{ || \widetilde{\vy} - \vy ||_2^2 \}
\end{equation}
where $||\cdot||_2$ denotes $l_2$ norm, and $\sY \subseteq \sR^n$ is output space of the layer.
We expand~\Cref{eq:problem2_objective} by substituting $\widetilde{\vy}$ and $\vy$:
\begin{equation}\label{eq:problem2_objective_expanded}
    \argmin_{\mC_k}\E_{\vx \sim \sX} \{ || \vx(\widetilde{\mW} + \mC_k) - \vx\mW ||_2^2 \}
\end{equation}
where $\sX \subseteq \sR^m$ is the input space.
In practice, the expectation can be approximated as a sample mean on a calibration dataset like a subset of the pretraining data set.
\end{problem}

Problem~\ref{problem:minimize-output-error} motivates some recent works~\citep{liu2023qllm,guo2023lqlora,zhang2024lqer} to involve activation-induced heuristics
in the optimization of $\mC_k$ \textit{but without a theoretical foundation}.
%
In the following two sections, we will derive the analytical solution to Problem~\ref{problem:minimize-output-error}.
More precisely, we present two solutions: one exact solution in~\Cref{sec:method:proof_exact} and an approximated solution based on a suitable statistical assumption in~\Cref{sec:method:proof-approx}.

\subsection{\texorpdfstring{\loqerrxx{}}{LoQER+ (rxx)}: Analytical Solution}
\label{sec:method:proof_exact}

\loqerrxx{} is our exact solution to Problem~\ref{problem:minimize-output-error}.
\loqerrxx{} is computationally expensive as it calculates the autocorrelation matrix of the input space $\sX$.
However, as we will show in~\Cref{sec:experiments}, \loqerrxx{} recovers significant model performance in extremely low-precision quantization.

\begin{theorem}[\loqerrxx{} solution]\label{theorem:loqerrxx}
    The solution to Problem~\ref{problem:minimize-output-error} is 
    \begin{equation}
        \begin{split}
            \mC_k & = \left(\mR_{\sX\sX}^{\frac{1}{2}}\right)^{-1}\mU_{:,:k} \mSigma_{:k,:k} \mV^T_{:k,:}
        \end{split}
    \end{equation}
    where $\mR_{\sX\sX}$ is the autocorrelation matrix respect to the input space $\sX$,
    \begin{equation}
        \mR_{\sX\sX} = \E_{\vx \sim \sX} \left\{ \vx^T \vx  \right\}
    \end{equation}
    $\mR_{\sX\sX}^{\frac{1}{2}}$ represents the unique symmetric positive semi-definite matrix square root of $\mR_{\sX\sX}$, and\newline
    $\mU_{:,:k}$, $\mSigma_{:k,:k}$, and $\mV_{:k,:}$ form the truncated SVD of the following scaled weight error matrix,
    \begin{equation}
        \mU\mSigma\mV^T = \mathrm{SVD}(\mR_{\sX\sX}^{\frac{1}{2}}(\mW - \widetilde{\mW}))
    \end{equation}

\end{theorem}

\begin{remark}\label{remark:inveritble-rxx}
    $\mR_{\sX\sX}^{\frac{1}{2}}$ is positive semi-definite. In the event that it has a zero eigenvalue, it would be normal to add a small diagonal perturbation to recover invertibility.
    In practice, we ran extensive experiments and find that $\mR_{\sX\sX}^{\frac{1}{2}}$ is invertible for all the pretrained models and datasets we present in~\Cref{sec:experiments}.
\end{remark}

\textbf{Proof of Theorem~\ref{theorem:loqerrxx}}
\begin{proof}
    Define $\mP:=\widetilde{\mW} + \mC_k - \mW$,
    and $\vp_i:=\mP_{i,:}$ is the $i$-th row of $\mP$.
    Then we substitute $(\widetilde{\mW} + \mC_k - \mW)$ in the expanded objective~\Cref{eq:problem2_objective_expanded} of Problem~\ref{problem:minimize-output-error} with $\mP$:
    \begin{equation} \label{eq:proof-approx_expansion_1}
        \begin{split}
            \E_{\vy \sim \sY} \{ || \widetilde{\vy} - \vy ||_2^2 \}
             & = \E_{\vx \sim \sX} \{ || \vx\mP ||_2^2 \}                                               \\
             & = \E_{\vx \sim \sX} \{ || \sum_{i=1}^{m} x_i \vp_i ||_2^2 \}                             \\
             & = \E_{\vx \sim \sX} \left\{ \sum_{i=1}^{m} \sum_{j=1}^{m} x_i x_j \vp_i \vp_j^T \right\}
        \end{split}
    \end{equation}
    We rewrite the last line of~\Cref{eq:proof-approx_expansion_1} as:
    \begin{equation}
        \begin{split}
            \E_{\vy \sim \sY} \{ || \widetilde{\vy} - \vy ||_2^2 \}
            %
            %
             &
            = \E_{\vx \sim \sX} \left\{\ve \cdot \left( (\vx^T\vx) \odot (\mP\mP^T)\right) \cdot \ve^T \right\}
        \end{split}
    \end{equation}
    where $\ve=\begin{bmatrix}1 & 1 & \hdots & 1\end{bmatrix}$ is a row vector of $m$ ones, and $\odot$ denotes the element-wise product.

    Using the property of the element-wise product~\citep{styan1973hadamard}, the RHS of the above can be simplified.
    \begin{equation}\label{eq:proof_exact_expansion:after_elementwise_product}
        \begin{split}
            \E_{\vy \sim \sY} \{ || \widetilde{\vy} - \vy ||_2^2 \}
             & = \E_{\vx \sim \sX} \left\{
            \Tr\left((\vx^T \vx) (\mP\mP^T)^T\right)
            \right\}                                                                      \\
             & = \Tr\left(\E_{\vx \sim \sX} \left\{ \vx^T \vx  \right\} \mP \mP^T \right) \\
             & = \Tr\left( \mR_{\sX\sX} \mP \mP^T \right)
        \end{split}
    \end{equation}
    where $\Tr(\cdot)$ denotes trace and $\mR_{\sX\sX} = \E_{\vx \sim \sX} \left\{ \vx^T \vx  \right\}$ is the autocorrelation matrix with respect to the input space $\sX$.

    Since $\mR_{\sX\sX}$ is a symmetric positive semi-definite matrix, it always has precisely one matrix square root,
    denoted as $\mR_{\sX\sX}^{\frac{1}{2}}$,
    that is also symmetric and positive semi-definite~\citep{horn2012matrix}. We reorganize~\Cref{eq:proof_exact_expansion:after_elementwise_product} as the following
    since both $\mR_{\sX\sX}$ and $(\mP\mP^T)$ are symmetric and positive semi-definite:
    \begin{equation}
        \begin{split}
            \E_{\vy \sim \sY} \{ || \widetilde{\vy} - \vy ||_2^2 \}
             & = \Tr\left( \mR_{\sX\sX}^{\frac{1}{2}} \mP \mP^T \mR_{\sX\sX}^{\frac{1}{2}} \right)     \\
             & = \Tr\left( \mR_{\sX\sX}^{\frac{1}{2}} \mP \mP^T (\mR_{\sX\sX}^{\frac{1}{2}})^T \right) \\
             & = || \mR_{\sX\sX}^{\frac{1}{2}} \mP ||_F^2
        \end{split}
    \end{equation}
    Now the objective of Problem~\ref{problem:minimize-output-error} (~\Cref{eq:problem2_objective}) is equivalent to:
    \begin{equation}
        \begin{split}
            \argmin_{\mC_k} \E_{\vy \sim \sY} \{ || \widetilde{\vy} - \vy ||_2^2 \}
             & = \argmin_{\mC_k} || \mR_{\sX\sX}^{\frac{1}{2}} \mP ||_F^2                            \\
             & = \argmin_{\mC_k} || \mR_{\sX\sX}^{\frac{1}{2}}(\widetilde{\mW} + \mC_k - \mW) ||_F^2
        \end{split}
    \end{equation}
    If we assign $\mQ :=  \mR_{\sX\sX}^{\frac{1}{2}}(\mW - \widetilde{\mW})$ and $\mQ_k :=  \mR_{\sX\sX}^{\frac{1}{2}}\mC_k$, the objective becomes:
    \begin{equation} \label{eq:proof_exact_final_objective}
        \argmin_{\mQ_k} || \mQ_k - \mQ ||_F^2
    \end{equation}
    Note that multiplication by the invertible matrix $\mR_{\sX\sX}^{\frac{1}{2}}$ (Remark~\ref{remark:inveritble-rxx}) does not change the rank of the matrix $\mC_k$.
    According to the Eckart-Young-Mirsky theorem~\citep{eckart1936approximation},
    the optimal rank $k$ approximation to $\mQ_k$ is the truncated SVD of $\mQ$:

    \begin{equation}\label{eq:proof_exact_final_objective_svd}
        \mQ_k = \mU_{:,:k} \mSigma_{:k,:k} \mV^T_{:k,:}
    \end{equation}
    where $\mU\mSigma\mV^T = \mathrm{SVD}(\mQ) = \mathrm{SVD}\left(\mR_{\sX\sX}^{\frac{1}{2}}(\mW - \widetilde{\mW})\right)$.
    %
    %
    Thus the optimal rank-$k$ solution to $\mC_k$ is:
    \begin{equation}
        \begin{split}
            \mC_k = \left(\mR_{\sX\sX}^{\frac{1}{2}}\right)^{-1} \mQ_k
            = \left(\mR_{\sX\sX}^{\frac{1}{2}}\right)^{-1}\mU_{:,:k} \mSigma_{:k,:k} \mV^T_{:k,:}
        \end{split}
    \end{equation}
\end{proof}
\ifdefined\isiclr
    \vspace{-1em}
\fi
In practice, we assign $\mA_k := \left(\mR_{\sX\sX}^{\frac{1}{2}}\right)^{-1}\mU_{:,:k}$ and $\mB_k := \mSigma_{:k,:k}\mV_{:k,:}^T$.
Note that \loqer{} adds no constraints to the quantization (and dequantization) function $\mathrm{q}(\cdot)$ (and $\mathrm{dq}(\cdot)$),
\ie{}, the low-precision $\widetilde{\mW}$ can be obtained by any quantization method.

\subsection{\texorpdfstring{\loqerdiag{}}{LoQER+ (diag)}: An Analytical Solution with the Uncorrelated Assumption}
\label{sec:method:proof-approx}

\loqerdiag{} is our analytical solution to Problem~\ref{problem:minimize-output-error}
based on the assumption that different embedding dimensions are uncorrelated.
This solution is more computationally efficient than the exact solution,
and the assumption is testable on real-world datasets.
The complete proof of \loqerdiag{} is in~\Cref{sec:proof-approx:complete}.

\begin{assumption}\label{assumption:independent-embedding}
    For a pretrained linear layer $\vy=\vx\mW$, the expectation of the product of different embedding dimensions is zero:
    \begin{equation}
        \E_{\vx \sim \sX} \{ x_i x_j \} = 0, \quad \forall i\neq j
    \end{equation}
    where $x_i$ and $x_j$ are the $i$-th and $j$-th elements of the input vector $\vx$.
\end{assumption}

We test this assumption on LLMs in~\Cref{sec:discussion}.

\begin{theorem}[\loqerdiag{} solution]\label{theorem:loqerdiag}
    The solution to Problem~\ref{problem:minimize-output-error} based on Assumption~\ref{assumption:independent-embedding} is:
    \begin{equation}
        \begin{split}
            \mC_k & = \mS^{-1}\mU_{:,:k} \mSigma_{:k,:k} \mV^T_{:k,:} 
        \end{split}
    \end{equation}
    where $\mS$ is a diagonal matrix built from activation statistics,
    \begin{equation}
        \mS = \mathrm{diag}(\sqrt{\E_{\vx \sim \sX} \{ x_1^2 \}}, \sqrt{\E_{\vx \sim \sX} \{ x_2^2 \}}, \hdots, \sqrt{\E_{\vx \sim \sX} \{ x_m^2 \}})
    \end{equation}
    and $\mU$, $\mSigma$, $\mV^T$ form the SVD of the following scaled weight error matrix,
    \begin{equation}
        \mU\mSigma\mV^T = \mathrm{SVD}(\mS(\mW - \widetilde{\mW}))
    \end{equation}
\end{theorem}
\begin{remark}
    For the diagonal matrix $\mS$ in Theorem~\ref{theorem:loqerdiag} to be invertible,
    we need $\E_{\vx \sim \sX} \{ x_i^2 \} \neq 0$ for all dimension $i$.
    In practice, this is almost always true for pretrained layers because
    no dimension in the input embeddings is always zero.
\end{remark}


For implementation, we assign $\mA_k := \mS^{-1}\mU_{:,:k}$ and $\mB_k := \mSigma_{:k,:k}\mV_{:k,:}$ to
form the low-rank terms to save the memory and computation cost.
Interestingly, \loqerdiag{} solution is similar to the activation-induced heuristics
in \LQER{}~\citep{zhang2024lqer}, which calibrates the average absolute value on the embedding dimension (Refer to \Cref{alg:lqer} in the Appendix).
In~\Cref{sec:experiments:ptq}, we will show that our solution is more effective in practice and
resolves the discrepancy between the recovered model performance and the number of calibration samples in \LQER{}.

%% file: sections/4_experiments.tex
\ifdefined\isiclr
    \vspace{-0.5em}
\fi
\section{Experiments}
\label{sec:experiments}
\ifdefined\isiclr
    \vspace{-0.5em}
\fi

In this section, we first introduce the experiment setup in~\Cref{sec:experiments:setup}.
Then we present the results of our experiments on QPEFT and PTQ in~\Cref{sec:experiments:peft} and~\Cref{sec:experiments:ptq} respectively.

\subsection{Experiment Setup}\label{sec:experiments:setup}

We perform QPEFT and PTQ experiments separately, and compare with their respective SoTA methods. The experiments take around 6400 GPU hours in total. The hardware platform, separate GPU hours, software dependencies, and random seed settings can be found in~\Cref{sec:appendix:detailed-exp-setup}.

For QPEFT experiments, we use \cref{theorem:loqerdiag}, noted as \loqerdiag{}, to initialize low-rank terms, and compare with full-finetuning, LoRA~\citep{hu2021lora}, \qlora{}~\citep{dettmers2024qlora}, and \loftq{}~\citep{li2023loftq}. Specifically, we adopt 5-iteration \loftq{},
which is the officially recommended setup.
We include both encoder-only model experiments (fine-tuning \roberta{}-base~\citep{liu2019roberta} on GLUE~\citep{ye2019glue})
and decoder-only LLM experiments (fine-tuning \llama{}-2~\citep{touvron2023llama} and \llama{}-3.1~\citep{dubey2024llama} on continuous pretraining task SlimPajama~\citep{cerebras2023slimpajama} and supervised fine-tuning task GSM8K~\citep{cobbe2021gsm8k}).
For each method/baseline, we sweep the learning rate and record the best result. The final results are averaged over three random seeds.
The learning rate ranges and batch sizes are listed in~\Cref{sec:appendix:detailed-exp-setup:qpeft-hyperparameters}.

For PTQ experiments, we use both \cref{theorem:loqerrxx}, noted as \loqerrxx{}, and \cref{theorem:loqerdiag} (\loqerdiag{})
to calculate the low-rank error reconstruction terms and report results separately. We compare with BF16, quantized model without error reconstruction terms ($w$-only),
\zeroquantii{}~\citep{yao2023zeroquant}, and \LQER{}~\citep{zhang2024lqer} at different precision setups. We also include HQQ~\citep{badri2023hqq}, a leading 4-bit method that does not use quantization error reconstruction.
We quantize LLMs of various sizes and model family, including TinyLlama~\citep{zhang2024tinyllama}, Gemma-2~\citep{team2024gemma}, Phi-3.5~\citep{abdin2024phi3} and \llama{}-2/-3.1~\citep{touvron2023llama,dubey2024llama}.
We use {\small\texttt{lm-evaluation-harness}} to report results on Wikitext2~\citep{merity2016wikitext2}, ARC (challenge)~\citep{allenai:arc}, BoolQ~\citep{clark2019boolq}, CommonSenseQA~\citep{talmor2019commonsenseqa},
Winogrande~\citep{ai2:winogrande}, MMLU~\citep{hendryckstest2021mmlu}, and BigBench-Hard~\citep{suzgun2022bigbench}.
We also evaluate instruction-tuned model, Vicuna-v1.5~\citep{zheng2023vicuna}, with AlpacaEval 2.0~\citep{dubois2024alpacaeval}, which is an automatic evaluation tool for instruction-following tasks.
Detailed setup is in~\Cref{sec:appendix:detailed-exp-setup:ptq-hyperparameters}.


\ifdefined\isiclr
    \vspace{-0.5em}
\fi
\subsection{Improved QPEFT}
\label{sec:experiments:peft}
\input{figures/fig_roberta-output-error-vs-rank.tex}

We first identify a pitfall in the commonly-used iterative~\Cref{alg:loftq}, that is,
minimizing the weight approximation error for each layer does not necessarily minimize the model output error.
Then we show that our \loqer{} initialization enables a clear reduction in the model output error at the start of fine-tuning,
leading to better fine-tuned accuracy/perplexity and faster convergence.

\input{figures/tab_roberta-base-glue-train.tex}
\input{figures/tab_llama_fine-tune}

\ifdefined\isiclr
    \vspace{-1em}
\fi
\paragraph{Reduced layer weight error $\ne$ reduced model output error}
\ifdefined\isieee
    \input{figures/fig_roberta-stsb-convergence-3bit.tex}
\fi
We apply 4-bit and 3-bit \qlora{}, \loftq{}, and \loqerdiag{} to \roberta{}-base and
inspect the \textit{model output error} on \roberta{}'s pretraining dataset before fine-tuning at rank $k=4, 8, 16, 32$.
For \loftq{}, we also sweep the number of iterations from 1 to 5.
In~\Cref{fig:roberta-output}, we observe that
\ifdefined\isiclr
    \vspace{-0.5em}
\fi
\begin{itemize}
    \item For \loftq{}, given a specific rank, increasing the optimization iterations does not guarantee a reduced model output error. Though all the layers' weight approximation errors monotonically decrease with the number of iterations (as illustrated in \Cref{fig:roberta-weight-approx-error-vs-num-iters} in Appendix), the model output error does not monotonically decrease. For example, in~\Cref{fig:roberta-output-error:rank}, the model output error of \loftq{} (5-iter) is larger than \loftq{} (3-iter) at rank $k=8$.
    \item For \loftq{}, given a specific number of iterations, increasing the rank does not guarantee a reduced model output error.
          For example, in~\Cref{fig:roberta-output-error:loftq-iter}, the output error of \loftq{} (rank $k=16$) is larger than \loftq{} (rank $k=4$) and $k=8$ at 2, 3, 4, and 5 iterations.
    \item The model output error of our \loqerdiag{} is always smaller than \loftq{} and \qlora{}, across all precision and rank settings.
          Moreover, the output error of \loqerdiag{} monotonically decreases as the rank increases.
\end{itemize}
\ifdefined\isiclr
    \vspace{-0.5em}
\fi

This empirical evidence suggests a strong correlation between the reduction of layer output error and the decrease in model output error in QER problem.
Conversely, minimizing weight approximation error using LoftQ does not have a comparable impact on overall model performance.
\ifdefined\isiclr
    \input{figures/fig_roberta-stsb-convergence-3bit.tex}
    \vspace{-1em}
\fi
\paragraph{Better optimization quality}
\Cref{tab:roberta-base-glue-train} and~\Cref{tab:llama-fine-tune} summarize the fine-tuning experiments of \roberta{}-base on GLUE,
and \llama{}-2-7B/-3.1-8B fine-tuned on SlimPajama and GSM8K, respectively.
\loqer{} outperforms both \loftq{} and \qlora{}. In GLUE experiments, at 4-bit, \loqer{} enables an average accuracy gain of 0.96\% and 0.79\% higher than \qlora{} and \loftq{} respectively, close to BF16 LoRA;
At 3-bit and 2-bit, \loqer{} achieves a 4.12\% and 6.05\% higher average accuracy than \loftq{} respectively.
Similar trends are observed on LLM fine-tuning experiments, \ie{},
\loqer{} outperforms \qlora{} and \loftq{}, and
the advantage of \loqer{} over \loftq{} is more obvious with more aggressive quantization.

\ifdefined\isiclr
    \vspace{-1em}
\fi
\paragraph{Faster Convergence} \loqer{} initialization also speeds up the training convergence.
For LLM fine-tuning, this is expected as \loqer{} initialization is closer to the full-precision model.
Interestingly, in encoder-only experiments on GLUE where the model classifier head is randomly initialized,
we also observe that \loqer{} converges faster,
especially on small subsets such as STSB and MRPC where only a few thousand samples are available
(in comparison MNLI has 393k samples and QQP has 364k samples).
For example, in~\Cref{fig:roberta-stsb-convergence-3bit},
the Spearman correlation coefficient of \loqer{} on STSB increases and converges faster than \loftq{} and \qlora{}, as the green line plateaus first.

\ifdefined\isiclr
    \vspace{-0.5em}
\fi
\subsection{Improved PTQ}\label{sec:experiments:ptq}
\ifdefined\isiclr
    \vspace{-0.5em}
\fi
In this part, we first demonstrate that \LQER{}, which depends on heuristics derived from activation values, does not guarantee improved performance with a larger calibration dataset. However, \loqer{} exhibits the opposite trend.
Through extensive experiments, we show that
\loqer{} consistently outperforms \zeroquantii{} and \LQER{},
and \loqerrxx{} exhibits better model performance than \loqerdiag{} at the cost of more computation in the quantization process.
These results verified the effectiveness of our analytical solution.
\ifdefined\isiclr
    \input{figures/fig_tinyllama-ppl-vs-num-samples.tex}

\fi
\ifdefined\isieee
    \input{figures/fig_tinyllama-ppl-vs-num-samples.tex}
\fi
\paragraph{Model performance \textit{vs.} calibration set size}\label{sec:experiments:ptq:calibration-size}
As mentioned at the end of~\Cref{sec:method:proof-approx},
the scale matrix in \LQER{}~\citep{zhang2024lqer} is similar to
the one in \loqerdiag{}, but is based on hand-crafted heuristics. As a result, we observe that
the model performance of \LQER{} varies randomly as the number of calibration samples increases
(the purple curve in~\Cref{fig:tinyllama-ppl-vs-num-samples}).
On the contrary, more calibration samples consistently lead to better model performance for \loqer{} until convergence.
\ifdefined\isiclr
    \vspace{-0.5em}
\fi

\input{figures/tab_llm_ptq_perplexity.tex}
\paragraph{Improved perplexity and downsteam task accuracy}
We apply \loqerdiag{} and \loqerrxx{} to a range of models and
evaluate on both pretraining task and downstream tasks in~\Cref{tab:llm-ptq-perplexity}
and~\Cref{tab:llm-ptq-downstream} respectively.
We also compare to HQQ, a SoTA method that does not use quantization error reconstruction.
On most models, \loqerdiag{} outperforms \zeroquantii{} and \LQER{},
while \loqerrxx{} achieves the best performance.
At 4-bit, \loqerrxx{} is nearly lossless.
At 3-bit, \loqerrxx{}'s improvement over \loqerdiag{} (\Cref{tab:llm-ptq-perplexity}) is clear,
indicating the superiority of \loqerrxx{} for aggressive quantization.
\input{figures/tab_llm_ptq_downstream.tex}

\ifdefined\isiclr
    \input{figures/fig_vicuna-alpaca-eval}
    \vspace{-0.5em}
\fi
\paragraph{Higher win rate on AlpacaEval 2.0}
To better understand the impact on instruction-tuned models,
we present the results of Vicuna-7b-v1.5 on AlpacaEval 2.0.
In~\Cref{fig:vicuna-alpaca-eval}, we evaluate the QER-based methods against the $w$-only quantization counterpart.
\loqer{} outperforms \zeroquantii{} and \LQER{} by a higher win rate, indicating a better response quality.

%% file: figures/fig_roberta-output-error-vs-rank.tex
\begin{figure}
    \ifdefined\isiclr
        \vspace{-2em}
    \fi
    \centering
    \begin{subfigure}[b]{0.48\textwidth}
        \includegraphics[width=\textwidth]{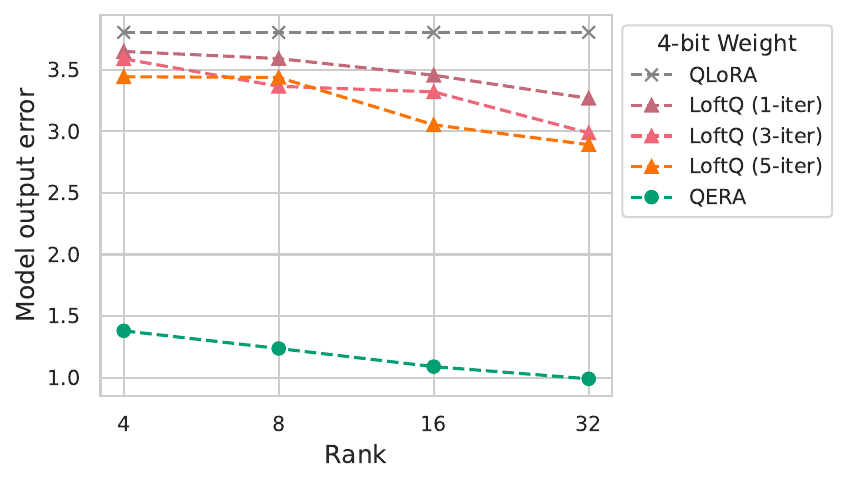}
        \caption{Model output error \textit{vs.} rank}
        \label{fig:roberta-output-error:rank}
    \end{subfigure}
    \hfill
    \begin{subfigure}[b]{0.48\textwidth}
        \includegraphics[width=\textwidth]{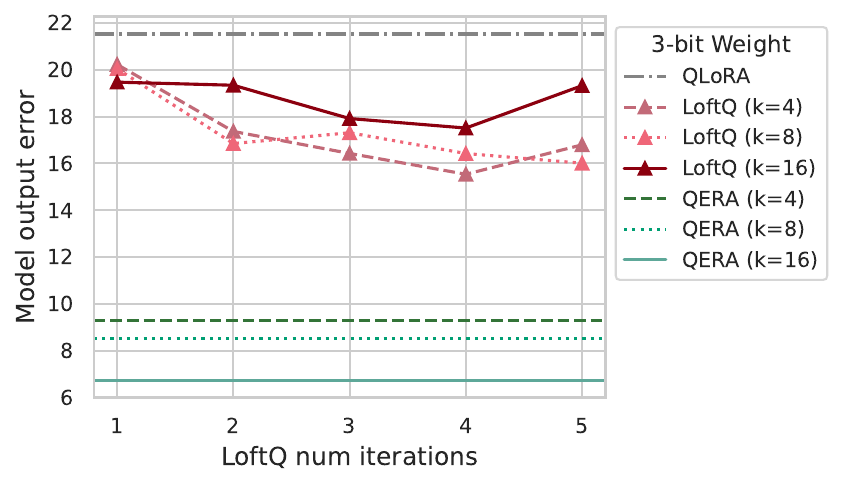}
        \caption{Model output error \textit{vs.} LoftQ iterations}
        \label{fig:roberta-output-error:loftq-iter}
    \end{subfigure}
    \caption{The model output error of \roberta{}-base before fine-tuning.
        We feed 128 samples from \roberta{}'s pretraining dataset and profile the output logits error between the adapted and the FP32 model.
        We sweep the rank $k$ and the iteration number of \loftq{} on 4-bit and 3-bit models.
        In \loftq{}, neither more iterations nor a higher rank guarantees lower model output error, though the weight approximation error of every layer decreases.
        In contrast, \loqerdiag{} consistently has the lowest model output error across all settings, and the error monotonically decreases as the rank increases.
    }
    \ifdefined\isiclr
        \vspace{-1em}
    \fi
    \label{fig:roberta-output}
\end{figure}

%% file: figures/tab_roberta-base-glue-train.tex
\begin{table}[t]
    \ifdefined\isiclr
        \vspace{-3em}
    \fi
    \caption{Fine-tuning results of \roberta{}-base on GLUE. \loqerdiag{} outperforms \loftq{} across all bit widths, and the improvement is more obvious with aggressive quantization.
        \loqer{} achieves $\Delta_{\text{acc}}$ = 4.12\%  higher than \loftq{} at 3-bit and 6.05\% at 2-bit.}
    \ifdefined\isiclr
        \vspace{-1em}
    \fi
    \begin{center}
        \begin{small}
            \resizebox{\linewidth}{!}{
                \begin{tabular}{@{}cclccccccccc@{}}
                    \toprule
                    \multirow{2}{*}{Rank} & \multirow{2}{*}{W-bits} & \multirow{2}{*}{Method} & MNLI         & QNLI         & RTE          & SST          & MRPC         & CoLA         & QQP          & STSB                      & \multirow{2}{*}{Avg.} \\
                                          &                         &                         & Acc          & Acc          & Acc          & Acc          & Acc          & Matt         & Acc          & P/S Corr                  &                       \\ \midrule
                    -                     & 16                      & Full FT                 & 87.61        & 92.95        & 73.16        & 94.88        & 92.15        & 60.41        & 91.61        & 90.44/90.25               & 85.38                 \\ \midrule
                    8                     & 16                      & LoRA                    & 87.85        & 92.84        & 69.55        & 94.46        & 89.99        & 57.52        & 89.83        & 89.92/89.83               & 84.00                 \\ \midrule
                    \multirow{3}{*}{8}    & \multirow{3}{*}{4.25}   & \qlora{}                & 87.21        & 92.32        & 63.90        & 94.08        & 88.24        & \good{56.08} & \good{90.55} & 89.59/89.56               & 82.75                 \\
                                          &                         & \loftq{} (5-iter)       & 87.27        & \good{92.48} & 67.13        & \good{94.38} & 88.24        & 54.59        & 90.51        & 88.75/88.79               & 82.92                 \\
                                          &                         & \loqerdiag{}            & \good{87.28} & 92.45        & \good{70.40} & \good{94.38} & \good{88.97} & 55.99        & 90.39        & \good{89.83}/\good{89.72} & \good{83.71}          \\ \midrule
                    \multirow{3}{*}{8}    & \multirow{3}{*}{3.25}   & \qlora{}                & 84.87        & 89.58        & 53.67        & 91.02        & 73.94        & 3.12         & 89.31        & 84.80/84.38               & 71.29                 \\
                                          &                         & \loftq{} (5-iter)       & 85.24        & 89.65        & 58.24        & 92.05        & 75.82        & 11.00        & 88.93        & 85.55/85.27               & 73.31                 \\
                                          &                         & \loqerdiag{}            & \good{85.58} & \good{90.74} & \good{58.48} & \good{92.59} & \good{82.19} & \good{32.98} & \good{89.41} & \good{87.43}/\good{87.08} & \good{77.43}          \\ \midrule
                    \multirow{3}{*}{64}   & \multirow{3}{*}{2.50}   & \qlora{}                & 77.87        & 85.26        & 54.15        & 90.02        & 71.00        & 0            & 87.93        & 74.72/75.31               & 67.62                 \\
                                          &                         & \loftq{} (5-iter)       & 80.15        & 87.65        & 52.95        & 90.94        & 74.35        & 3.43         & 89.17        & 82.76/82.90               & 70.18                 \\
                                          &                         & \loqerrxx{}             & \good{84.64} & \good{90.05} & \good{58.48} & \good{92.32} & \good{84.72} & \good{26.43} & \good{89.69} & \good{86.48}/\good{86.40} & \good{76.23}          \\ \bottomrule
                \end{tabular}
            }
        \end{small}
    \end{center}
    \label{tab:roberta-base-glue-train}
\end{table}

%% file: figures/tab_llama_fine-tune.tex
\begin{table}[h]
    \caption{Fine-tuning results of \llama{}-2-7B and \llama{}-3.1-8B on SlimPajama and GSM8K. A trend similar to \roberta{} experiments are observed, \ie{},
        \loqer{} outperforms \qlora{} and \loftq{} and the improvement is more obvious on aggressive quantization.}
    \ifdefined\isiclr
        \vspace{-1em}
    \fi
    \begin{center}
        \begin{small}
            \begin{tabular}{@{}clcccc@{}}
                \toprule
                \multirow{2}{*}{W-bits} & \multirow{2}{*}{Method} & \multicolumn{2}{c}{LLaMA-2-7B}     & \multicolumn{2}{c}{LLaMA-3.1-8B}                                                                      \\ \cmidrule(l){3-6}
                                        &                         & SlimPajama ($\Delta_{\text{ppl}}$) & GSM8K ($\Delta_{\text{acc}}$)    & SlimPajama ($\Delta_{\text{ppl}}$) & GSM8K ($\Delta_{\text{acc}}$) \\ \midrule
                16                      & LoRA                    & 6.17                               & 39.40                            & 8.07                               & 55.72                         \\ \midrule
                \multirow{3}{*}{4.25}   & \qlora{}                & 6.44 (+0.27)                       & 30.71 (-8.69)                    & 8.70 (+0.63)                       & 54.81 (-0.91)                 \\
                                        & \loftq{} (5-iter)       & 6.39 (+0.22)                       & 28.58 (-10.82)                   & 8.73 (+0.66)                       & 54.23 (-1.49)                 \\
                                        & \loqerdiag{}            & \good{6.33 (+0.16)}                & \good{32.26 (-7.14)}             & \good{8.68 (+0.61)}                & \good{55.24 (-0.48)}          \\ \midrule
                \multirow{3}{*}{2.25}   & \qlora{}                & 53.95 (+47.78)                     & 12.79 (-18.31)                   & 71.90 (+63.83)                     & 5.08 (-50.64)                 \\
                                        & \loftq{} (5-iter)       & 12.30 (+6.13)                      & 18.37 (-12.73)                   & 27.16 (+19.09)                     & 13.72 (-42.00)                \\
                                        & \loqerdiag{}            & \good{10.56 (+4.39)}               & \good{18.78 (-12.32)}            & \good{20.07 (+12.00)}              & \good{19.41 (-36.31)}         \\ \bottomrule
            \end{tabular}
        \end{small}
    \end{center}
    \ifdefined\isiclr
        \vspace{-1.5em}
    \fi
    \label{tab:llama-fine-tune}
\end{table}

%% file: figures/fig_roberta-stsb-convergence-3bit.tex
\begin{wrapfigure}{R}{0.36\textwidth}
    \ifdefined\isiclr
        \vspace{-1em}
    \fi
    \centering
    \includegraphics[width=0.35\textwidth]{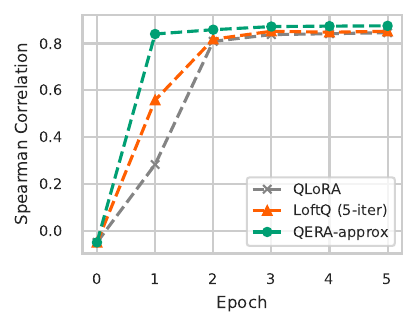}
    \ifdefined\isiclr
        \vspace{-0.5em}
    \fi
    \caption{Faster convergence of \loqerdiag{} on STSB.}
    \ifdefined\isiclr
        \vspace{-1em}
    \fi
    \label{fig:roberta-stsb-convergence-3bit}
\end{wrapfigure}

%% file: figures/fig_tinyllama-ppl-vs-num-samples.tex
\begin{wrapfigure}{R}{0.38\textwidth}
    \ifdefined\isiclr
        \vspace{-2em}
    \fi
    \centering
    \includegraphics[width=0.35\textwidth]{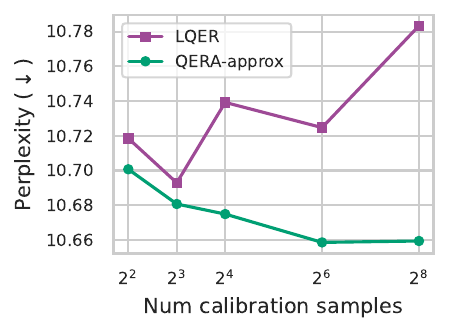}
    \ifdefined\isiclr
        \vspace{-1em}
    \fi
    \caption{\loqer{} resolves the discrepancy between the recovered model performance and the number of calibration samples in \LQER{}.}
    \ifdefined\isieee
        \vspace{1em}
    \fi
    \label{fig:tinyllama-ppl-vs-num-samples}
\end{wrapfigure}

%% file: figures/tab_llm_ptq_perplexity.tex
\begin{table}[t]
    \ifdefined\isiclr
        \vspace{-3em}
    \fi
    \caption{Perplexity ($\downarrow$) of LLMs on WikiText2.
        $w$-only denotes the quantized model without low-rank error reconstruction.
        \loqerdiag{} outperforms \LQER{} on almost all setups and \loqerrxx{} achieves the lowest perplexity.
        The advantage of \loqer{} is pronounced at 3-bit.}
    \ifdefined\isiclr
        \vspace{-1em}
    \fi
    \begin{center}
        \begin{small}
            \begin{tabular}{@{}clcccccccc@{}}
                \toprule
                \multirow{2}{*}{W-bits} & \multirow{2}{*}{Method} & \multirow{2}{*}{Rank} & TinyLlama      & Gemma-2               & Phi-3.5               & \multicolumn{2}{c}{LLaMA-2} & \multicolumn{2}{c}{LLaMA-3.1}                                               \\ \cmidrule(l){4-10}
                                        &                         &                       & 1.1B           & 2B                    & 3.8B                  & 7B                          & 13B                           & 8B                   & 80B                  \\ \midrule
                -                       & BF16                    & -                     & 13.98          & 13.08                 & 11.50                 & 8.71                        & 7.68                          & 7.55                 & 3.06                 \\ \midrule
                4.25                    & HQQ                     & -                     & \textit{15.02} & 14.29                 & 14.63                 & 9.59                        & 8.27                          & 8.72                 & 3.97                 \\ \midrule
                \multirow{5}{*}{4.25}   & $w$-only                & -                     & 19.40          & 16.23                 & 14.16                 & 9.45                        & 8.06                          & 8.78                 & 4.55                 \\
                                        & ZeroQuant-V2            & \multirow{4}{*}{32}   & 18.03          & 15.71                 & 14.09                 & 9.42                        & 8.07                          & 8.83                 & 4.48                 \\
                                        & LQER                    &                       & 16.23          & 14.55                 & 12.88                 & 9.22                        & 7.96                          & 8.45                 & 4.10                 \\
                                        & QERA-approx             &                       & \good{15.66}   & 14.60                 & 12.81                 & 9.17                        & 7.95                          & 8.45                 & 4.10                 \\
                                        & QERA-exact              &                       & 16.16          & \textit{\good{14.12}} & \textit{\good{12.30}} & \textit{\good{9.12}}        & \textit{\good{7.93}}          & \textit{\good{8.33}} & \textit{\good{3.82}} \\ \midrule
                \multirow{5}{*}{3.25}   & $w$-only                & -                     & 32.82          & 41.13                 & 47.78                 & 13.32                       & 10.24                         & 18.96                & 16.46                \\
                                        & ZeroQuant-V2            & \multirow{4}{*}{64}   & 27.80          & 33.56                 & 42.64                 & 13.00                       & 10.03                         & 19.29                & 10.12                \\
                                        & LQER                    &                       & 20.60          & 21.99                 & 18.27                 & 14.00                       & 9.09                          & 11.86                & 7.05                 \\
                                        & QERA-approx             &                       & 20.43          & 21.93                 & \good{17.99}          & 10.99                       & 9.04                          & 11.73                & 6.99                 \\
                                        & QERA-exact              &                       & \good{19.51}   & \good{19.97}          & 20.37                 & \good{10.67}                & \good{8.97}                   & \good{11.39}         & \good{6.68}          \\ \bottomrule
            \end{tabular}
        \end{small}
    \end{center}
    \ifdefined\isiclr
        \vspace{-1em}
    \fi
    \label{tab:llm-ptq-perplexity}
\end{table}

%% file: figures/tab_llm_ptq_downstream.tex
\begin{table}[t]
    \vspace{-3em}
    \caption{Average accuracy ($\uparrow$) of LLMs on six downstream tasks.
        \loqerrxx{} outperforms other quantization-error reconstruction-based methods across almost all models.
        We also compare to HQQ~\citep{badri2023hqq}, a SoTA PTQ method that does not adopt quantization-error reconstruction or activation heuristics. \loqerrxx{} achieves an average accuracy on par with HQQ.
    }
    \ifdefined\isiclr
        \vspace{-1em}
    \fi
    \begin{center}
        \begin{small}
            \begin{tabular}{@{}clcccccccc@{}}
                \toprule
                \multirow{2}{*}{W-bits} & \multirow{2}{*}{Method} & \multirow{2}{*}{Rank} & TinyLlama\protect\footnotemark & Gemma-2        & Phi-3.5      & \multicolumn{2}{c}{LLaMA-2} & \multicolumn{2}{c}{LLaMA-3.1}                                          \\ \cmidrule(l){4-10}
                                        &                         &                       & 1.1B                           & 2B             & 3.8B         & 7B                          & 13B                           & 8B             & 80B                   \\ \midrule
                -                       & BF16                    & -                     & 40.59                          & 53.96          & 66.91        & 49.61                       & 55.74                         & 63.88          & 72.05                 \\ \midrule
                4.25                    & HQQ                     & -                     & 40.35                          & \textit{52.54} & 59.17        & 48.26                       & 54.53                         & \textit{62.59} & 71.31                 \\ \midrule
                \multirow{5}{*}{4.25}   & $w$-only                & -                     & 36.56                          & 48.33          & 64.52        & 47.62                       & 55.12                         & 61.53          & 68.46                 \\
                                        & \zeroquantii{}          & \multirow{4}{*}{32}   & 37.26                          & 48.24          & 64.44        & 47.43                       & 55.15                         & 61.70          & 68.45                 \\
                                        & \LQER{}                 &                       & \textit{\good{40.45}}          & 49.77          & 64.46        & 48.47                       & 55.40                         & 61.75          & 70.94                 \\
                                        & \loqerdiag{}            &                       & 40.02                          & 49.29          & 64.53        & 48.52                       & 55.20                         & 61.68          & 70.80                 \\
                                        & \loqerrxx{}             &                       & 40.36                          & \good{51.73}   & \good{65.08} & \textit{\good{48.91}}       & \textit{\good{55.42}}         & \good{62.05}   & \textit{\good{71.42}} \\ \bottomrule 
            \end{tabular}
        \end{small}
    \end{center}
    \ifdefined\isiclr
        \vspace{-1em}
    \fi
    \label{tab:llm-ptq-downstream}
\end{table}
\footnotetext{The average accuracy of TinyLlama-1.1B excludes BoolQ, CommonsenseQA, and MMLU since TinyLlama-1.1B has random guess accuracy on these tasks.}

%% file: figures/fig_vicuna-alpaca-eval.tex
\begin{wrapfigure}{R}{0.38\textwidth}
    \ifdefined\isiclr
        \vspace{-2em}
    \fi
    \centering
    \includegraphics[width=0.35\textwidth]{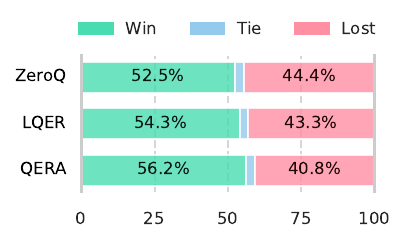}
    \ifdefined\isiclr
        \vspace{-1em}
    \fi
    \caption{AlpacaEval 2.0 evaluation results. We compare quantized models to the counterpart without quantization-error reconstruction.
        A higher \textcolor{Green}{win rate} ($\uparrow$) indicates better instruction-following performance.}
    \ifdefined\isiclr
        \vspace{-2em}
    \fi
    \label{fig:vicuna-alpaca-eval}
\end{wrapfigure}

%% file: sections/5_discussion.tex
\ifdefined\isiclr
    \vspace{-0.5em}
\fi
\ifdefined\isieee
    \newpage
\fi
\section{Discussion}
\label{sec:discussion}
\ifdefined\isiclr
    \vspace{-0.5em}
\fi
\ifdefined\isieee
    \input{figures/fig_vicuna-alpaca-eval}
\fi
In this section, we revisit the arguments, design choices, and observations made
in the previous sections, including a test of Assumption~\ref{assumption:independent-embedding},
and the choice of the calibration set for PEFT.
We offer an extended discussion of the numeric stability and scalability in~\Cref{sec:appendix:scalability-stability},
and LoRA rank and model choices of PEFT experiments in~\Cref{sec:appendix:choice-of-ranks-and-models}.

\vspace{-0.5em}

\paragraph{Test of Assumption~\ref{assumption:independent-embedding}}\label{sec:discussion:test-independent-embedding-dim}
To test Assumption~\ref{assumption:independent-embedding},
we profile the autocorrelation matrix $\mR_{\sX\sX}$ of the linear layer inputs in \llama{}-2-7B and \llama{}-3-8B.
Note that ${\R_{\sX\sX}}_{i,j} =  \E_{\vx \sim \sX} \{ x_i x_j \}$,
which assumes to be zero for $i\ne j$ in Assumption~\ref{assumption:independent-embedding}.
~\Cref{fig:llama-3-rxx-example}
shows the normalized ${\R_{\sX\sX}}$ magnitude, $\frac{\mathrm{abs}(\R_{\sX\sX})}{||\R_{\sX\sX} ||_F}$, of four representative layers in \llama{}-3-8B
where darker elements denote values closer to zero. There are several layers with some input dimensions strongly correlated with others,
such as the inputs to the third attention layer in~\Cref{fig:llama3-rxx-example:first-k},
but for most layers, our assumption holds, especially the MLP layers, such as~\Cref{fig:llama-3-rxx-example:normal-k,fig:llama-3-rxx-example:normal-o,fig:llama-3-rxx-example:normal-down}.
More ${\R_{\sX\sX}}$ plots are in~\Cref{sec:appendix:test-assumption}.

\ifdefined\isiclr
    \input{figures/fig_llama-3-8b_rxx_example}
    \vspace{-0.5em}
\fi

\paragraph{Choice of calibration set for QPEFT} One problem is to determine the calibration set for \loqer{} before fine-tuning.
In 2-bit \roberta{}-base fine-tuning experiment on SST2 (\Cref{sec:appendix:choice-of-calibration-set-roberta}),
we find that calibrating on the pretraining dataset, WikiText2, helps the loss to decrease.
However, the loss of the model calibrated on the fine-tuning dataset does not follow the same trend.
We hypothesize that the massive padding tokens in preprocessed SST2 samples cause this discrepancy, especially considering
that the sequence length of the raw SST2 dataset changes fiercely.
\ifdefined\isieee
    \input{figures/fig_llama-3-8b_rxx_example}
\fi

%% file: figures/fig_llama-3-8b_rxx_example.tex
\begin{figure}[h]
    \ifdefined\isiclr
        \vspace{-1em}
    \fi
    \centering
    \begin{subfigure}[b]{0.22\textwidth}
        \includegraphics[height=\textwidth]{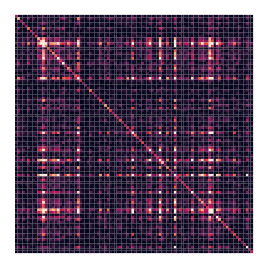}
        \ifdefined\isiclr
            \vspace{-1em}
        \fi
        \caption{{\small \texttt{3.o\_proj}}}
        \label{fig:llama3-rxx-example:first-k}
    \end{subfigure}
    \hfill
    \begin{subfigure}[b]{0.22\textwidth}
        \includegraphics[height=\textwidth]{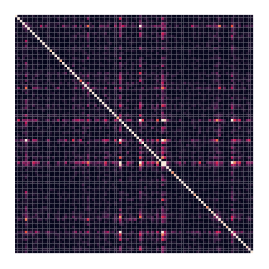}
        \ifdefined\isiclr
            \vspace{-1em}
        \fi
        \caption{{\small\texttt{7.k/q/v\_proj}}}
        \label{fig:llama-3-rxx-example:normal-k}
    \end{subfigure}
    \hfill
    \begin{subfigure}[b]{0.22\textwidth}
        \includegraphics[height=\textwidth]{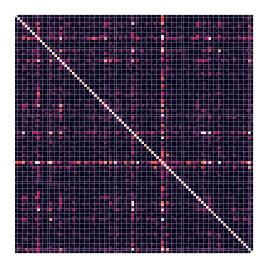}
        \ifdefined\isiclr
            \vspace{-1em}
        \fi
        \caption{{\small\texttt{7.o\_proj}}}
        \label{fig:llama-3-rxx-example:normal-o}
    \end{subfigure}
    \hfill
    \begin{subfigure}[b]{0.22\textwidth}
        \includegraphics[height=\textwidth]{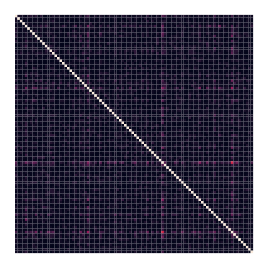}
        \ifdefined\isiclr
            \vspace{-1em}
        \fi
        \caption{{\small\texttt{7.gate\_proj}}}
        \label{fig:llama-3-rxx-example:normal-down}
    \end{subfigure}
    \hfill
    \ifdefined\isiclr
        \vspace{-0.5em}
    \fi
    \caption{Normalized $\mathrm{abs}(\mR_{\sX\sX})$ of the layer inputs in LLaMA-3-8B. Dark elements denotes value close to zero.
        There are a few layers with input dimensions strongly correlated with others,
        such as the third attention layer in (a), but for most layers,
        our assumption of zero-expectation holds.
    }
    \label{fig:llama-3-rxx-example}
    \ifdefined\isiclr
        \vspace{-1em}
    \fi
\end{figure}

%% file: sections/6_conclusion.tex
\ifdefined\isiclr
    \vspace{-0.5em}
\fi
\section{Conclusion}
\label{sec:conclusion}
\ifdefined\isiclr
    \vspace{-0.5em}
\fi

In this paper, we formulate the problem of quantization error reconstruction and propose \loqer{} as an analytical solution.
Applying \loqer{} to related works for efficient fine-tuning or inference,
we show that \loqer{} resolves the discrepancy in existing methods,
and outperforms SoTA methods in both fine-tuning and quantization tasks by a clear margin.

%% file: sections/7_acknowlegement.tex
\subsubsection*{Acknowledgments}

This work was sponsored by \href{https://www.aria.org.uk/}{Advanced Research + Invention Agency (ARIA), UK}. 
We also thank ARIA for their research network.

%% file: sections/99_appendix.tex
\section{Appendix}
\label{sec:appendix}

\subsection{Algorithms in Related Work}

Here we summarize the algorithm of~\loftq{}~\citep{li2023loftq} in~\Cref{alg:loftq} and~\LQER{}~\citep{zhang2024lqer} in~\Cref{alg:lqer} respectively.
LQ-LoRA~\citep{guo2023lqlora} adopts a variant of~\Cref{alg:loftq}.
\zeroquantii{}~\citep{yao2023zeroquant} can be considered as~\Cref{alg:loftq} with one iteration, or a special case of~\Cref{alg:lqer} where the scale matrix $\mS$ is an identity matrix.

\ifdefined\isiclr
    \input{figures/alg_loftq.tex}
    \input{figures/alg_lqer.tex}
\else
    \input{figures/alg_loftq_ieee}
    \input{figures/alg_lqer_ieee}
\fi

\subsection{Proof of Theorem~\ref{theorem:loqerdiag}}\label{sec:proof-approx:complete}
Here we present the full proof of~\loqerdiag{}. \loqerdiag{} is an approximated solution to Problem~\ref{problem:minimize-output-error} based on Assumption~\ref{assumption:independent-embedding}, which is suitable to initialize the low-rank terms in fine-tuning for lower computation complexity.

\input{sections/999_proof_approx}

\input{sections/999_appendix_caldera.tex}

\subsection{Detailed Experiment Setup}\label{sec:appendix:detailed-exp-setup}

We mainly use \href{https://pytorch.org/}{\texttt{\small{PyTorch}}}, \href{https://github.com/huggingface/transformers}{\texttt{\small{Transformers}}}, \href{https://github.com/huggingface/peft}{\texttt{\small{PEFT}}}, and \href{https://github.com/huggingface/accelerate/tree/main}{\texttt{\small{Accelerate}}} to implement \loqer{}. We use \href{https://docs.scipy.org/doc/scipy/index.html}{\texttt{\small SciPy}}'s implementation of blocked Schur algorithm~\citep{deadman2012blockedsqrtm} to calculate the matrix square root, which runs on CPUs. The evaluation is performed with \href{https://github.com/EleutherAI/lm-evaluation-harness}{\texttt{\small lm-evaluation-harness}}, \href{https://github.com/huggingface/evaluate/tree/main}{\texttt{\small Evaluate}}, and \href{https://github.com/tatsu-lab/alpaca_eval}{AlpacaEval 2.0}~\citep{dubois2024alpacaeval}.

\subsubsection{QPEFT Hyperparameters}\label{sec:appendix:detailed-exp-setup:qpeft-hyperparameters}
We perform fine-tuning experiments on four NVIDIA A100 80GB GPUs with AMD EPYC 64-Core Processor with 1024GB RAM.
The total fine-tuning time is around 2100 GPU hours.

\paragraph{\roberta{}-base on GLUE} We sweep learning rates for each ({\small \texttt{method}, \texttt{task}}), and collect the best accuracy.
Thus each ({\small \texttt{method}, \texttt{task}}) pair has its own tailored learning rate,
ensuring the best performance of baselines and \loqer{} under the same trainable parameter budget.
The reported accuracy is the average value across random seeds 42, 1, and 2.
The total batch size is 64 for all GLUE experiments and we train the models for 5 epochs. For 4-bit experiments, we use 4-bit floating point from the \qlora{} implementation in \texttt{\small PEFT}.
For 3-bit experiments, we use emulated MXINT~\citep{darvish2023mxformat} with block size = 32 and for 2-bit experiments we use MXINT with block size = 16.
\Cref{tab:hyperparams-roberta-glue} lists the learning rates for each experiment.

\paragraph{\llama{}-2-7B/-3.1-8B on SlimPajama and GSM8K} We adopt the learning rates in~\citet{meng2024pissa}. The reported perplexity/accuracy is the average value across random seeds 42, 1, and 2. For SlimPajama, we fine-tune the model on a subset for 1000 steps with rank = 8, total batch size = 64, sequence length = 1024, learning rate = 3e-5. For GSM8K, we fine-tune the model for 10 epochs with rank = 64, total batch size = 128, sequence length = 384, and learning rate = 3e-5.

\input{figures/tab_hyper-params-roberta-glue}

\subsubsection{PTQ Hyperparameters}\label{sec:appendix:detailed-exp-setup:ptq-hyperparameters}
We perform PTQ experiments on eight NVIDIA A6000 48GB GPUs with AMD EPYC 256-Core Processor with 1024GB RAM.
The total quantization and evaluation time is around 4500 GPU hours. We report 0-shot accuracy or normalized accuracy (if available) for all tasks except WikiText2, in which we report word perplexity.
The sequence length for reporting word perplexity is the model's context length by default, except for Phi-3.5 and \llama{}-3.1. For these two models, we set the sequence length = 2048.
We use the HuggingFace \texttt{\small Transformers}'s implementation of HQQ, and reimplement \zeroquantii{} and \LQER{} as baselines.
We use MXINT with block size = 32 as the quantization format for all quantization methods except HQQ, which uses its built-in INT format with group size = 64.
Thus, both formats have an average W-bits of 4.25.
We evaluate quantized Vicuna-v1.5-7B, which is an instruction-tuned \llama{}-2-7B, with AlpacaEval 2.0. and use GPT4-Turbo as the evaluator. The reported win rate is the length-controlled win rate, which is a debiased version of the win rate that controls for the length of the generated outputs.

\subsection{Decreasing Weight Error \texorpdfstring{$\ne$}{!=} Decreasing Output Error for \loftq{}}

We provide the weight approximation error, $||\mW - \widetilde{\mW} - \mC_k||_F$, in~\Cref{fig:roberta-weight-approx-error-vs-num-iters}, of all linaer layers in \roberta{}-base by sweeping the number of iterations.
We observe that the weight approximation error monotonically decreases with the number of iterations, but as shown in~\Cref{fig:roberta-output}, the model output error may increase.
This observation indicates that the commonly used objective of minimizing the weight approximation error and the corresponding algorithm are not ideal for the quantization error reconstruction problem.

\input{figures/fig_roberta-weight-approx-error-vs-num-iters.tex}

\subsection{Choice of Calibration Set}\label{sec:appendix:choice-of-calibration-set-roberta}

We compare the \loqer{}-adapted models calibrated on the pretraining dataset and the downstream dataset.
Specifically, we fine-tune two \loqer{}-adapted 2-bit \roberta{}-base models. One is calibrated on its pretraining dataset, WikiText2, and the other on SST2.
~\Cref{fig:roberta-qera-calibrated-on-pretraining-vs-downstream} shows the loss curves of the two models across three learning rates.
None loss curves of the models calibrated on SST2 decreases,
but the ones calibrated on WikiText2 successfully decrease and converge.
We hypothesize that this is due to the massive padding tokens in preprocessed SST2 considering that the raw sample lengths change fiercely.
However, WikiText2 samples were preprocessed in the masked language modeling style, which means that only a few special tokens are added to the grouped texts.

\input{figures/fig_roberta-qera-calibrated-on-pretraining-vs-downstream}

\input{figures/fig_stability-and-scalability}
\subsection{Scalability and Numerical Stability of \loqer{}}\label{sec:appendix:scalability-stability}
One may notice the diminishing model performance improvement of \loqerrxx{} over \loqerdiag{} as the model size increases.
The main reason is that larger LLMs are more resistant to quantization~\citep{chee2024quip}.
Another reason can be the error ratio of the matrix square root calculation of the autocorrelation matrix increases with model hidden size (\Cref{fig:scalability-and-stability:sqrtm-error-vs-hidden-size}). 

We find that the data type used in the calibration is important for the numeric stability of \loqerrxx{} due to the calculation of the matrix square root and SVD.
To improve the stability of the calculation in \loqerrxx{}, a good practice we find is to perform the outer product of $\mR_{\sX\sX}$ in FP32,
accumulated outer product in FP64, and calculate the matrix square root in FP64 using the blocked Schur algorithm~\citep{deadman2012blockedsqrtm}. \Cref{fig:scalability-and-stability:runtime-vs-model-size} illustrates the quantization time of~\loqerdiag{} and~\loqerrxx{} on the platform described in~\Cref{sec:appendix:detailed-exp-setup} where the linear layers are quantized sequentially. \loqerrxx{} is slow due to the calculation of matrix square roots on CPUs. GPU-accelerated matrix square root will be the key optimization to reduce the quantization time. Note that in \loqer{}, the quantization of individual layers is independent, allowing more parallelization and acceleration of the quantization process.
%

\subsection{Choice of Solutions for QPEFT and PTQ}\label{sec:appendix:choice-of-solutions-for-qpeft-ptq}


QPEFT and PTQ are two different application scenarios of \loqer{}.
We recommend \loqerdiag{} for QPEFT and \loqerrxx{} for PTQ.
PTQ aims to recover the model performance as much as possible without re-training.
For PTQ, it is desirable to recover more model performance even if it takes longer to compute low-rank terms.
Note that the low-rank terms are pre-computed once offline.
At inference time, \loqerrxx{} introduces no overhead to the hardware
since \LQER{}, \loqerdiag{}, and \loqerrxx{} all takes the same form of $\vy=\vx(\widetilde{\mW}+\mA_k\mB_k)$.

However, for QPEFT experiments, it is unreasonable to pay a long time for initializing the low-rank terms
for the limited improvement in output approximation error (\text{i.e.}, QERA-exact/\caldera{}),
because 1) fine-tuning can recover the error, and 2) instead of spending much time on initialization,
increasing training steps or increasing the rank number brings more gain in the fine-tuned accuracy.
We run controlled experiments to support this claim.
In~\Cref{tab:appendix:choice-of-solutions-mrpc} and~\Cref{tab:appendix:choice-of-solutions-slimpajama},
we run QPEFT experiments of RoBERTA-base on MRPC and \llama{}-2-7B on SlimPajama respectively.
Compared to QERA-exact (Caldera's Lemma 4.2), QERA-approx achieves better accuracy/perplexity while taking $\frac{2}{3}\sim \frac{1}{2}$ of the time.

\input{figures/tab_choice_of_solutions_mrpc.tex}
\input{figures/tab_choice_of_solutions_slimpajama.tex}

\subsection{Choices of LoRA Ranks, Models, and Precisions for QPEFT}\label{sec:appendix:choice-of-ranks-and-models}

\paragraph{Rank = 8 for GLUE experiments} We notice \loftq{} paper uses a large rank of 16 and 32 for fine-tuning on GLUE,
which is larger than the commonly-used rank value of LoRA (4 or 8 in LoRA paper~\citep{hu2021lora}).
If we consider LoRA as the upper limit of \qlora{}-like QPEFT methods (including \loftq{} and \loqer{}),
to effectively compare these QPEFT methods, one easy way is to set the rank as the minimum value required by LoRA
and check which QPEFT method achieves an accuracy closest to LoRA.
This is why we choose rank = 8 for GLUE experiments (For 2-bit GLUE experiments we use a large rank 64 since the quantization is very aggressive).
If we use rank = 32, LoRA and all the QPEFT methods may be over-parameterized and it will be hard to make a fair comparison in terms of fine-tuned accuracy.
To support this claim, we sweep the rank of LoRA-adapted RoBERTA-base on SST2 and MRPC
and show a large rank $k$ like 16 in LoftQ has over-parallelization problem in~\Cref{tab:appendix:over-parameterization-sst2} and~\Cref{tab:appendix:over-parameterization-mrpc}.

\paragraph{\roberta{} \textit{vs.} DeBERTa} When investigating the related work,
we find that both \roberta{} and DeBERTaV3~\citep{he2021debertav3} are used in QPEFT experiments~\citep{guo2023lqlora,li2023loftq,meng2024pissa,guo2023lqlora,zhang2023adalora}.
The reason why we chose \roberta{} is that the \roberta{} checkpoint on HuggingFace\footnote{\roberta{}-base checkpoint:
    \href{https://huggingface.co/FacebookAI/roberta-base}{link}} is complete
and compatible with both HuggingFace's official examples of sequence classification\footnote{HuggingFace example of sequence classification: \href{https://github.com/huggingface/transformers/tree/main/examples/pytorch/text-classification}{link}}
and masked language modeling\footnote{HuggingFace example of masked language modeling: \href{https://github.com/huggingface/transformers/tree/main/examples/pytorch/language-modeling}{link}}.
Specifically, the \roberta{} checkpoint contains both the base model and the masked language modeling head but the DeBERTaV3's checkpoint\footnote{DeBERTaV3's checkpoint: \href{https://huggingface.co/microsoft/deberta-v3-base}{link}} only contains the base model. As we know, the base model is enough for fine-tuning on downstream tasks.
However, to calibrate on the pretraining dataset, we need the language modeling head to verify if our implementation of data preprocessing and calibration matches how the model was originally pretrained. Note that the quality of the statistic values in~\loqer{} like ${\R_{\sX\sX}}$ depends on the quality of the calibration set.
Thus, without the language modeling head in the checkpoint, we cannot perform the \loqer{}'s calibration for DeBERTaV3 properly, ensure the correctness of statistics in~\loqer{}, and explore the effect of the choice of calibration sets.

\input{figures/tab_choose_qpeft_rank_sst2.tex}
\input{figures/tab_choose_qpeft_rank_mrpc.tex}

\subsection{Detailed PTQ Results}
\label{sec:appendix:detailed-ptq-results}

Here we offer the detailed evaluation results for each downstream task in~\Cref{tab:ptq-downstream-detailed:tinyllama,tab:ptq-downstream-detailed:gemma2-2b,tab:ptq-downstream-detailed:phi-3.5-mini,tab:ptq-downstream-detailed:llama-2-7b,tab:ptq-downstream-detailed:llama-2-13b,tab:ptq-downstream-detailed:llama-31-8b,tab:ptq-downstream-detailed:llama-31-70b}.

\input{figures/tab_llm_ptq_downstream_detailed}

\subsection{Test of Assumption \ref{assumption:independent-embedding}}
\label{sec:appendix:test-assumption}

We provide more plots of normalized ${\R_{\sX\sX}}$ magnitude, $\frac{\mathrm{abs}(\R_{\sX\sX})}{||\R_{\sX\sX} ||_F}$, across \llama{}-3.1-8B, \llama{}-2-7B, Mistral-7B-v0.3, and TinyLlama-1.1B
in~\Cref{fig:llama_3_8b_rxx_k_proj,fig:llama_3_8b_rxx_o_proj,fig:llama_3_8b_rxx_gate_proj,fig:llama_3_8b_rxx_down_proj,fig:llama_2_7b_rxx_k_proj,fig:llama_2_7b_rxx_o_proj,fig:llama_2_7b_rxx_gate_proj,fig:llama_2_7b_rxx_down_proj,fig:mistral_7b_rxx_k_proj,fig:mistral_7b_rxx_o_proj,fig:mistral_7b_rxx_gate_proj,fig:mistral_7b_rxx_down_proj,fig:tinyllama_rxx_k_proj,fig:tinyllama_rxx_o_proj,fig:tinyllama_rxx_gate_proj,fig:tinyllama_rxx_down_proj},
where dark pixels are elements close to zeros. There are strongly correlated embedding channels in some \texttt{\small k\_proj} and \texttt{\small o\_proj} layers.
The assumption fits better in MLP layers (\texttt{\small gate\_proj}, \texttt{\small up\_proj}, and \texttt{\small down\_proj}), and holds for over 60\% of the layers in LLMs.

\input{figures/fig_llama-3-8b_rxx_plots.tex}
\input{figures/fig_llama-2-7b_rxx_plots.tex}
\input{figures/fig_mistral-7b_rxx_plots.tex}
\input{figures/fig_tinyllama-1b_rxx_plots.tex}

%% file: figures/alg_loftq.tex
\vspace{-0.5em}
\begin{algorithm}
    \caption{LoftQ~\citep{li2023loftq}}\label{alg:loftq}
    \begin{small}
    \begin{algorithmic}[1]
        \Require {Pretrained weight $\mW$, target rank $k$,
            quantization function $\mathrm{q}(\cdot)$,
            dequantization function $\mathrm{dq}(\cdot)$, number of iterations $T$}
        \State {$\mA_{k} \leftarrow \mathbf{0}, \mB_{k} \leftarrow \mathbf{0}$}
        \For {$i$ = $1$ to $T$}
        \State {$\mW_{q} \leftarrow \mathrm{q}(\mW - \mA_{k}\mB_{k})$} \Comment{Update quantized weight matrix}
        \State {$\widetilde{\mW} \leftarrow \mathrm{dq}(\mW_{q})$}
        \State {$\mU, \mSigma, \mV^T \leftarrow \mathrm{SVD}(\mW - \widetilde{\mW})$} \Comment{SVD-based rank-$k$ approximation}
        \State {$\mA_{k} \leftarrow \mU_{:,:k} \sqrt{\mSigma_{:k,:k}}$, $\mB_{k} \leftarrow \sqrt{\mSigma_{:k,:k}}\mV^T_{:k,:}$}
        \EndFor
    \end{algorithmic}
    \end{small}
\end{algorithm}
\vspace{-0.5em}

%% file: figures/alg_lqer.tex
\begin{algorithm}
    \caption{LQER~\citep{zhang2024lqer}}\label{alg:lqer}
    \begin{small}
    \begin{algorithmic}[1]
        \Require {Pretrained weight $\mW$, target rank $k$,
            quantization function $\mathrm{q}(\cdot)$,
            dequantization function $\mathrm{dq}(\cdot)$,
            calibration dataset $\sX=\{\vx_i \in \sR^{m} | i=1,\dots,N\}$}
        \State {Initialize vector $\vs \leftarrow \mathbf{0}$}
        \For {sample $\vx$ in $\sX$} \Comment{Calibration}
        \State {$\vs \leftarrow \vs + \mathrm{abs}(\vx)$} \Comment{Accumulate activation magnitude on each dimension}
        \EndFor
        \State {$\mS\leftarrow  \frac{1}{N}\mathrm{diag}(\vs)$} \Comment{Construct a diagonal matrix $\mS$}
        \State {$\mW_q \leftarrow \mathrm{q}(\mW)$}
        \State {$\widetilde{\mW} \leftarrow \mathrm{dq}(\mW_q)$}
        \State {$\mU, \mSigma, \mV^T \leftarrow \mathrm{SVD}(\mS (\mW-\ \widetilde{\mW}))$} \Comment{SVD on the scaled weight error}
        \State {$\mA_k \leftarrow \mS^{-1}\mU_{:,:k}$, $\mB_k \leftarrow \mSigma_{:k,:k}\mV^T_{:k,:}$} \Comment{Rank-$k$ approximation with un-scaling}
    \end{algorithmic}
    \end{small}
\end{algorithm}

%% file: figures/alg_loftq_ieee.tex
\begin{algorithm}[H]
    \caption{LoftQ~\citep{li2023loftq}}\label{alg:loftq}
    \begin{small}
    \begin{algorithmic}[1]
        \REQUIRE {Pretrained weight $\mW$, target rank $k$,
            quantization function $\mathrm{q}(\cdot)$,
            dequantization function $\mathrm{dq}(\cdot)$, number of iterations $T$}
        \STATE {$\mA_{k} \leftarrow \mathbf{0}, \mB_{k} \leftarrow \mathbf{0}$}
        \FOR {$i$ = $1$ to $T$}
        \STATE {$\mW_{q} \leftarrow \mathrm{q}(\mW - \mA_{k}\mB_{k})$} \COMMENT{Update quantized weight matrix}
        \STATE {$\widetilde{\mW} \leftarrow \mathrm{dq}(\mW_{q})$}
        \STATE {$\mU, \mSigma, \mV^T \leftarrow \mathrm{SVD}(\mW - \widetilde{\mW})$} \COMMENT{SVD-based rank-$k$ approximation}
        \STATE {$\mA_{k} \leftarrow \mU_{:,:k} \sqrt{\mSigma_{:k,:k}}$, $\mB_{k} \leftarrow \sqrt{\mSigma_{:k,:k}}\mV^T_{:k,:}$}
        \ENDFOR
    \end{algorithmic}
    \end{small}
\end{algorithm}

%% file: figures/alg_lqer_ieee.tex
\begin{algorithm}[H]
    \caption{LQER~\citep{zhang2024lqer}}\label{alg:lqer}
    \begin{small}
    \begin{algorithmic}[1]
        \REQUIRE {Pretrained weight $\mW$, target rank $k$,
            quantization function $\mathrm{q}(\cdot)$,
            dequantization function $\mathrm{dq}(\cdot)$,
            calibration dataset $\sX=\{\vx_i \in \sR^{m} | i=1,\dots,N\}$}
        \STATE {Initialize vector $\vs \leftarrow \mathbf{0}$}
        \FOR {sample $\vx$ in $\sX$} 
        \STATE {$\vs \leftarrow \vs + \mathrm{abs}(\vx)$} \COMMENT{Calibrate by accumulating activation magnitude on each dimension}
        \ENDFOR
        \STATE {$\mS\leftarrow  \frac{1}{N}\mathrm{diag}(\vs)$} \COMMENT{Construct a diagonal matrix $\mS$}
        \STATE {$\mW_q \leftarrow \mathrm{q}(\mW)$}
        \STATE {$\widetilde{\mW} \leftarrow \mathrm{dq}(\mW_q)$}
        \STATE {$\mU, \mSigma, \mV^T \leftarrow \mathrm{SVD}(\mS (\mW-\ \widetilde{\mW}))$} \COMMENT{SVD on the scaled weight error}
        \STATE {$\mA_k \leftarrow \mS^{-1}\mU_{:,:k}$, $\mB_k \leftarrow \mSigma_{:k,:k}\mV^T_{:k,:}$} \COMMENT{Rank-$k$ approximation with un-scaling}
    \end{algorithmic}
    \end{small}
\end{algorithm}

%% file: sections/999_proof_approx.tex
\textbf{Proof of Theorem~\ref{theorem:loqerdiag}}
\begin{proof}
    %
    We continue at~\Cref{eq:proof-approx_expansion_1}.
    Since $\E_{\vx \sim \sX}$ is the expectation with respect to the input space,
    we move the expectation inside the summation of RHS of~\Cref{eq:proof-approx_expansion_1}.
    \begin{equation} \label{eq:proof-approx_expansion_2}
        \begin{split}
            \E_{\vy \sim \sY} \{ || \widetilde{\vy} - \vy ||_2^2 \} = \sum_{i=1}^{m} \sum_{j=1}^{m} \E_{\vx \sim \sX} \{ x_i x_j \} \vp_i \vp_j^T
        \end{split}
    \end{equation}
    Under Assumption~\ref{assumption:independent-embedding},
    $\E_{\vx \sim \sX} \{ x_i x_j \} = 0$ for $i\neq j$,
    the RHS of~\Cref{eq:proof-approx_expansion_2} simplifies to:
    \begin{equation}\label{eq:proof-approx_expansion_3}
        \begin{split}
            \E_{\vy \sim \sY} \{ || \widetilde{\vy} - \vy ||_2^2 \}
             & = \sum_{i=1}^{m} \E_{\vx \sim \sX} \{ x_i^2 \} \vp_i \vp_i^T
        \end{split}
    \end{equation}
    We can define diagonal matrix $\mS = \mathrm{diag}(\sqrt{\E_{\vx \sim \sX} \{ x_1^2 \}}, \sqrt{\E_{\vx \sim \sX} \{ x_2^2 \}}, \hdots, \sqrt{\E_{\vx \sim \sX} \{ x_m^2 \}})$
    and rewrite the RHS of~\Cref{eq:proof-approx_expansion_3} as:
    \begin{equation} \label{eq:proof-approx_expansion_4}
        \begin{split}
            \E_{\vy \sim \sY} \{ || \widetilde{\vy} - \vy ||_2^2 \}
             & = \Tr(\mS \mP \mP^T \mS^T)
            = || \mS \mP ||_F^2
        \end{split}
    \end{equation}
    where $\Tr(\cdot)$ denotes the trace of a matrix.

    Therefore, the objective of Problem 2 (\Cref{eq:problem2_objective}) is equivalent to:
    \begin{equation}
        \begin{split}
            \argmin_{\mC_k}\E_{\vy \sim \sY} \{ || \widetilde{\vy} - \vy ||_2^2 \}
             & = \argmin_{\mC_k} || \mS \mP ||_F^2                            \\
             & = \argmin_{\mC_k} || \mS(\widetilde{\mW} + \mC_k - \mW) ||_F^2
        \end{split}
    \end{equation}
    If we assign $\mQ = \mS(\mW - \widetilde{\mW})$ and $\mQ_k = \mS\mC_k$, the objective becomes:
    \begin{equation} \label{eq:proof-approx_final_objective}
        \argmin_{\mQ} || \mQ_k - \mQ ||_F^2
    \end{equation}
    Note that the invertible matrix $\mS$ in $\mQ_k$ does not change the rank of the matrix $\mC_k$.
    According to the Eckart-Young-Mirsky theorem,
    the optimal rank $k$ approximation to $\mQ$ is the truncated SVD of $\mQ$:

    \begin{equation}\label{eq:proof-approx_final_objective_svd}
        \mQ_k = \mU_{:,:k} \mSigma_{:k,:k} \mV^T_{:k,:}
    \end{equation}
    where $\mU\mSigma\mV^T = \mathrm{SVD}(\mQ) = \mathrm{SVD}\left(\mS(\mW - \widetilde{\mW})\right)$.

    Finally, we get the optimal solution to the low-rank term $\mC_k$:
    \begin{equation}
        \begin{split}
            \mC_k = \mS^{-1}\mQ_k
            = \mS^{-1}\mU_{:,:k} \mSigma_{:k,:k} \mV^T_{:k,:}
        \end{split}
    \end{equation}

\end{proof}

%% file: sections/999_appendix_caldera.tex
\subsection{Connection and Difference between \caldera{} and QERA}
\label{sec:appendix:caldera_vs_qera}

\caldera{}~\citep{saha2024caldera} is the concurrent work close to QERA.
Here we elaborate the connection and difference between \caldera{} and QERA,
and highlight the contributions of QERA.

\caldera{} focuses on a different problem setup. Specifically, \caldera{} focuses on the following problem:
\begin{equation}
    \mathrm{min}_{\widetilde{\mW}, \mA_{k,q}, \mB_{k,q}} || \mX \mW - \mX (\widetilde{\mW} + \mA_{q,k}\mB_{q,k})||_F^2
\end{equation}\label{eq:appendix:caldera_problem}
where $\mX\in\sR^{b\times m}$ denotes a batch of calibration samples, and $\widetilde{\mW}$, $\mA_{k,q}$, and $\mB_{k,q}$ are all quantized variables to optimize.
Note that this problem setup is different from QERA (\Cref{eq:problem2_objective_expanded}):
\begin{equation}
    \argmin_{\mC_k}\E_{\vx \sim \sX} \{ || \vx(\widetilde{\mW} + \mC_k) - \vx\mW ||_2^2 \}
\end{equation}\label{eq:appendix:qera_problem}
where only the low-rank high-precision $\mC_k:=\mA_k\mB_k$ is the variable to optimize,
and the quantized weight $\widetilde{\mW}$ is predefined given a quantization method.

\input{figures/tab_caldera_notation.tex}

We find that \caldera{}'s Lemma 4.2 is equivalent to Theorem~\ref{theorem:loqerrxx} in QERA.
Note that the proof of QERA-exact is different from Caldera's Lemma 4.2, though the final closed-form solution is equivalent.
Here we additionally show the derivation of the equivalence between QERA-exact and Caldera's Lemma 4.2 using the notation table in \Cref{tab:caldera_equivalence_notation}.
For convenience, we remove the quantized weight term $\widetilde{\mW}$ from QERA (Problem 2 in \Cref{eq:problem2_objective_expanded}),
which does not change the proof.
Now the problem becomes finding the optimal low-rank approximation of the weight matrix, $\mC_k$ that minimizes the layer output error.

First we note that the objective of QERA,~\Cref{eq:problem2_objective_expanded}, is equivalent to \caldera{}'s Eq(5) scaled by a constant $n$:
\begin{equation}\label{eq:appendix:different_problem_setup}
    \begin{split}
         & \text{QERA}:    \text{min}_{\mC_k} E_\mathbf{x}\{||\mathbf{x}(\mC_k - \mW)||_2^2\} \\
         & \text{CALDERA}: \text{min}_{\mC_k}||\mX(\mC_k-\mW)||_F^2
    \end{split}
\end{equation}

Then we show that Theorem~\ref{theorem:loqerrxx} (QERA-exact) is equal to Caldera's Lemma 4.2.
\begin{equation}\label{eq:appendix:ck_qera_exact}
    \text{QERA-exact}: \mC_k=(\mR_{\sX\sX}^\frac{1}{2})^{-1} \cdot\mathrm{SVD}_k(\mR_{\sX\sX}^\frac{1}{2}\mW)
\end{equation}
\begin{equation}\label{eq:appendix:ck_caldera}
    \text{CALDERA}:    \mC_k'=\mV\mSigma \cdot\mathrm{SVD}_k(\mU^T\mY)
\end{equation}
We firstly show that $(\mR_{\sX\sX}^\frac{1}{2})^{-1}$ in~\Cref{eq:appendix:ck_qera_exact}
equals to $\mV\mSigma$ in~\Cref{eq:appendix:ck_caldera} scaled by a constant ${\sqrt{b}}$:
\begin{equation}
    \begin{split}
        \mR_{\sX\sX}                    & =\frac{1}{b} (\mX^T\mX)=\mV\mSigma \mU^T \mU\mSigma \mV^T=\mV\mSigma^2 \mV^T \\
        \mR_{\sX\sX}^\frac{1}{2}        & =\frac{1}{\sqrt{b}} \mSigma \mV^T                                            \\
        (\mR_{\sX\sX}^\frac{1}{2})^{-1} & =\sqrt{b}\mV\mSigma^{-1}
    \end{split}
\end{equation}
Then we show that $\mR_{\sX\sX}^\frac{1}{2}\mW$ in~\Cref{eq:appendix:ck_qera_exact} equals to $\mU^T\mY$ in~\Cref{eq:appendix:ck_caldera} scaled by the constant $\frac{1}{\sqrt{b}}$:
\begin{equation}
    \begin{split}
        \mU^T\mY                    & = \mU^T\mX\mW = \mU^T\mU\mSigma \mV^T \mW = \mSigma \mV^T \mW={\sqrt{b}}\mR_{\sX\sX}^\frac{1}{2}\mW \\
        \mR_{\sX\sX}^\frac{1}{2}\mW & =\frac{1}{\sqrt{b}} \mU^T\mY
    \end{split}
\end{equation}
Therefore $C_k$ equals to $C_k'$, and the two solutions are equivalent.
Despite of the equivalence, we shortlist the differences between \caldera{} and our work:
\begin{itemize}
    \item Different problem setup (\Cref{eq:appendix:different_problem_setup}).
    \item We simplify QERA-exact and derive QERA-approx, which is a computationally-efficient approximated solution. Specifically, QERA-approx is more suitable for parameter-efficient fine-tuning than QERA-exact/\caldera{}.
          Moreover, QERA-approx overcomes the pitfalls in existing methods and explains why previous heuristic methods like \LQER{} work.
    \item The optimization objective is similar (vector form \textit{vs} matrix form), and the final closed-form solution is equivalent, but the proof of QERA-exact is different from \caldera{}.
\end{itemize}

%% file: figures/tab_caldera_notation.tex
\begin{table}[H]
    \caption{Notation Table for the Equivalence Derivation}
    \ifdefined\isiclr
        \vspace{-1em}
    \fi
    \begin{center}
        \begin{small}
            \begin{tabular}{@{}lll@{}}
                \toprule
                Notation                & Description                                & Comments           \\ \midrule
                $b$                     & Number of calibration samples (vectors)    &                    \\
                $m$                     & Layer input feature size                   &                    \\
                $n$                     & Layer output feature size                  &                    \\
                $\mX$                   & Calibration set                            & Shape: $b\times m$ \\
                $\vx$                   & A sample in the calibration set            & Shape: $1\times m$ \\
                $\mW$                   & Original full-precision layer weights      & Shape: $m\times n$ \\
                $\mY$                   & Layer output matrix corresponding to $\mX$ & Shape: $b\times n$ \\
                $\vy$                   & Layer output vector corresponding to $\vx$ & Shape: $1\times n$ \\
                $k$                     & Rank of the low-rank approximation         &                    \\
                $\mC_k$                 & Approximated rank-$k$ weight               & Shape: $m\times n$ \\
                $\mU, \mSigma, \mV$     & SVD decomposition of $\mX$                 &                    \\
                $\mathrm{SVD}_k(\cdot)$ & Truncated rank-$k$ SVD                     &                    \\ \bottomrule
            \end{tabular}
        \end{small}
    \end{center}
    \label{tab:caldera_equivalence_notation}
\end{table}

%% file: figures/tab_hyper-params-roberta-glue.tex
\begin{table}[h]
    \caption{Learning rates of \roberta{}-base experiments on GLUE.}
    \ifdefined\isiclr
        \vspace{-1em}
    \fi
    \begin{center}
        \begin{small}

            \begin{tabular}{@{}cclc@{}}
                \toprule
                Rank & W-bits & Method                         & Learning rates                           \\ \midrule
                -    & 16     & Full FT                        & 7e-5, 5e-5, 2e-5                         \\ \midrule
                8    & 16     & LoRA                           & 1e-4, 2e-4, 3e-4                         \\
                8    & 4.25   & \qlora{}/\loftq{}/\loqerdiag{} & 1e-4, 2e-4, 3e-4                         \\
                8    & 3.25   & \qlora{}/\loftq{}/\loqerdiag{} & 1e-4, 2e-4, 3e-4                         \\
                64   & 2.50   & \qlora{}/\loftq{}/\loqerrxx{}  & 2e-5, 3e-5, 4e-5, 5e-5, 6e-5, 9e-5, 1e-4 \\ \bottomrule
            \end{tabular}

        \end{small}
    \end{center}
    \ifdefined\isiclr
        \vspace{-1em}
    \fi
    \label{tab:hyperparams-roberta-glue}
\end{table}

%% file: figures/fig_roberta-weight-approx-error-vs-num-iters.tex
\begin{figure}[ht]
    \centering
    \includegraphics[width=\textwidth]{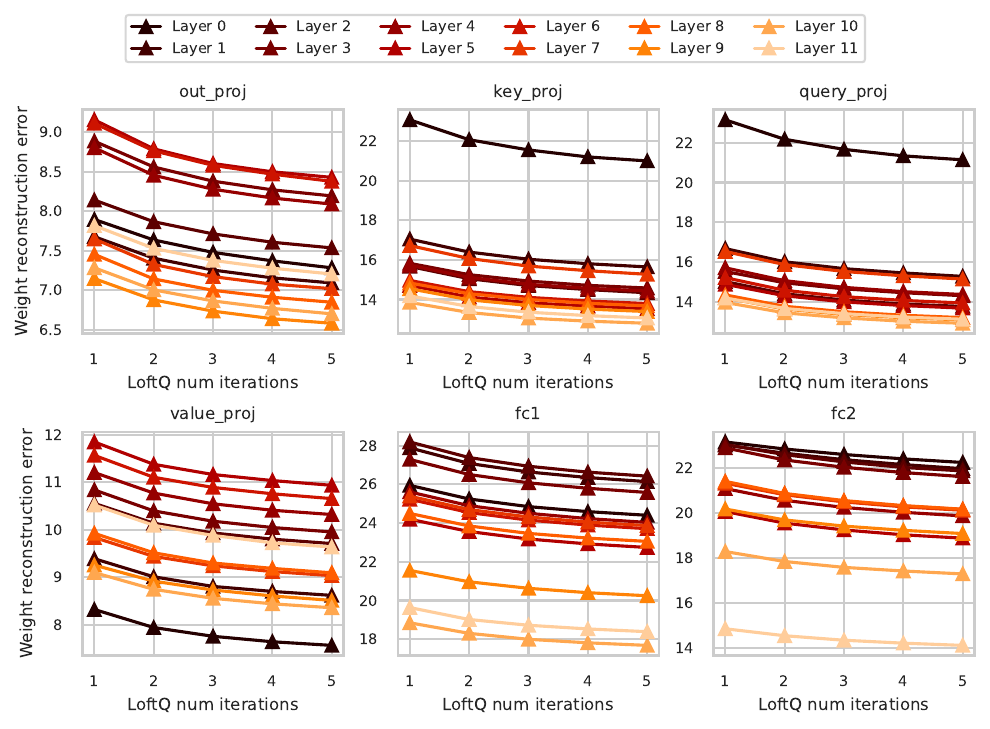}
    \caption{Weight approximation error of 3-bit rank-16 \loftq{} with different numbers of iterations on \roberta{}-base.
        We observe that the weight reconstruction error of all the layers decreases as the number of iterations increases,
        but as shown in~\Cref{fig:roberta-output-error:loftq-iter}, the model output error ($k$=16) increases from the 4-th to 5-th iteration.
    }
    \label{fig:roberta-weight-approx-error-vs-num-iters}
\end{figure}

%% file: figures/fig_roberta-qera-calibrated-on-pretraining-vs-downstream.tex
\begin{figure}[]
    \centering
    \includegraphics[width=0.9\textwidth]{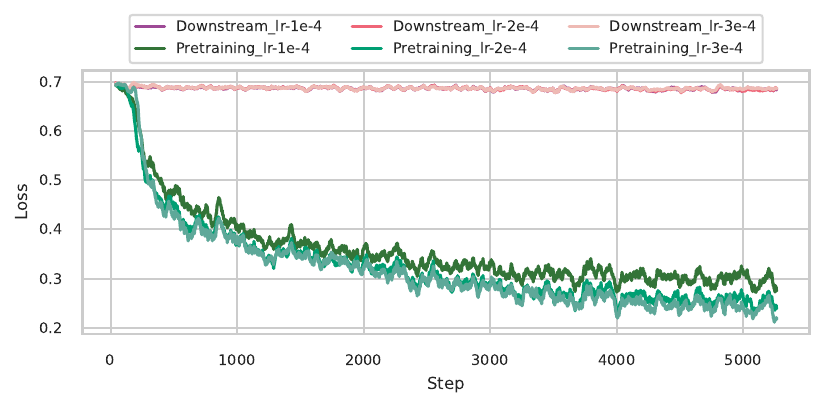}
    \caption{The fine-tuning loss curves of \loqer{}-adapted 2-bit \roberta{}-base on SST2. The loss fails to decrease if the calibration is performed on the downstream task SST2 due to the massive padding tokens in preprocessed SST2 samples. In pretraining dataset, there are only a few special tokens like padding tokens and mask tokens.}
    \label{fig:roberta-qera-calibrated-on-pretraining-vs-downstream}
\end{figure}

%% file: figures/fig_stability-and-scalability.tex
\begin{figure}
    \centering
    \ifdefined\isiclr
        \vspace{-1em}
    \fi
    \begin{subfigure}[b]{0.48\textwidth}
        \includegraphics[width=\textwidth]{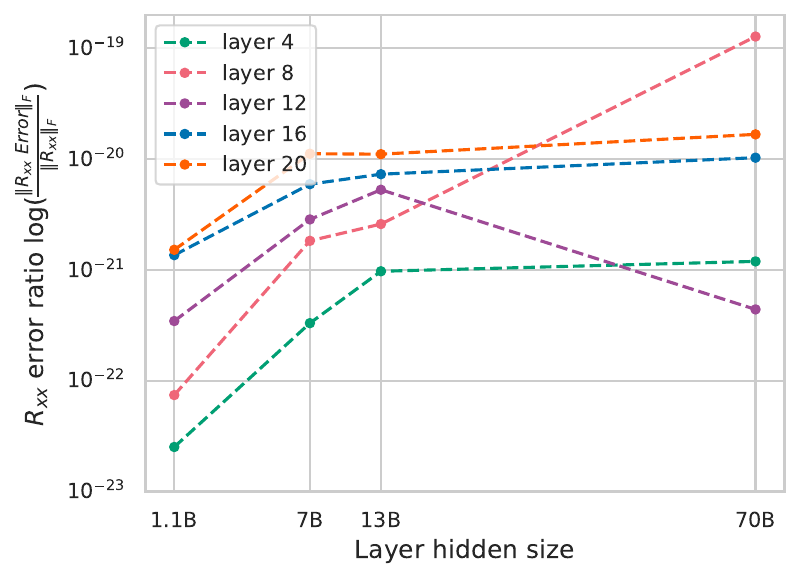}
        \caption{Estimated error ratio of the square root of $\R_{\sX\sX}$}
        \label{fig:scalability-and-stability:sqrtm-error-vs-hidden-size}
    \end{subfigure}
    \hfill
    \begin{subfigure}[b]{0.48\textwidth}
        \includegraphics[width=\textwidth]{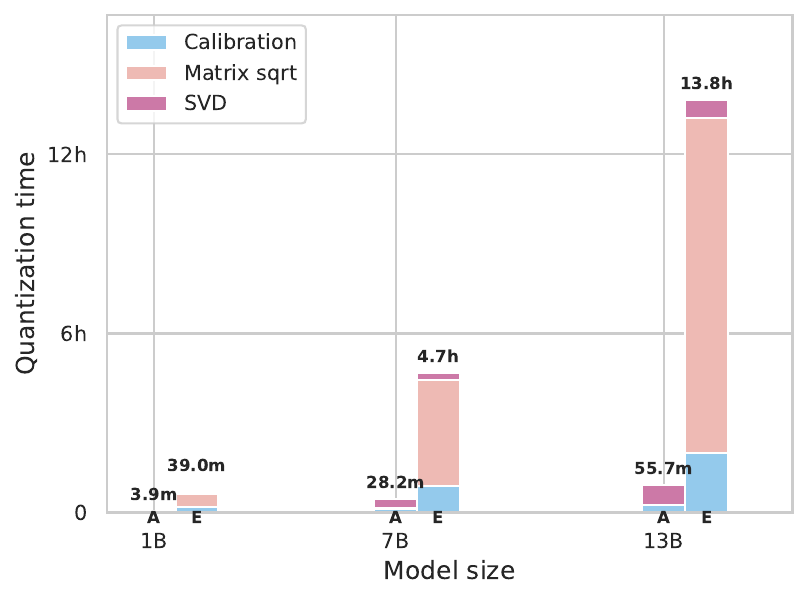}
        \caption{\loqer{} quantization time}
        \label{fig:scalability-and-stability:runtime-vs-model-size}
    \end{subfigure}
    \caption{Scalability of \loqer{}. (a) plots the estimated error ratio of the matrix square root calculation of $\R_{\sX\sX}$ of some layers where the error increases as the model goes larger. (b) compares the quantization time of \loqerdiag{} and \loqerrxx{} if all layers are quantized sequentially. The matrix square root is time-consuming since it is executed on CPUs. One key optimization for accelerating the quantization process of \loqerrxx{} will be the GPU-accelerated matrix square root.}
    \ifdefined\isiclr
        \vspace{-1em}
    \fi
    \label{fig:scalability-and-stability}
\end{figure}

%% file: figures/tab_choice_of_solutions_mrpc.tex
\begin{table}[]
    \caption{Runtime comparison of QERA-exact and QERA-approx on MRPC. It is recommended using QERA-approx for QPEFT instead of QERA-exact.}
    \centering
    \begin{small}
        \begin{tabular}{@{}lcccccc@{}}
            \toprule
            Method      & Rank & Epochs & Init. time & Training time & Total time ($\downarrow$) & Acc. ($\uparrow$) \\ \midrule
            QERA-exact  & 8    & 4      & 1.6min     & 2.2min        & 3.8min                    & 88.97             \\
            QERA-approx & 12   & 4      & 21s        & 2.2min        & \textbf{2.6min}           & \textbf{89.95}    \\
            QERA-approx & 8    & 5      & 21s        & 2.7min        & \textbf{3.1min}           & \textbf{89.97}    \\ \bottomrule
        \end{tabular}
    \end{small}
    \label{tab:appendix:choice-of-solutions-mrpc}
\end{table}

%% file: figures/tab_choice_of_solutions_slimpajama.tex
\begin{table}[]
    \caption{Runtime comparison of QERA-exact and QERA-approx on SlimPajama. It is recommended using QERA-approx for QPEFT instead of QERA-exact.}
    \centering
    \begin{small}
        \begin{tabular}{@{}lcccccc@{}}
            \toprule
            Method      & Rank & Epochs & Init. time & Training time & Total time ($\downarrow$) & PPL. ($\downarrow$) \\ \midrule
            QERA-exact  & 16   & 2      & 4.9h       & 1.9h          & 6.8h                      & 6.31                \\
            QERA-approx & 64   & 2      & 29.6min    & 2.1h          & \textbf{2.6h}             & \textbf{6.18}       \\
            QERA-approx & 16   & 4      & 28.2min    & 4.0h          & \textbf{4.5h}             & \textbf{6.21}       \\ \bottomrule
        \end{tabular}
    \end{small}
    \label{tab:appendix:choice-of-solutions-slimpajama}
\end{table}

%% file: figures/tab_choose_qpeft_rank_sst2.tex
\begin{table}[]
    \caption{Over-parameterization problem. We sweep the rank $k$ of LoRA on SST2 and reported fine-tuned accuracy.
        The highest accuracy at rank $k=12$ indicates over-parameterization happens for $k\ge 12$.}
    \ifdefined\isiclr
        \vspace{-1em}
    \fi
    \centering
    \begin{small}
        \begin{tabular}{@{}lccc@{}}
            \toprule
            Method                & Rank $k$ & Learning rates                & Best Acc.      \\ \midrule
            \multirow{5}{*}{LoRA} & 4        & 1e-4/2e-4/3e-4/4e-4/5e-4/6e-4 & 94.38          \\
                                  & 8        & 1e-4/2e-4/3e-4/4e-4/5e-4/6e-4 & 94.46          \\
                                  & 12       & 1e-4/2e-4/3e-4/4e-4/5e-4/6e-4 & \textbf{94.73} \\
                                  & 16       & 1e-4/2e-4/3e-4/4e-4/5e-4/6e-4 & 94.50          \\
                                  & 20       & 1e-4/2e-4/3e-4/4e-4/5e-4/6e-4 & 94.50          \\ \bottomrule
        \end{tabular}
    \end{small}
    \label{tab:appendix:over-parameterization-sst2}
\end{table}

%% file: figures/tab_choose_qpeft_rank_mrpc.tex
\begin{table}[]
    \caption{Over-parameterization problem. We sweep the rank $k$ of LoRA on MRPC and reported fine-tuned accuracy.
        The highest accuracy at rank $k=12$ indicates over-parameterization happens for $k\ge 12$.}
    \ifdefined\isiclr
        \vspace{-1em}
    \fi
    \centering
    \begin{small}
        \begin{tabular}{@{}lccc@{}}
            \toprule
            Method                & Rank $k$ & Learning rates                & Best Acc.      \\ \midrule
            \multirow{5}{*}{LoRA} & 4        & 1e-4/2e-4/3e-4/4e-4/5e-4/6e-4 & 87.99          \\
                                  & 8        & 1e-4/2e-4/3e-4/4e-4/5e-4/6e-4 & 88.97          \\
                                  & 12       & 1e-4/2e-4/3e-4/4e-4/5e-4/6e-4 & \textbf{89.95} \\
                                  & 16       & 1e-4/2e-4/3e-4/4e-4/5e-4/6e-4 & 89.46          \\
                                  & 20       & 1e-4/2e-4/3e-4/4e-4/5e-4/6e-4 & 89.71          \\ \bottomrule
        \end{tabular}
    \end{small}
    \label{tab:appendix:over-parameterization-mrpc}
\end{table}

%% file: figures/tab_llm_ptq_downstream_detailed.tex
\begin{table}[h]
    \caption{Post-training quantization evaluation of TinyLlama-1.1B.}
    \begin{center}
        \begin{small}
            \resizebox{\linewidth}{!}{
                \begin{tabular}{@{}clcccccccc@{}}
                    \toprule
                    \multirow{2}{*}{rank} & \multirow{2}{*}{Method} & \multirow{2}{*}{w-bits} & ARC (challenge) & BoolQ & CommonSenseQA & BBH       & MMLU  & WikiText2 & Winogrande \\ \cmidrule(l){4-10}
                                          &                         &                         & Acc\_norm       & Acc   & Acc           & Acc\_norm & Acc   & Word ppl  & Acc        \\ \midrule
                    -                     & BF16                    & 16                      & 32.51           & 55.93 & 20.07         & 29.68     & 25.35 & 13.98     & 59.59      \\ \midrule
                    -                     & HQQ                     & \multirow{6}{*}{4.25}   & 32.00           & 58.13 & 20.15         & 29.70     & 25.75 & 15.02     & 59.35      \\
                    -                     & $w$-only                &                         & 28.67           & 58.23 & 19.49         & 28.99     & 23.81 & 19.40     & 52.01      \\
                    \multirow{4}{*}{32}   & \zeroquantii{}          &                         & 29.69           & 57.86 & 19.41         & 29.53     & 24.85 & 18.03     & 52.57      \\
                                          & \LQER{}                 &                         & 32.00           & 52.42 & 18.59         & 29.60     & 25.31 & 16.23     & 59.75      \\
                                          & \loqerdiag{}            &                         & 31.83           & 52.08 & 17.20         & 29.51     & 25.22 & 15.66     & 58.72      \\
                                          & \loqerrxx{}             &                         & 32.00           & 51.31 & 19.33         & 29.42     & 25.19 & 16.16     & 59.67      \\ \bottomrule
                \end{tabular}
            }
        \end{small}
    \end{center}
    \ifdefined\isiclr
        \vspace{-1em}
    \fi
    \label{tab:ptq-downstream-detailed:tinyllama}
\end{table}

\begin{table}[h]
    \caption{Post-training quantization evaluation of Gemma-2-2B.}
    \ifdefined\isiclr
        \vspace{-1em}
    \fi
    \begin{center}
        \begin{small}
            \resizebox{\linewidth}{!}{
                \begin{tabular}{@{}clcccccccc@{}}
                    \toprule
                    \multirow{2}{*}{rank} & \multirow{2}{*}{Method} & \multirow{2}{*}{W-bits} & ARC (challenge) & BoolQ & CommonSenseQA & BBH       & MMLU  & WikiText2 & Winogrande \\ \cmidrule(l){4-10}
                                          &                         &                         & Acc\_norm       & Acc   & Acc           & Acc\_norm & Acc   & Word ppl  & Acc        \\ \midrule
                    -                     & BF16                    & 16                      & 49.91           & 72.60 & 50.29         & 32.67     & 49.44 & 13.08     & 68.82      \\ \midrule
                    -                     & HQQ                     & \multirow{6}{*}{4.25}   & 48.81           & 71.77 & 48.40         & 32.32     & 46.52 & 14.29     & 67.40      \\
                    -                     & w-only                  &                         & 44.62           & 69.91 & 34.07         & 31.96     & 42.90 & 16.23     & 66.54      \\
                    \multirow{4}{*}{32}   & \zeroquantii{}          &                         & 44.45           & 69.94 & 34.07         & 31.50     & 43.27 & 15.71     & 66.22      \\
                                          & \LQER{}                 &                         & 46.08           & 68.84 & 37.59         & 32.60     & 45.78 & 14.55     & 67.72      \\
                                          & \loqerdiag{}            &                         & 45.31           & 68.99 & 36.20         & 32.04     & 45.80 & 14.60     & 67.40      \\
                                          & \loqerrxx{}             &                         & 46.84           & 72.32 & 42.75         & 33.36     & 47.29 & 14.12     & 67.80      \\ \bottomrule
                \end{tabular}
            }
        \end{small}
    \end{center}
    \ifdefined\isiclr
        \vspace{-1em}
    \fi
    \label{tab:ptq-downstream-detailed:gemma2-2b}
\end{table}

\begin{table}[h]
    \caption{Post-training quantization evaluation of Phi3-3.5-mini.}
    \ifdefined\isiclr
        \vspace{-1em}
    \fi
    \begin{center}
        \begin{small}
            \resizebox{\linewidth}{!}{
                \begin{tabular}{@{}clcccccccc@{}}
                    \toprule
                    \multirow{2}{*}{rank} & \multirow{2}{*}{Method} & \multirow{2}{*}{W-bits} & ARC (challenge) & BoolQ & CommonSenseQA & BBH       & MMLU  & WikiText2 & Winogrande \\ \cmidrule(l){4-10}
                                          &                         &                         & Acc\_norm       & Acc   & Acc           & Acc\_norm & Acc   & Word ppl  & Acc        \\ \midrule
                    -                     & BF16                    & 16                      & 59.39           & 84.65 & 71.91         & 48.19     & 64.58 & 11.50     & 72.77      \\ \midrule
                    -                     & HQQ                     & \multirow{6}{*}{4.25}   & 57.00           & 74.34 & 60.20         & 38.22     & 56.00 & 14.63     & 69.61      \\
                    -                     & w-only                  &                         & 59.73           & 82.72 & 68.22         & 44.45     & 61.54 & 14.16     & 70.48      \\
                    \multirow{4}{*}{32}   & \zeroquantii{}          &                         & 59.64           & 82.94 & 68.06         & 44.58     & 62.00 & 14.09     & 69.77      \\
                                          & \LQER{}                 &                         & 59.39           & 84.01 & 70.76         & 45.67     & 62.21 & 12.88     & 70.74      \\
                                          & \loqerdiag{}            &                         & 59.45           & 84.82 & 70.84         & 45.67     & 62.26 & 12.81     & 70.17      \\
                                          & \loqerrxx{}             &                         & 58.70           & 83.73 & 69.45         & 45.37     & 62.01 & 13.00     & 71.19      \\ \bottomrule
                \end{tabular}
            }
        \end{small}
    \end{center}
    \ifdefined\isiclr
        \vspace{-1em}
    \fi
    \label{tab:ptq-downstream-detailed:phi-3.5-mini}
\end{table}

\begin{table}[h]
    \caption{Post-training quantization evaluation of \llama{}-2-7B.}
    \ifdefined\isiclr
        \vspace{-1em}
    \fi
    \begin{center}
        \begin{small}
            \resizebox{\linewidth}{!}{
                \begin{tabular}{@{}clcccccccc@{}}
                    \toprule
                    \multirow{2}{*}{rank} & \multirow{2}{*}{Method} & \multirow{2}{*}{W-bits} & ARC (challenge) & BoolQ & CommonSenseQA & BBH       & MMLU  & WikiText2 & Winogrande \\ \cmidrule(l){4-10}
                                          &                         &                         & Acc\_norm       & Acc   & Acc           & Acc\_norm & Acc   & Word ppl  & Acc        \\ \midrule
                    -                     & BF16                    & 16                      & 46.25           & 77.83 & 33.09         & 30.74     & 40.64 & 8.71      & 69.14      \\ \midrule
                    -                     & HQQ                     & \multirow{6}{*}{4.25}   & 44.03           & 75.87 & 29.40         & 30.50     & 40.14 & 9.59      & 69.61      \\
                    -                     & w-only                  &                         & 45.22           & 75.87 & 25.47         & 30.71     & 40.03 & 9.45      & 68.43      \\
                    \multirow{4}{*}{32}   & \zeroquantii{}          &                         & 45.82           & 75.90 & 24.82         & 29.99     & 39.84 & 9.42      & 68.19      \\
                                          & \LQER{}                 &                         & 44.28           & 76.15 & 29.81         & 30.72     & 40.66 & 9.22      & 69.22      \\
                                          & \loqerdiag{}            &                         & 44.28           & 75.96 & 30.96         & 30.72     & 40.59 & 9.17      & 68.59      \\
                                          & \loqerrxx{}             &                         & 44.80           & 76.39 & 31.61         & 30.57     & 40.86 & 9.12      & 69.22      \\ \bottomrule
                \end{tabular}
            }
        \end{small}
    \end{center}
    \ifdefined\isiclr
        \vspace{-1em}
    \fi
    \label{tab:ptq-downstream-detailed:llama-2-7b}
\end{table}

\begin{table}[h]
    \caption{Post-training quantization evaluation of \llama{}-2-13B.}
    \ifdefined\isiclr
        \vspace{-1em}
    \fi
    \begin{center}
        \begin{small}
            \resizebox{\linewidth}{!}{
                \begin{tabular}{@{}clcccccccc@{}}
                    \toprule
                    \multirow{2}{*}{rank} & \multirow{2}{*}{Method} & \multirow{2}{*}{W-bits} & ARC (challenge) & BoolQ & CommonSenseQA & BBH       & MMLU  & WikiText2 & Winogrande \\ \cmidrule(l){4-10}
                                          &                         &                         & Acc\_norm       & Acc   & Acc           & Acc\_norm & Acc   & Word ppl  & Acc        \\ \midrule
                    -                     & BF16                    & 16                      & 49.49           & 80.58 & 47.34         & 32.65     & 52.18 & 7.68      & 72.22      \\ \midrule
                    -                     & HQQ                     & \multirow{6}{*}{4.25}   & 49.06           & 78.69 & 45.05         & 32.41     & 50.85 & 8.27      & 71.11      \\
                    -                     & w-only                  &                         & 50.43           & 80.58 & 44.06         & 33.45     & 50.21 & 8.06      & 71.98      \\
                    \multirow{4}{*}{32}   & \zeroquantii{}          &                         & 50.00           & 81.04 & 44.47         & 33.50     & 50.31 & 8.07      & 71.59      \\
                                          & \LQER{}                 &                         & 51.02           & 81.25 & 44.47         & 32.41     & 51.24 & 7.96      & 71.98      \\
                                          & \loqerdiag{}            &                         & 51.11           & 80.83 & 44.06         & 32.48     & 51.07 & 7.95      & 71.67      \\
                                          & \loqerrxx{}             &                         & 50.77           & 81.10 & 44.55         & 32.91     & 51.23 & 7.93      & 71.98      \\ \bottomrule
                \end{tabular}
            }
        \end{small}
    \end{center}
    \ifdefined\isiclr
        \vspace{-1em}
    \fi
    \label{tab:ptq-downstream-detailed:llama-2-13b}
\end{table}

\begin{table}[h]
    \caption{Post-training quantization evaluation of \llama{}-3.1-8B.}
    \ifdefined\isiclr
        \vspace{-1em}
    \fi
    \begin{center}
        \begin{small}
            \resizebox{\linewidth}{!}{
                \begin{tabular}{@{}clcccccccc@{}}
                    \toprule
                    \multirow{2}{*}{rank} & \multirow{2}{*}{Method} & \multirow{2}{*}{W-bits} & ARC (challenge) & BoolQ & CommonSenseQA & BBH       & MMLU  & WikiText2 & Winogrande \\ \cmidrule(l){4-10}
                                          &                         &                         & Acc\_norm       & Acc   & Acc           & Acc\_norm & Acc   & Word ppl  & Acc        \\ \midrule
                    -                     & BF16                    & 16                      & 53.50           & 82.05 & 71.42         & 39.07     & 63.27 & 7.55      & 73.95      \\ \midrule
                    -                     & HQQ                     & \multirow{6}{*}{4.25}   & 52.73           & 81.19 & 69.86         & 35.60     & 62.14 & 8.72      & 74.03      \\
                    -                     & w-only                  &                         & 50.68           & 81.31 & 67.24         & 37.34     & 59.03 & 8.78      & 73.56      \\
                    \multirow{4}{*}{32}   & \zeroquantii{}          &                         & 51.11           & 81.25 & 66.99         & 38.43     & 58.94 & 8.83      & 73.48      \\
                                          & \LQER{}                 &                         & 50.34           & 80.98 & 67.49         & 38.05     & 60.23 & 8.45      & 73.40      \\
                                          & \loqerdiag{}            &                         & 50.77           & 81.04 & 66.75         & 37.94     & 60.09 & 8.45      & 73.48      \\
                                          & \loqerrxx{}             &                         & 51.28           & 80.18 & 68.83         & 37.48     & 60.60 & 8.33      & 73.95      \\ \bottomrule
                \end{tabular}
            }
        \end{small}
    \end{center}
    \ifdefined\isiclr
        \vspace{-1em}
    \fi
    \label{tab:ptq-downstream-detailed:llama-31-8b}
\end{table}

\begin{table}[h]
    \caption{Post-training quantization evaluation of \llama{}-3.1-70B.}
    \begin{center}
        \begin{small}
            \resizebox{\linewidth}{!}{
                \begin{tabular}{@{}clcccccccc@{}}
                    \toprule
                    \multirow{2}{*}{rank} & \multirow{2}{*}{Method} & \multirow{2}{*}{W-bits} & ARC (challenge) & BoolQ & CommonSenseQA & BBH       & MMLU  & WikiText2 & Winogrande \\ \cmidrule(l){4-10}
                                          &                         &                         & Acc\_norm       & Acc   & Acc           & Acc\_norm & Acc   & Word ppl  & Acc        \\ \midrule
                    -                     & BF16                    & 16                      & 65.10           & 85.38 & 78.46         & 48.53     & 75.28 & 3.06      & 79.56      \\ \midrule
                    -                     & HQQ                     & \multirow{5}{*}{4.25}   & 63.99           & 85.02 & 77.48         & 48.19     & 75.20 & 3.97      & 77.98      \\
                    -                     & w-only                  &                         & 60.58           & 83.82 & 73.63         & 41.28     & 73.06 & 4.55      & 78.37      \\
                    32                    & \zeroquantii{}          &                         & 59.90           & 83.61 & 73.55         & 42.75     & 73.15 & 4.48      & 77.74      \\
                                          & \LQER{}                 &                         & 62.97           & 83.88 & 76.25         & 48.67     & 74.26 & 4.10      & 79.64      \\
                                          & \loqerdiag{}            &                         & 62.12           & 83.79 & 76.74         & 48.53     & 73.98 & 4.10      & 79.64      \\ \bottomrule
                \end{tabular}
            }
        \end{small}
    \end{center}
    \ifdefined\isiclr
        \vspace{-1em}
    \fi
    \label{tab:ptq-downstream-detailed:llama-31-70b}
\end{table}

%% file: figures/fig_llama-3-8b_rxx_plots.tex
\begin{figure}[h]
    \centering
    \includegraphics[width=0.85\textwidth]{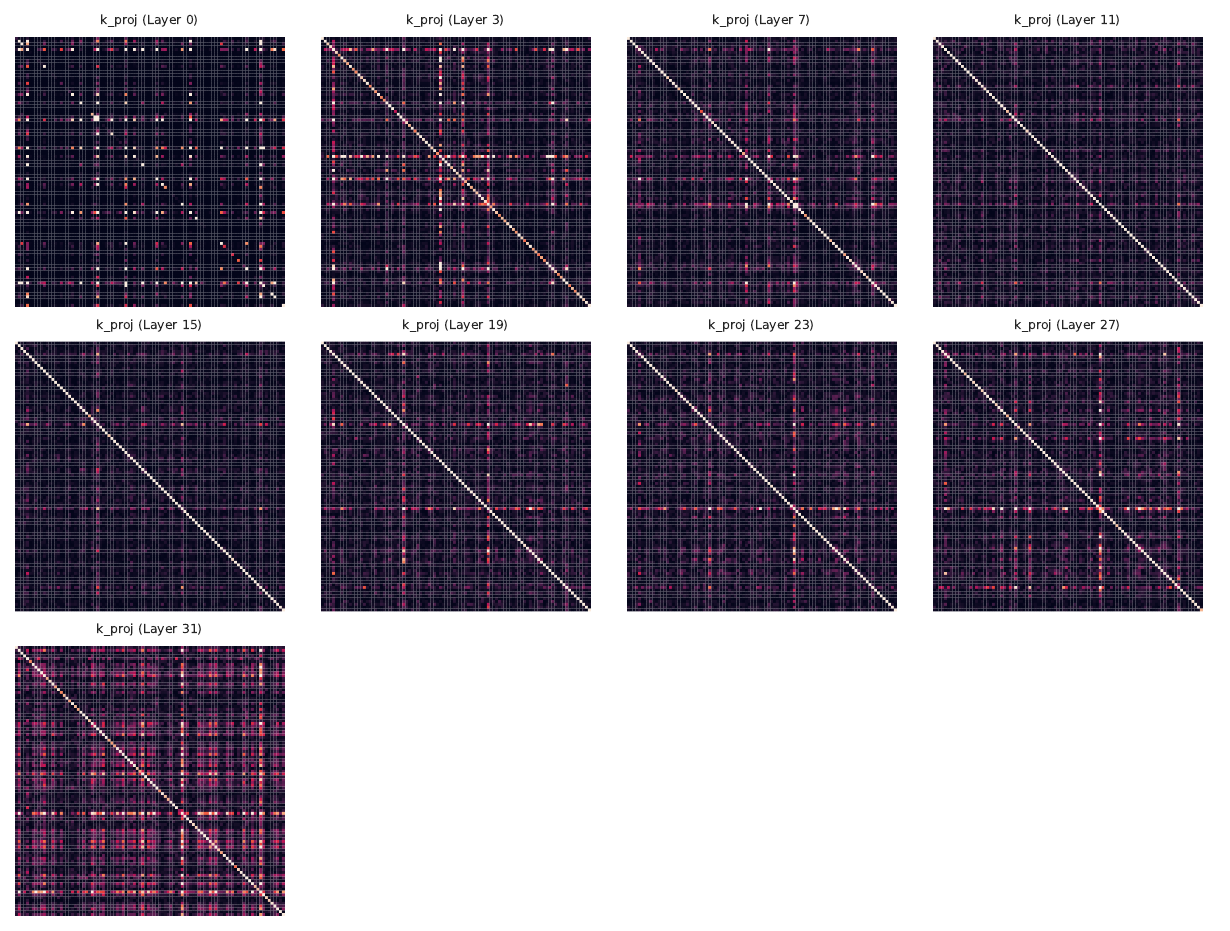}
    \caption{Normalized $\mathrm{abs}({\R_{\sX\sX}})$ of inputs of {\small\texttt{k\_proj}} layers in \llama{}-3-8B. Note that the {\small\texttt{q\_proj}} and {\small\texttt{v\_proj}} share the same inputs. Layers are sampled and only the first 96 dimensions are plotted for clarity.}
    \ifdefined\isiclr
        \vspace{-1em}
    \fi
    \label{fig:llama_3_8b_rxx_k_proj}
\end{figure}
\begin{figure}[h]
    \ifdefined\isiclr
        \vspace{-0.5em}
    \fi
    \centering
    \includegraphics[width=0.85\textwidth]{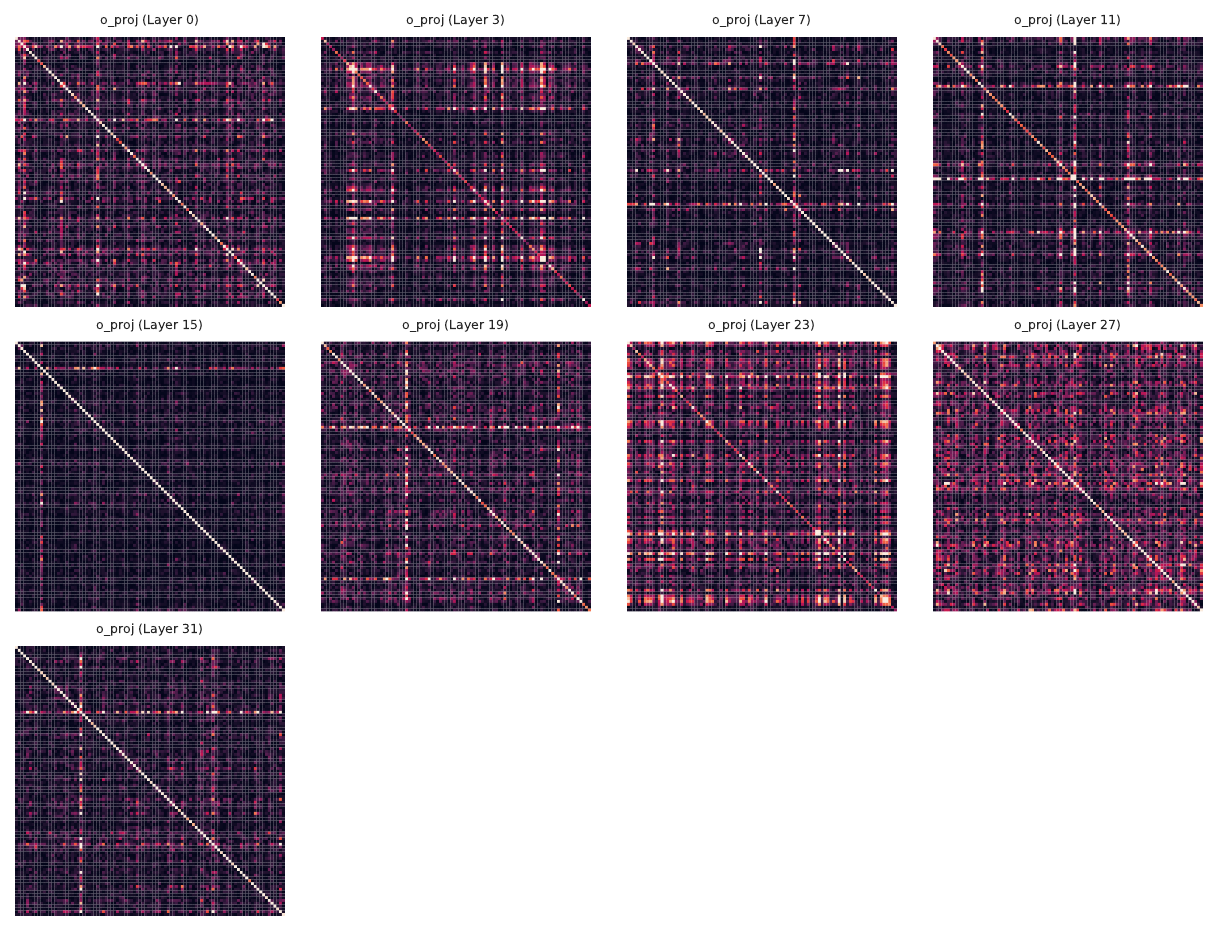}
    \caption{Normalized $\mathrm{abs}({\R_{\sX\sX}})$ of inputs of {\small\texttt{o\_proj}} layers in \llama{}-3-8B. Layers are sampled and only the first 96 dimensions are plotted for clarity.}
    \ifdefined\isiclr
        \vspace{-1em}
    \fi
    \label{fig:llama_3_8b_rxx_o_proj}
\end{figure}
\begin{figure}[h]
    \ifdefined\isiclr
        \vspace{-2em}
    \fi
    \centering
    \includegraphics[width=0.85\textwidth]{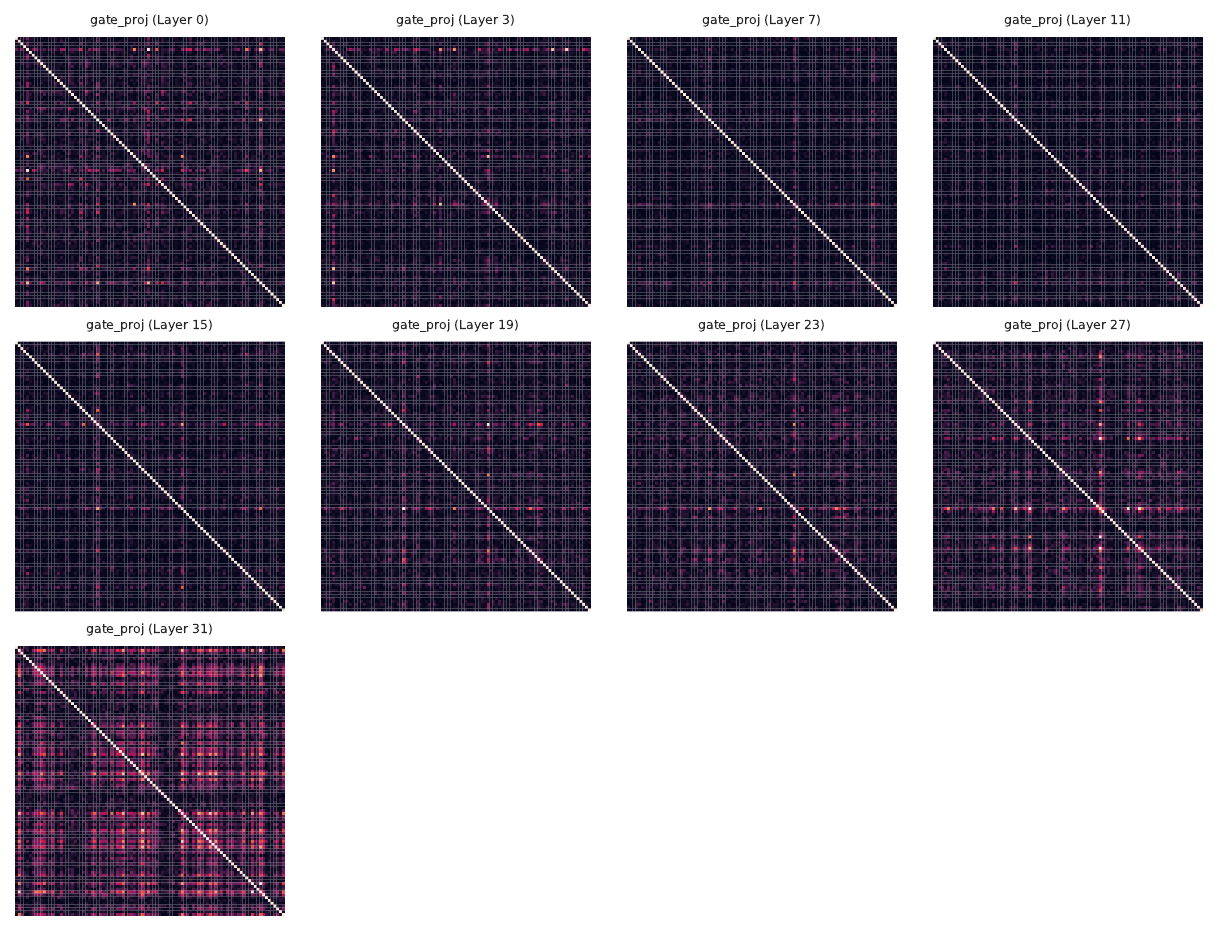}
    \caption{Normalized $\mathrm{abs}({\R_{\sX\sX}})$ of inputs of {\small\texttt{gate\_proj}} layers in \llama{}-3-8B. Note that the {\small\texttt{up\_proj}} shares the same inputs. Layers are sampled and only the first 96 dimensions are plotted for clarity.}
    \ifdefined\isiclr
        \vspace{-1em}
    \fi
    \label{fig:llama_3_8b_rxx_gate_proj}
\end{figure}
\begin{figure}[h]
    \ifdefined\isiclr
        \vspace{-0.5em}
    \fi
    \centering
    \includegraphics[width=0.85\textwidth]{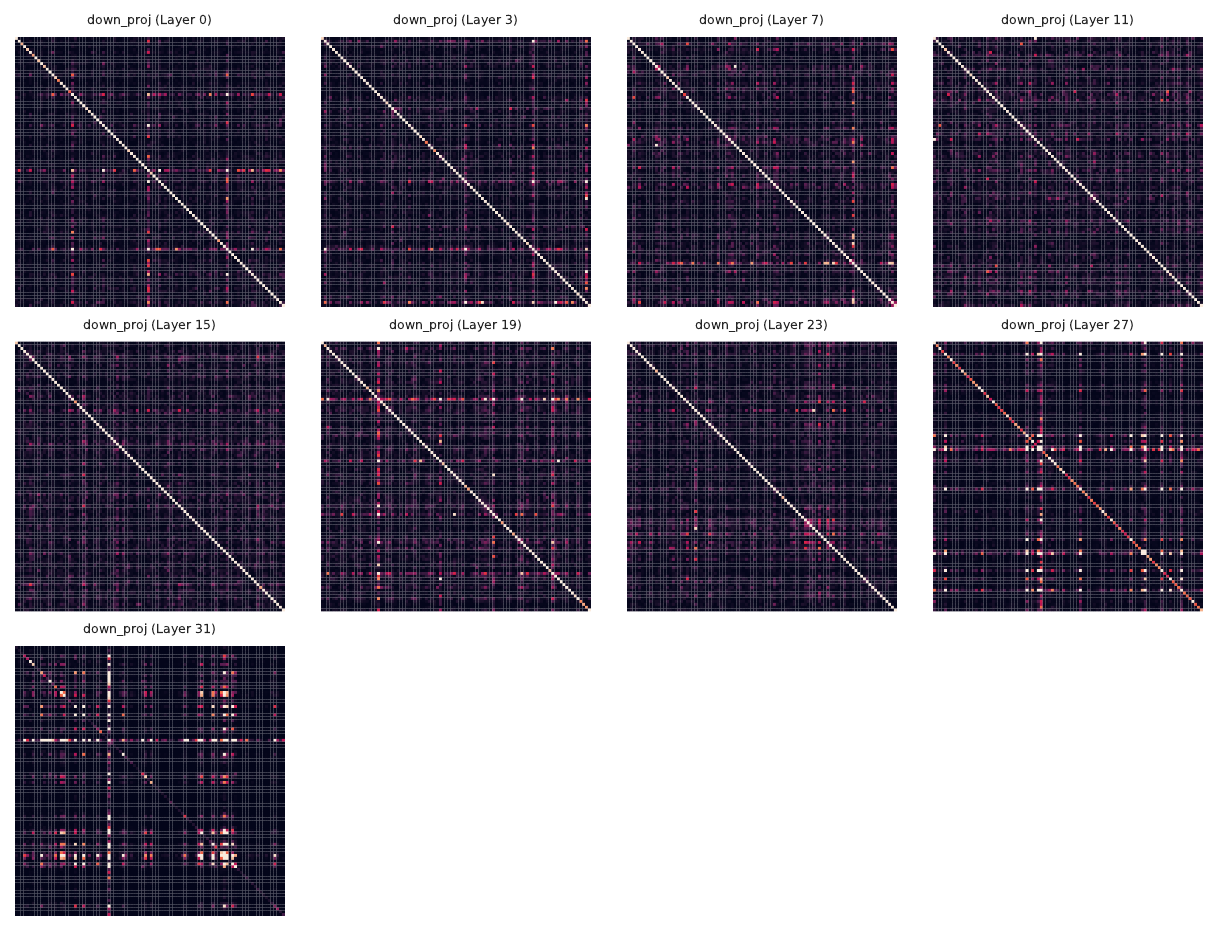}
    \caption{Normalized $\mathrm{abs}({\R_{\sX\sX}})$ of inputs of {\small\texttt{down\_proj}} layers in \llama{}-3-8B. Layers are sampled and only the first 96 dimensions are plotted for clarity.}
    \label{fig:llama_3_8b_rxx_down_proj}
\end{figure}

%% file: figures/fig_llama-2-7b_rxx_plots.tex
\begin{figure}[h]
    \centering
    \includegraphics[width=0.85\textwidth]{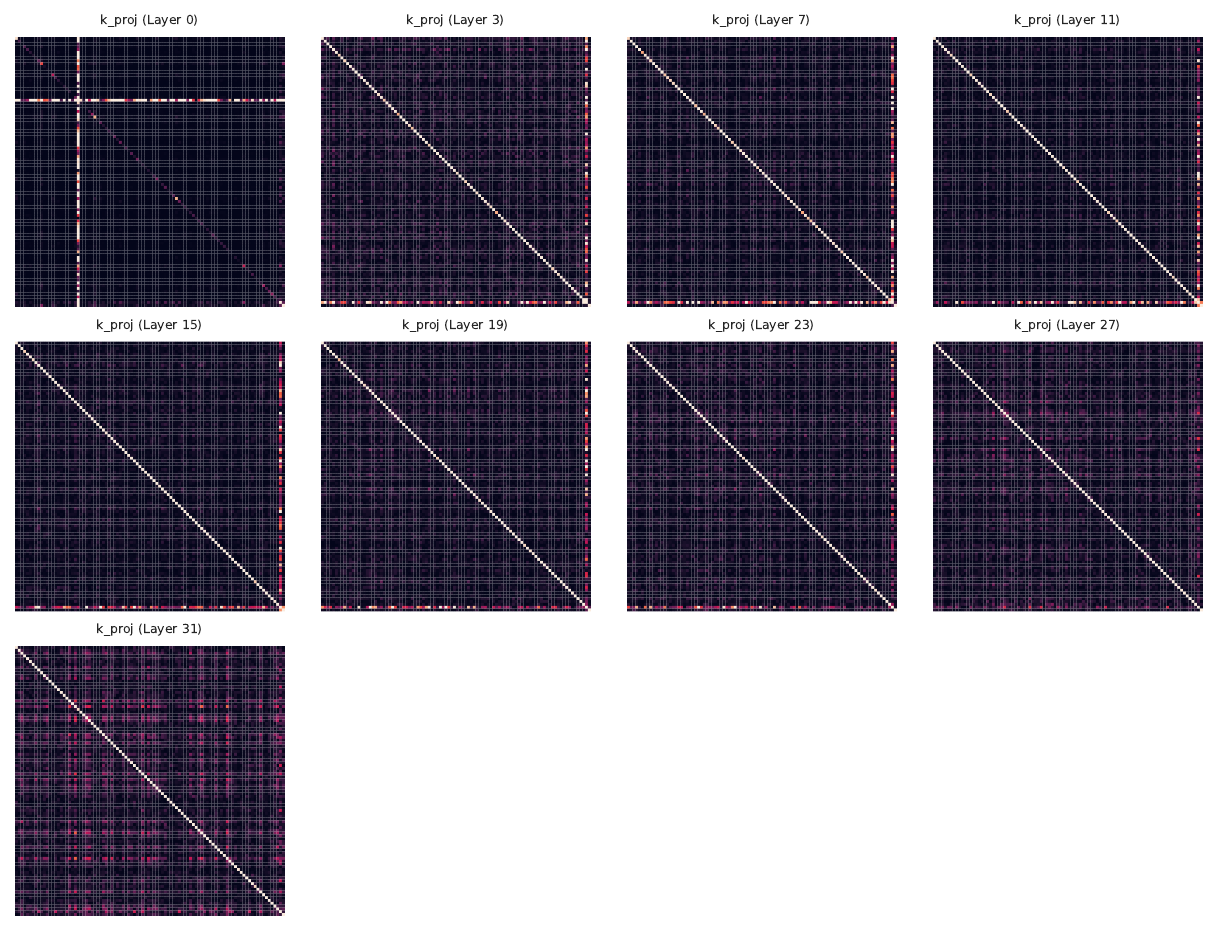}
    \caption{Normalized $\mathrm{abs}({\R_{\sX\sX}})$ of inputs of {\small\texttt{k\_proj}} layers in \llama{}-2-7B. Note that the {\small\texttt{q\_proj}} and {\small\texttt{v\_proj}} share the same inputs. Layers are sampled and only the first 96 dimensions are plotted for clarity.}
    \ifdefined\isiclr
        \vspace{-1em}
    \fi
    \label{fig:llama_2_7b_rxx_k_proj}
\end{figure}
\begin{figure}[h]
    \ifdefined\isiclr
        \vspace{-0.5em}
    \fi
    \centering
    \includegraphics[width=0.85\textwidth]{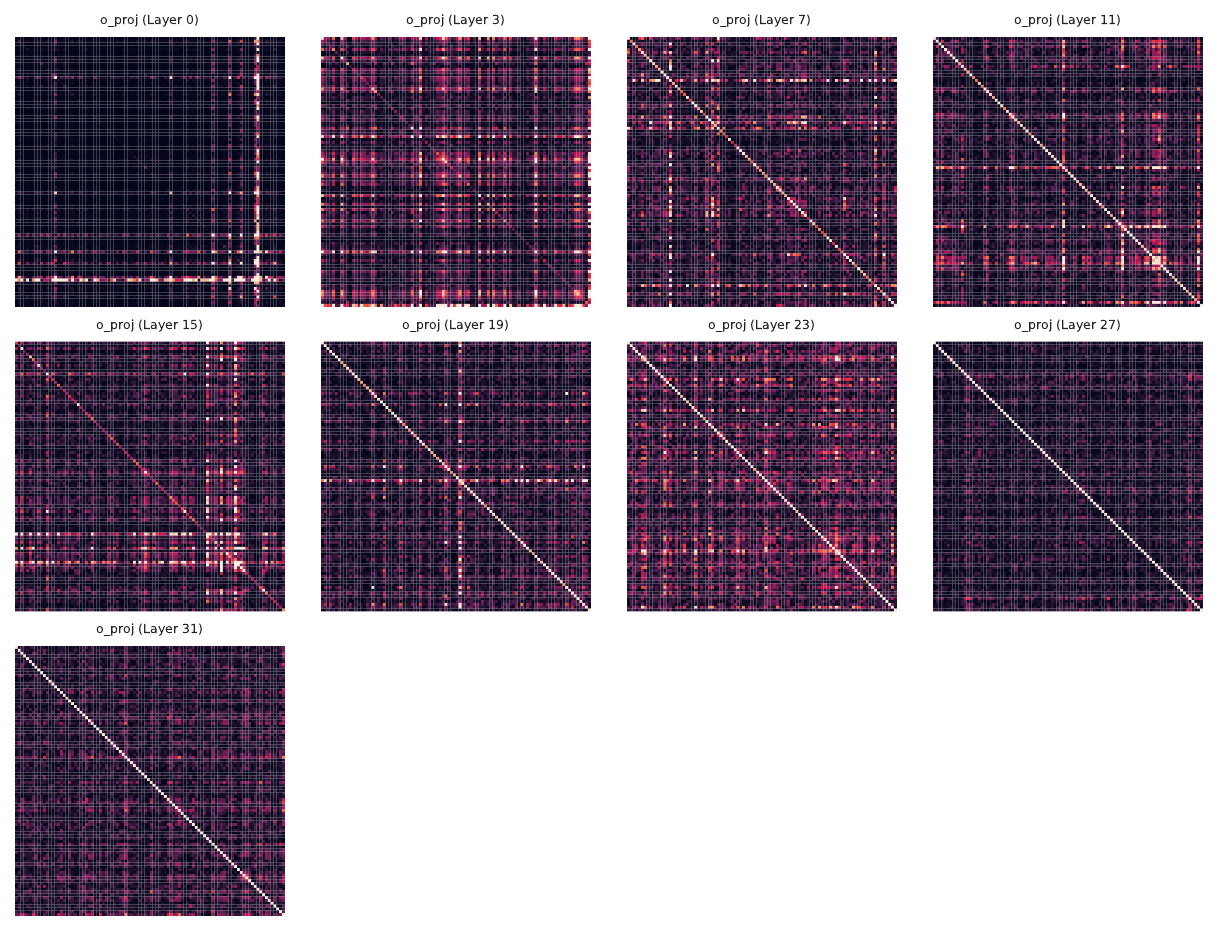}
    \caption{Normalized $\mathrm{abs}({\R_{\sX\sX}})$ of inputs of {\small\texttt{o\_proj}} layers in \llama{}-2-7B. Layers are sampled and only the first 96 dimensions are plotted for clarity.}
    \ifdefined\isiclr
        \vspace{-1em}
    \fi
    \label{fig:llama_2_7b_rxx_o_proj}
\end{figure}
\begin{figure}[h]
    \ifdefined\isiclr
        \vspace{-2em}
    \fi
    \centering
    \includegraphics[width=0.85\textwidth]{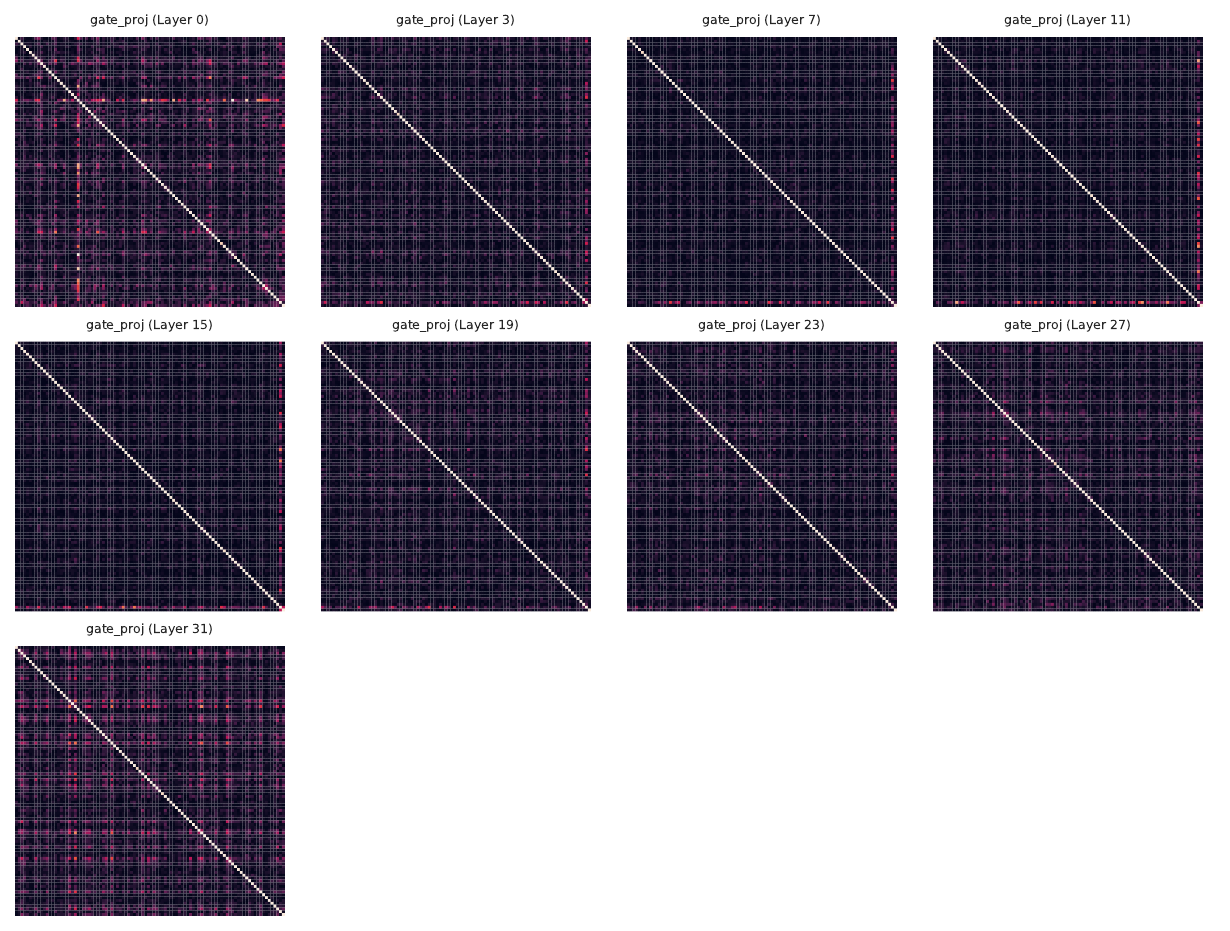}
    \caption{Normalized $\mathrm{abs}({\R_{\sX\sX}})$ of inputs of {\small\texttt{gate\_proj}} layers in \llama{}-2-7B. Note that the {\small\texttt{up\_proj}} shares the same inputs. Layers are sampled and only the first 96 dimensions are plotted for clarity.}
    \ifdefined\isiclr
        \vspace{-1em}
    \fi
    \label{fig:llama_2_7b_rxx_gate_proj}
\end{figure}
\begin{figure}[h]
    \ifdefined\isiclr
        \vspace{-0.5em}
    \fi
    \centering
    \includegraphics[width=0.85\textwidth]{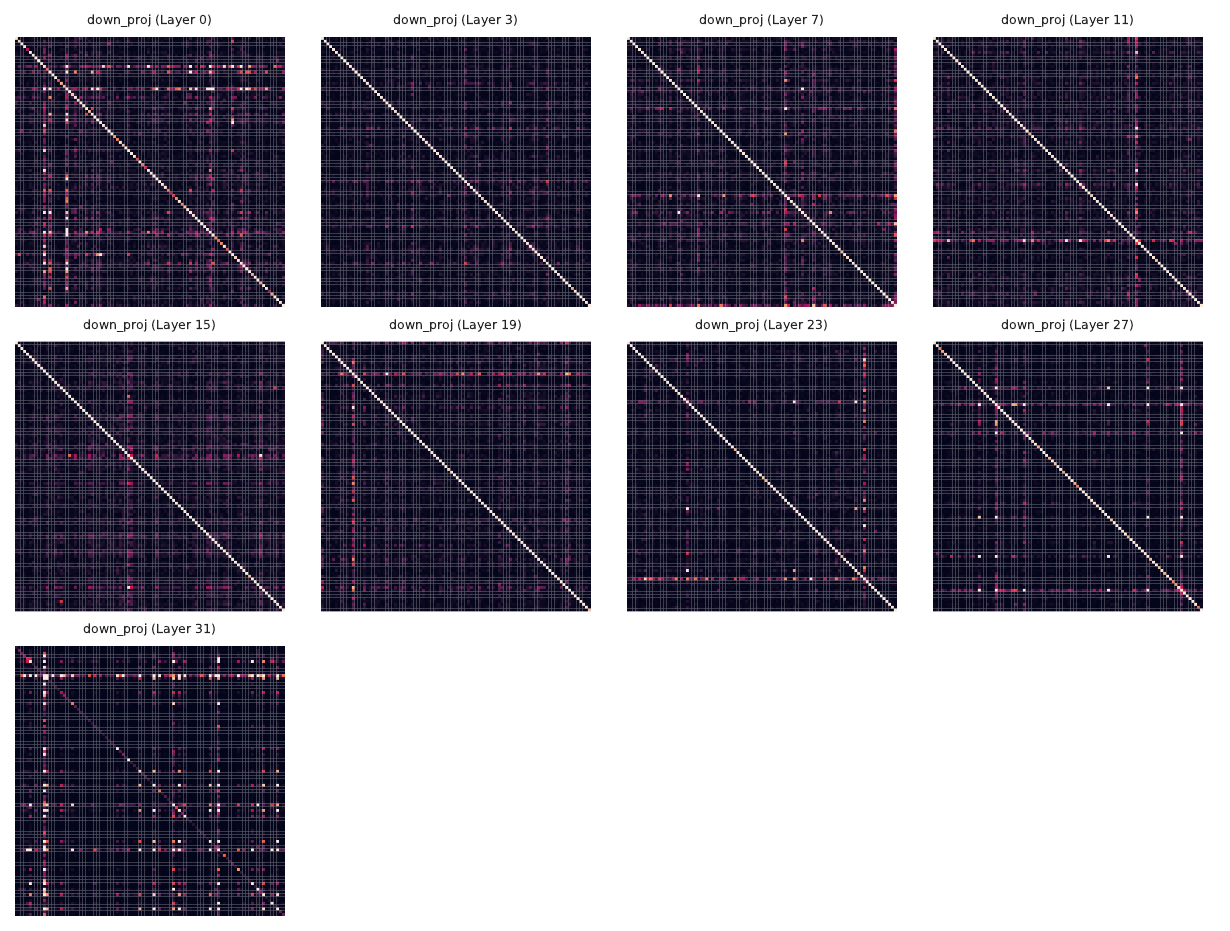}
    \caption{Normalized $\mathrm{abs}({\R_{\sX\sX}})$ of inputs of {\small\texttt{down\_proj}} layers in \llama{}-2-7B. Layers are sampled and only the first 96 dimensions are plotted for clarity.}
    \label{fig:llama_2_7b_rxx_down_proj}
\end{figure}

%% file: figures/fig_mistral-7b_rxx_plots.tex
\begin{figure}[h]
    \centering
    \includegraphics[width=0.85\textwidth]{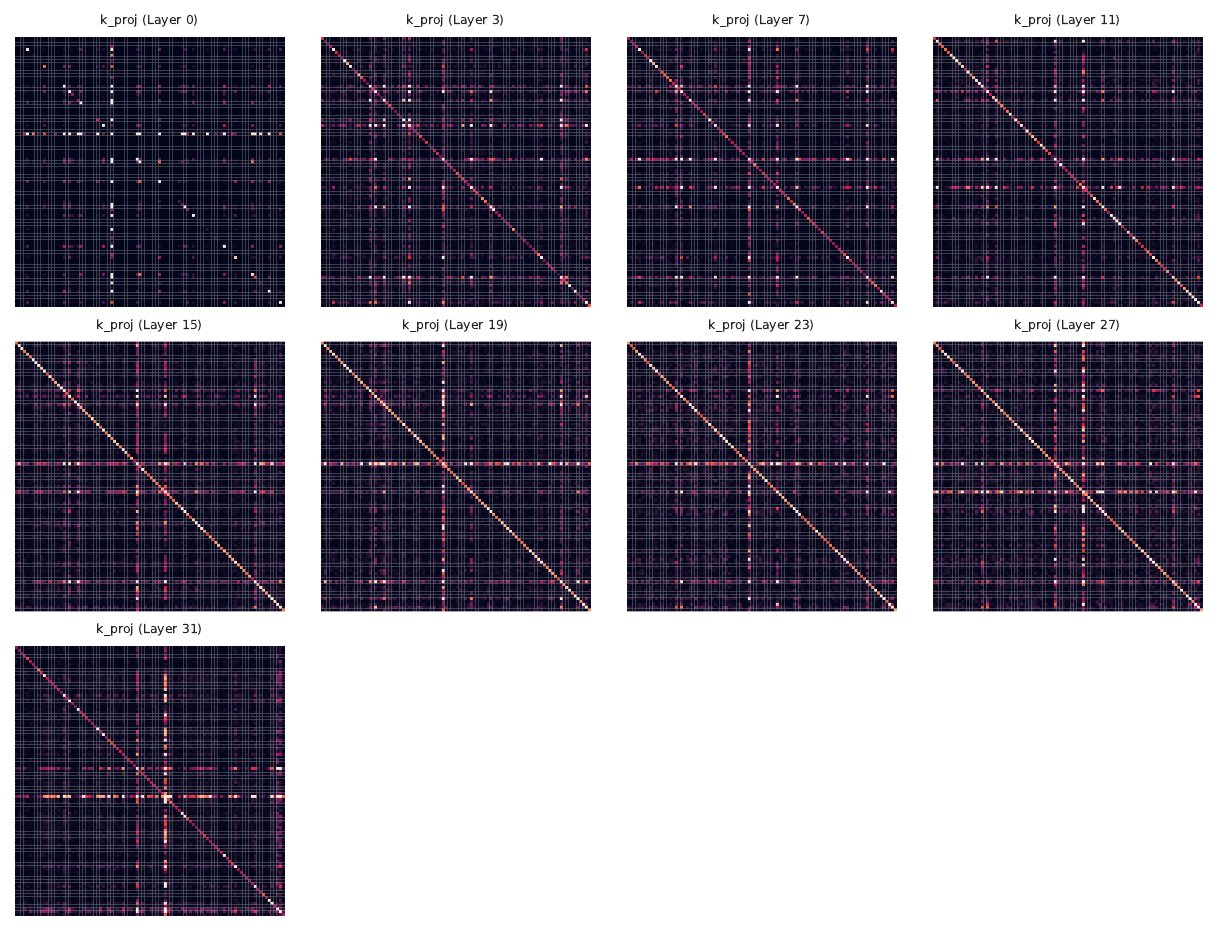}
    \caption{Normalized $\mathrm{abs}({\R_{\sX\sX}})$ of inputs of {\small\texttt{k\_proj}} layers in Mistral-7B-v0.3. Note that the {\small\texttt{q\_proj}} and {\small\texttt{v\_proj}} share the same inputs. Layers are sampled and only the first 96 dimensions are plotted for clarity.}
    \ifdefined\isiclr
        \vspace{-1em}
    \fi
    \label{fig:mistral_7b_rxx_k_proj}
\end{figure}
\begin{figure}[h]
    \ifdefined\isiclr
        \vspace{-0.5em}
    \fi
    \centering
    \includegraphics[width=0.85\textwidth]{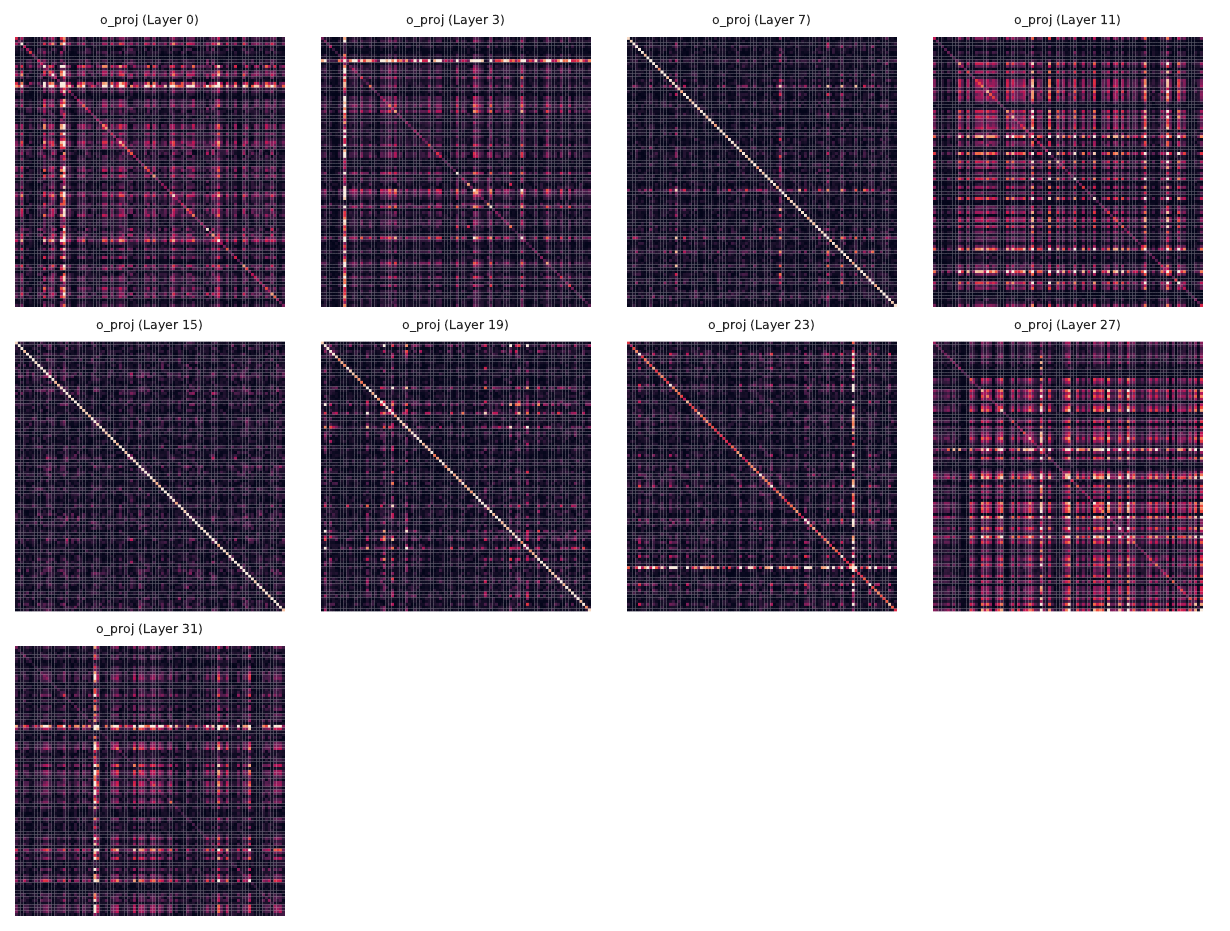}
    \caption{Normalized $\mathrm{abs}({\R_{\sX\sX}})$ of inputs of {\small\texttt{o\_proj}} layers in Mistral-7B-v0.3. Layers are sampled and only the first 96 dimensions are plotted for clarity.}
    \ifdefined\isiclr
        \vspace{-1em}
    \fi
    \label{fig:mistral_7b_rxx_o_proj}
\end{figure}
\begin{figure}[h]
    \ifdefined\isiclr
        \vspace{-2em}
    \fi
    \centering
    \includegraphics[width=0.85\textwidth]{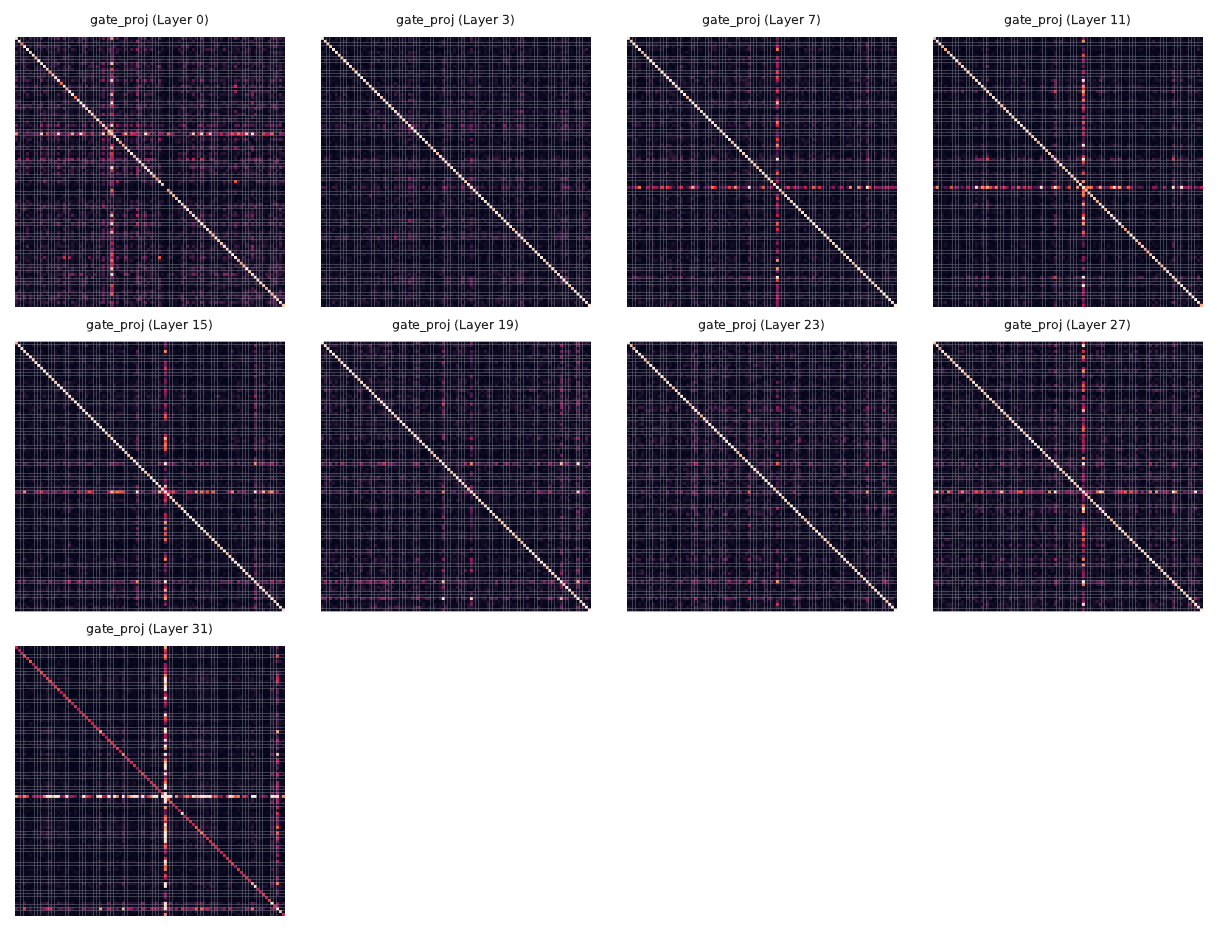}
    \caption{Normalized $\mathrm{abs}({\R_{\sX\sX}})$ of inputs of {\small\texttt{gate\_proj}} layers in Mistral-7B-v0.3. Note that the {\small\texttt{up\_proj}} shares the same inputs. Layers are sampled and only the first 96 dimensions are plotted for clarity.}
    \ifdefined\isiclr
        \vspace{-1em}
    \fi
    \label{fig:mistral_7b_rxx_gate_proj}
\end{figure}
\begin{figure}[h]
    \ifdefined\isiclr
        \vspace{-0.5em}
    \fi
    \centering
    \includegraphics[width=0.85\textwidth]{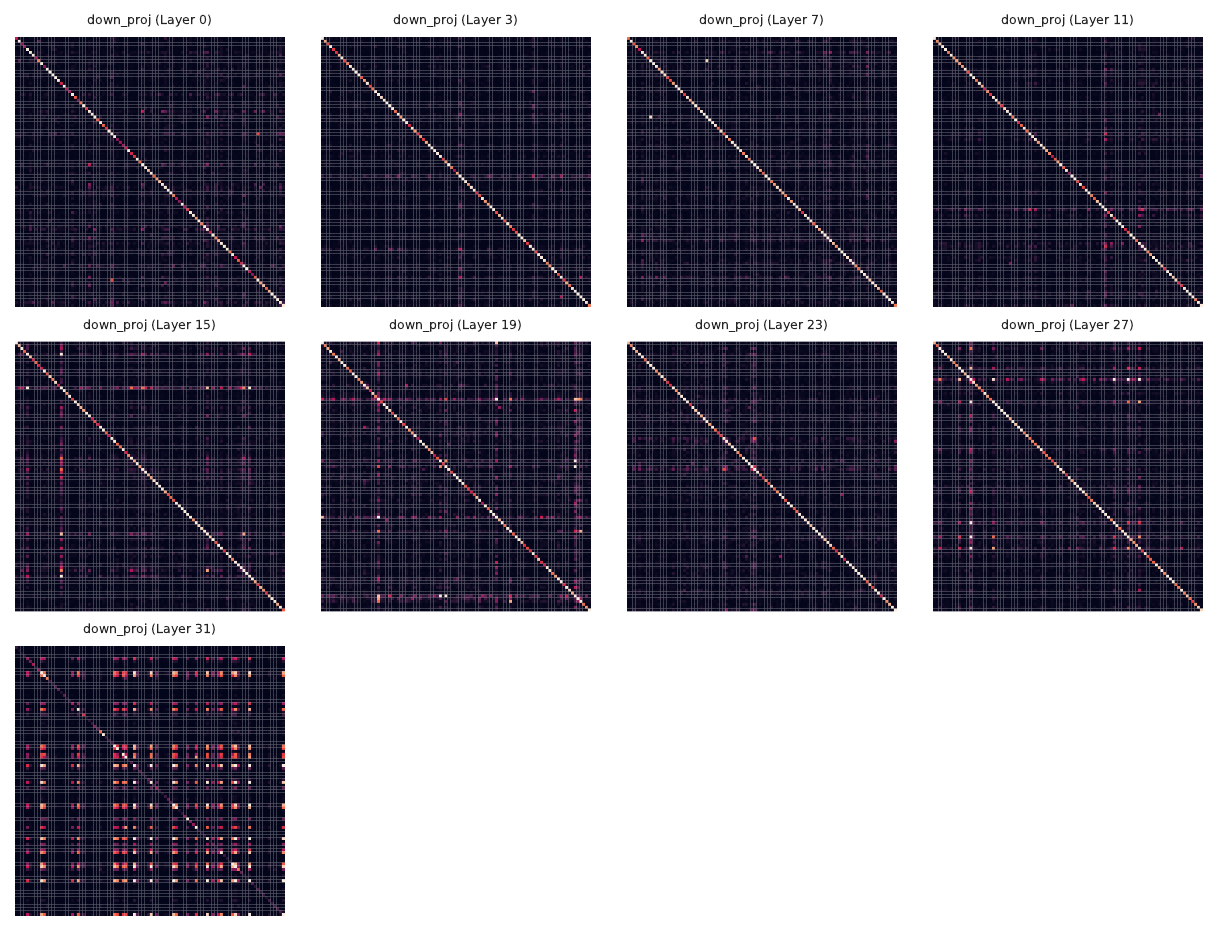}
    \caption{Normalized $\mathrm{abs}({\R_{\sX\sX}})$ of inputs of {\small\texttt{down\_proj}} layers in Mistral-7B-v0.3. Layers are sampled and only the first 96 dimensions are plotted for clarity.}
    \label{fig:mistral_7b_rxx_down_proj}
\end{figure}

%% file: figures/fig_tinyllama-1b_rxx_plots.tex
\begin{figure}[h]
    \centering
    \includegraphics[width=0.85\textwidth]{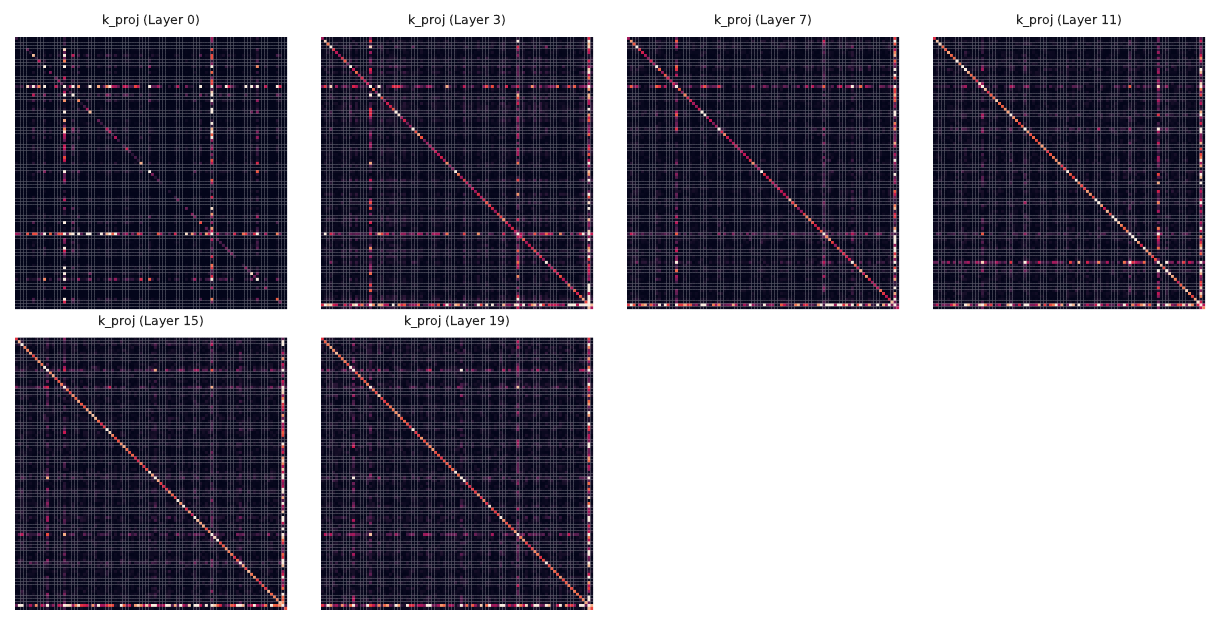}
    \caption{Normalized $\mathrm{abs}({\R_{\sX\sX}})$ of inputs of {\small\texttt{k\_proj}} layers in TinyLlama-1.1B. Note that the {\small\texttt{q\_proj}} and {\small\texttt{v\_proj}} share the same inputs. Layers are sampled and only the first 96 dimensions are plotted for clarity.}
    \ifdefined\isiclr
        \vspace{-1em}
    \fi
    \label{fig:tinyllama_rxx_k_proj}
\end{figure}
\begin{figure}[h]
    \ifdefined\isiclr
        \vspace{-0.5em}
    \fi
    \centering
    \includegraphics[width=0.85\textwidth]{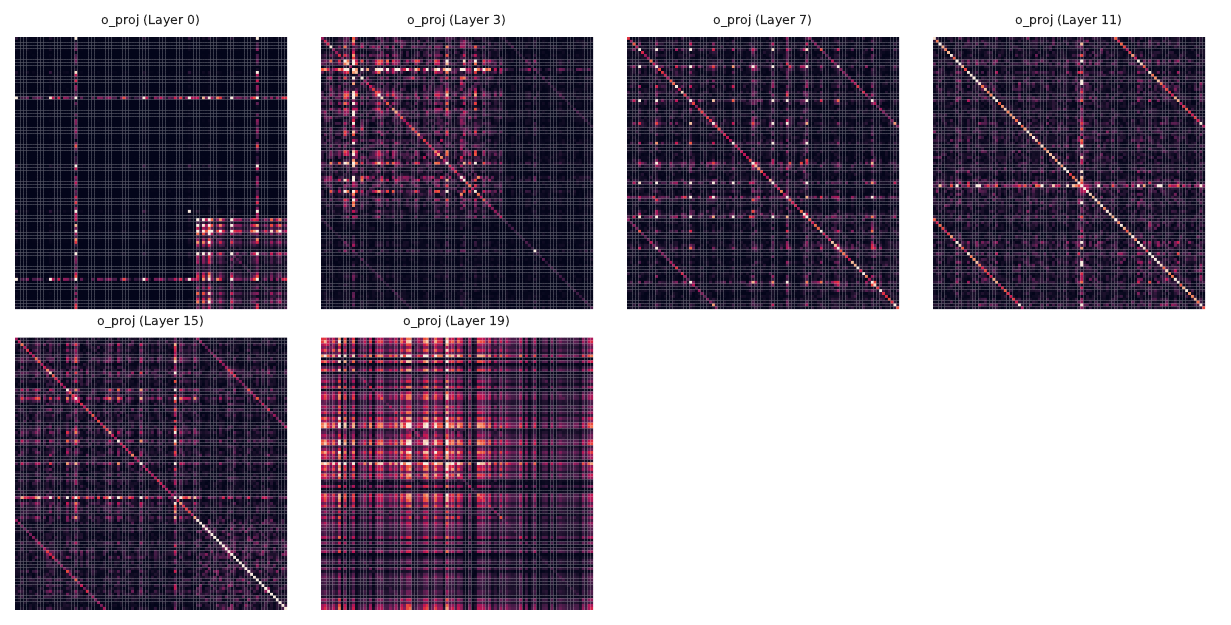}
    \caption{Normalized $\mathrm{abs}({\R_{\sX\sX}})$ of inputs of {\small\texttt{o\_proj}} layers in TinyLlama-1.1B. Layers are sampled and only the first 96 dimensions are plotted for clarity.}
    \ifdefined\isiclr
        \vspace{-1em}
    \fi
    \label{fig:tinyllama_rxx_o_proj}
\end{figure}
\begin{figure}[h]
    \ifdefined\isiclr
        \vspace{-2em}
    \fi
    \centering
    \includegraphics[width=0.85\textwidth]{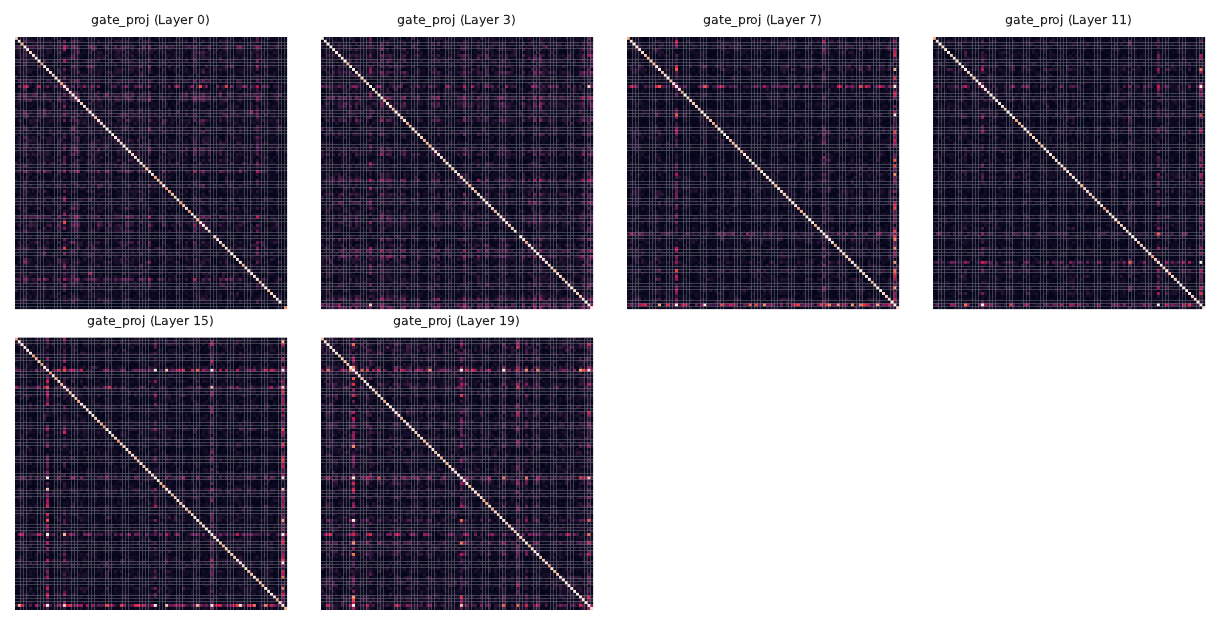}
    \caption{Normalized $\mathrm{abs}({\R_{\sX\sX}})$ of inputs of {\small\texttt{gate\_proj}} layers in TinyLlama-1.1B. Note that the {\small\texttt{up\_proj}} shares the same inputs. Layers are sampled and only the first 96 dimensions are plotted for clarity.}
    \ifdefined\isiclr
        \vspace{-1em}
    \fi
    \label{fig:tinyllama_rxx_gate_proj}
\end{figure}
\begin{figure}[h]
    \ifdefined\isiclr
        \vspace{-0.5em}
    \fi
    \centering
    \includegraphics[width=0.85\textwidth]{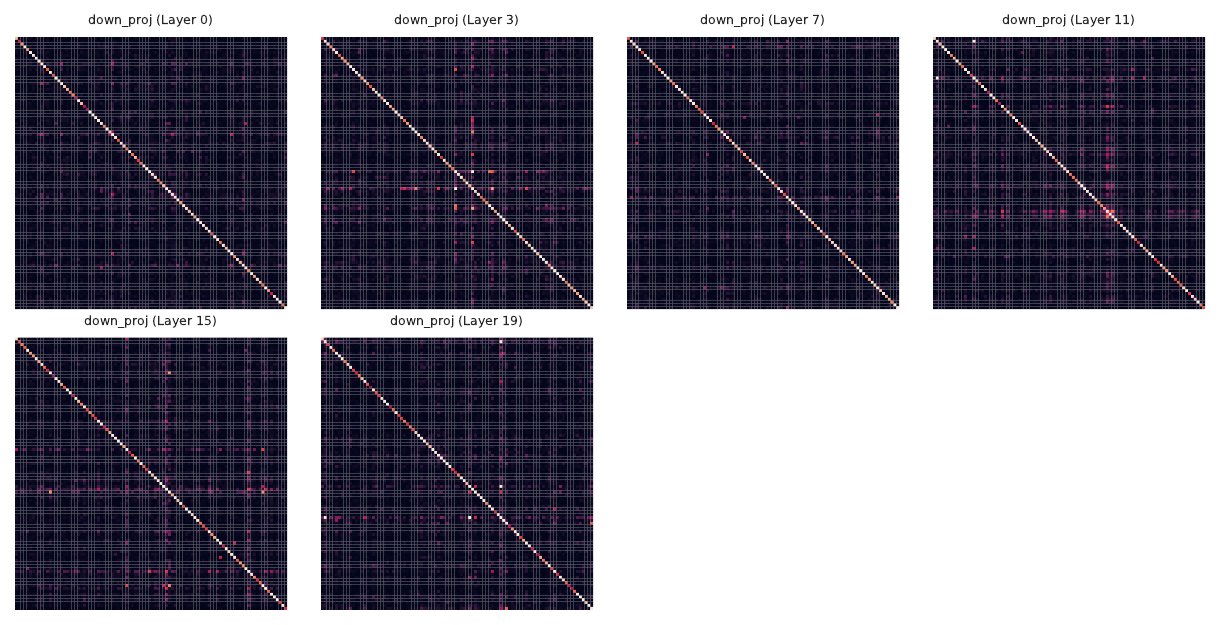}
    \caption{Normalized $\mathrm{abs}({\R_{\sX\sX}})$ of inputs of {\small\texttt{down\_proj}} layers in TinyLlama-1.1B. Layers are sampled and only the first 96 dimensions are plotted for clarity.}
    \label{fig:tinyllama_rxx_down_proj}
\end{figure}

%% file: main.bbl
\begin{thebibliography}{42}
\providecommand{\natexlab}[1]{#1}
\providecommand{\url}[1]{\texttt{#1}}
\expandafter\ifx\csname urlstyle\endcsname\relax
  \providecommand{\doi}[1]{doi: #1}\else
  \providecommand{\doi}{doi: \begingroup \urlstyle{rm}\Url}\fi

\bibitem[Abdin et~al.(2024)Abdin, Jacobs, Awan, Aneja, Awadallah, Awadalla, Bach, Bahree, Bakhtiari, Behl, et~al.]{abdin2024phi3}
Marah Abdin, Sam~Ade Jacobs, Ammar~Ahmad Awan, Jyoti Aneja, Ahmed Awadallah, Hany Awadalla, Nguyen Bach, Amit Bahree, Arash Bakhtiari, Harkirat Behl, et~al.
\newblock Phi-3 technical report: A highly capable language model locally on your phone.
\newblock \emph{arXiv preprint arXiv:2404.14219}, 2024.

\bibitem[Badri \& Shaji(2023)Badri and Shaji]{badri2023hqq}
Hicham Badri and Appu Shaji.
\newblock Half-quadratic quantization of large machine learning models, November 2023.
\newblock URL \url{https://mobiusml.github.io/hqq_blog/}.

\bibitem[Chee et~al.(2024)Chee, Cai, Kuleshov, and De~Sa]{chee2024quip}
Jerry Chee, Yaohui Cai, Volodymyr Kuleshov, and Christopher~M De~Sa.
\newblock Quip: 2-bit quantization of large language models with guarantees.
\newblock \emph{Advances in Neural Information Processing Systems}, 36, 2024.

\bibitem[Clark et~al.(2019)Clark, Lee, Chang, Kwiatkowski, Collins, and Toutanova]{clark2019boolq}
Christopher Clark, Kenton Lee, Ming-Wei Chang, Tom Kwiatkowski, Michael Collins, and Kristina Toutanova.
\newblock Boolq: Exploring the surprising difficulty of natural yes/no questions.
\newblock \emph{arXiv preprint arXiv:1905.10044}, 2019.

\bibitem[Clark et~al.(2018)Clark, Cowhey, Etzioni, Khot, Sabharwal, Schoenick, and Tafjord]{allenai:arc}
Peter Clark, Isaac Cowhey, Oren Etzioni, Tushar Khot, Ashish Sabharwal, Carissa Schoenick, and Oyvind Tafjord.
\newblock Think you have solved question answering? try arc, the ai2 reasoning challenge.
\newblock \emph{arXiv:1803.05457v1}, 2018.

\bibitem[Cobbe et~al.(2021)Cobbe, Kosaraju, Bavarian, Chen, Jun, Kaiser, Plappert, Tworek, Hilton, Nakano, Hesse, and Schulman]{cobbe2021gsm8k}
Karl Cobbe, Vineet Kosaraju, Mohammad Bavarian, Mark Chen, Heewoo Jun, Lukasz Kaiser, Matthias Plappert, Jerry Tworek, Jacob Hilton, Reiichiro Nakano, Christopher Hesse, and John Schulman.
\newblock Training verifiers to solve math word problems.
\newblock \emph{arXiv preprint arXiv:2110.14168}, 2021.

\bibitem[Darvish~Rouhani et~al.(2023)Darvish~Rouhani, Zhao, Elango, Shafipour, Hall, Mesmakhosroshahi, More, Melnick, Golub, Varatkar, et~al.]{darvish2023mxformat}
Bita Darvish~Rouhani, Ritchie Zhao, Venmugil Elango, Rasoul Shafipour, Mathew Hall, Maral Mesmakhosroshahi, Ankit More, Levi Melnick, Maximilian Golub, Girish Varatkar, et~al.
\newblock With shared microexponents, a little shifting goes a long way.
\newblock In \emph{Proceedings of the 50th Annual International Symposium on Computer Architecture}, pp.\  1--13, 2023.

\bibitem[Deadman et~al.(2012)Deadman, Higham, and Ralha]{deadman2012blockedsqrtm}
Edvin Deadman, Nicholas~J Higham, and Rui Ralha.
\newblock Blocked schur algorithms for computing the matrix square root.
\newblock In \emph{International Workshop on Applied Parallel Computing}, pp.\  171--182. Springer, 2012.

\bibitem[Dettmers et~al.(2024)Dettmers, Pagnoni, Holtzman, and Zettlemoyer]{dettmers2024qlora}
Tim Dettmers, Artidoro Pagnoni, Ari Holtzman, and Luke Zettlemoyer.
\newblock Qlora: Efficient finetuning of quantized llms.
\newblock \emph{Advances in Neural Information Processing Systems}, 36, 2024.

\bibitem[Ding et~al.(2023)Ding, Qin, Yang, Wei, Yang, Su, Hu, Chen, Chan, Chen, et~al.]{ding2023parameter}
Ning Ding, Yujia Qin, Guang Yang, Fuchao Wei, Zonghan Yang, Yusheng Su, Shengding Hu, Yulin Chen, Chi-Min Chan, Weize Chen, et~al.
\newblock Parameter-efficient fine-tuning of large-scale pre-trained language models.
\newblock \emph{Nature Machine Intelligence}, 5\penalty0 (3):\penalty0 220--235, 2023.

\bibitem[Dubey et~al.(2024)Dubey, Jauhri, Pandey, Kadian, Al-Dahle, Letman, Mathur, Schelten, Yang, Fan, et~al.]{dubey2024llama}
Abhimanyu Dubey, Abhinav Jauhri, Abhinav Pandey, Abhishek Kadian, Ahmad Al-Dahle, Aiesha Letman, Akhil Mathur, Alan Schelten, Amy Yang, Angela Fan, et~al.
\newblock The llama 3 herd of models.
\newblock \emph{arXiv preprint arXiv:2407.21783}, 2024.

\bibitem[Dubois et~al.(2024)Dubois, Galambosi, Liang, and Hashimoto]{dubois2024alpacaeval}
Yann Dubois, Bal{\'a}zs Galambosi, Percy Liang, and Tatsunori~B Hashimoto.
\newblock Length-controlled alpacaeval: A simple way to debias automatic evaluators.
\newblock \emph{arXiv preprint arXiv:2404.04475}, 2024.

\bibitem[Eckart \& Young(1936)Eckart and Young]{eckart1936approximation}
Carl Eckart and Gale Young.
\newblock The approximation of one matrix by another of lower rank.
\newblock \emph{Psychometrika}, 1\penalty0 (3):\penalty0 211--218, 1936.

\bibitem[Faiz et~al.(2023)Faiz, Kaneda, Wang, Osi, Sharma, Chen, and Jiang]{faiz2023llmcarbon}
Ahmad Faiz, Sotaro Kaneda, Ruhan Wang, Rita Osi, Parteek Sharma, Fan Chen, and Lei Jiang.
\newblock Llmcarbon: Modeling the end-to-end carbon footprint of large language models.
\newblock \emph{arXiv preprint arXiv:2309.14393}, 2023.

\bibitem[Guo et~al.(2023)Guo, Greengard, Xing, and Kim]{guo2023lqlora}
Han Guo, Philip Greengard, Eric~P Xing, and Yoon Kim.
\newblock Lq-lora: Low-rank plus quantized matrix decomposition for efficient language model finetuning.
\newblock \emph{arXiv preprint arXiv:2311.12023}, 2023.

\bibitem[He et~al.(2021)He, Gao, and Chen]{he2021debertav3}
Pengcheng He, Jianfeng Gao, and Weizhu Chen.
\newblock Debertav3: Improving deberta using electra-style pre-training with gradient-disentangled embedding sharing.
\newblock \emph{arXiv preprint arXiv:2111.09543}, 2021.

\bibitem[Hendrycks et~al.(2021)Hendrycks, Burns, Basart, Zou, Mazeika, Song, and Steinhardt]{hendryckstest2021mmlu}
Dan Hendrycks, Collin Burns, Steven Basart, Andy Zou, Mantas Mazeika, Dawn Song, and Jacob Steinhardt.
\newblock Measuring massive multitask language understanding.
\newblock \emph{Proceedings of the International Conference on Learning Representations (ICLR)}, 2021.

\bibitem[Horn \& Johnson(2012)Horn and Johnson]{horn2012matrix}
Roger~A Horn and Charles~R Johnson.
\newblock \emph{Matrix analysis}.
\newblock Cambridge university press, 2012.

\bibitem[Hu et~al.(2021)Hu, Shen, Wallis, Allen-Zhu, Li, Wang, Wang, and Chen]{hu2021lora}
Edward~J Hu, Yelong Shen, Phillip Wallis, Zeyuan Allen-Zhu, Yuanzhi Li, Shean Wang, Lu~Wang, and Weizhu Chen.
\newblock Lora: Low-rank adaptation of large language models.
\newblock \emph{arXiv preprint arXiv:2106.09685}, 2021.

\bibitem[Kaplan et~al.(2020)Kaplan, McCandlish, Henighan, Brown, Chess, Child, Gray, Radford, Wu, and Amodei]{kaplan2020scaling}
Jared Kaplan, Sam McCandlish, Tom Henighan, Tom~B Brown, Benjamin Chess, Rewon Child, Scott Gray, Alec Radford, Jeffrey Wu, and Dario Amodei.
\newblock Scaling laws for neural language models.
\newblock \emph{arXiv preprint arXiv:2001.08361}, 2020.

\bibitem[Li et~al.(2023)Li, Yu, Liang, He, Karampatziakis, Chen, and Zhao]{li2023loftq}
Yixiao Li, Yifan Yu, Chen Liang, Pengcheng He, Nikos Karampatziakis, Weizhu Chen, and Tuo Zhao.
\newblock Loftq: Lora-fine-tuning-aware quantization for large language models.
\newblock \emph{arXiv preprint arXiv:2310.08659}, 2023.

\bibitem[Lin et~al.(2024)Lin, Tang, Tang, Yang, Chen, Wang, Xiao, Dang, Gan, and Han]{lin2024awq}
Ji~Lin, Jiaming Tang, Haotian Tang, Shang Yang, Wei-Ming Chen, Wei-Chen Wang, Guangxuan Xiao, Xingyu Dang, Chuang Gan, and Song Han.
\newblock Awq: Activation-aware weight quantization for on-device llm compression and acceleration.
\newblock \emph{Proceedings of Machine Learning and Systems}, 6:\penalty0 87--100, 2024.

\bibitem[Liu et~al.(2023{\natexlab{a}})Liu, Gong, Wei, Dong, Cai, and Zhuang]{liu2023qllm}
Jing Liu, Ruihao Gong, Xiuying Wei, Zhiwei Dong, Jianfei Cai, and Bohan Zhuang.
\newblock Qllm: Accurate and efficient low-bitwidth quantization for large language models.
\newblock \emph{arXiv preprint arXiv:2310.08041}, 2023{\natexlab{a}}.

\bibitem[Liu(2019)]{liu2019roberta}
Yinhan Liu.
\newblock Roberta: A robustly optimized bert pretraining approach.
\newblock \emph{arXiv preprint arXiv:1907.11692}, 2019.

\bibitem[Liu et~al.(2023{\natexlab{b}})Liu, Oguz, Zhao, Chang, Stock, Mehdad, Shi, Krishnamoorthi, and Chandra]{liu2023llmqat}
Zechun Liu, Barlas Oguz, Changsheng Zhao, Ernie Chang, Pierre Stock, Yashar Mehdad, Yangyang Shi, Raghuraman Krishnamoorthi, and Vikas Chandra.
\newblock Llm-qat: Data-free quantization aware training for large language models.
\newblock \emph{arXiv preprint arXiv:2305.17888}, 2023{\natexlab{b}}.

\bibitem[Meng et~al.(2024)Meng, Wang, and Zhang]{meng2024pissa}
Fanxu Meng, Zhaohui Wang, and Muhan Zhang.
\newblock Pissa: Principal singular values and singular vectors adaptation of large language models.
\newblock \emph{arXiv preprint arXiv:2404.02948}, 2024.

\bibitem[Merity et~al.(2016)Merity, Xiong, Bradbury, and Socher]{merity2016wikitext2}
Stephen Merity, Caiming Xiong, James Bradbury, and Richard Socher.
\newblock Pointer sentinel mixture models, 2016.

\bibitem[Saha et~al.(2024)Saha, Sagan, Srivastava, Goldsmith, and Pilanci]{saha2024caldera}
Rajarshi Saha, Naomi Sagan, Varun Srivastava, Andrea~J Goldsmith, and Mert Pilanci.
\newblock Compressing large language models using low rank and low precision decomposition.
\newblock \emph{arXiv preprint arXiv:2405.18886}, 2024.

\bibitem[Sakaguchi et~al.(2019)Sakaguchi, Le~Bras, Bhagavatula, and Choi]{ai2:winogrande}
Keisuke Sakaguchi, Ronan Le~Bras, Chandra Bhagavatula, and Yejin Choi.
\newblock An adversarial winograd schema challenge at scale.
\newblock \emph{arXiv preprint arXiv:1907.10641}, 2019.

\bibitem[Shao et~al.(2023)Shao, Chen, Zhang, Xu, Zhao, Li, Zhang, Gao, Qiao, and Luo]{shao2023omniquant}
Wenqi Shao, Mengzhao Chen, Zhaoyang Zhang, Peng Xu, Lirui Zhao, Zhiqian Li, Kaipeng Zhang, Peng Gao, Yu~Qiao, and Ping Luo.
\newblock Omniquant: Omnidirectionally calibrated quantization for large language models.
\newblock \emph{arXiv preprint arXiv:2308.13137}, 2023.

\bibitem[Soboleva et~al.(2023)Soboleva, Al-Khateeb, Myers, Steeves, Hestness, and Dey]{cerebras2023slimpajama}
Daria Soboleva, Faisal Al-Khateeb, Robert Myers, Jacob~R Steeves, Joel Hestness, and Nolan Dey.
\newblock {SlimPajama: A 627B token cleaned and deduplicated version of RedPajama}.
\newblock \url{https://www.cerebras.net/blog/slimpajama-a-627b-token-cleaned-and-deduplicated-version-of-redpajama}, 2023.
\newblock URL \url{https://huggingface.co/datasets/cerebras/SlimPajama-627B}.

\bibitem[Styan(1973)]{styan1973hadamard}
George~PH Styan.
\newblock Hadamard products and multivariate statistical analysis.
\newblock \emph{Linear algebra and its applications}, 6:\penalty0 217--240, 1973.

\bibitem[Suzgun et~al.(2022)Suzgun, Scales, Sch{\"a}rli, Gehrmann, Tay, Chung, Chowdhery, Le, Chi, Zhou, , and Wei]{suzgun2022bigbench}
Mirac Suzgun, Nathan Scales, Nathanael Sch{\"a}rli, Sebastian Gehrmann, Yi~Tay, Hyung~Won Chung, Aakanksha Chowdhery, Quoc~V Le, Ed~H Chi, Denny Zhou, , and Jason Wei.
\newblock Challenging big-bench tasks and whether chain-of-thought can solve them.
\newblock \emph{arXiv preprint arXiv:2210.09261}, 2022.

\bibitem[Talmor et~al.(2019)Talmor, Herzig, Lourie, and Berant]{talmor2019commonsenseqa}
Alon Talmor, Jonathan Herzig, Nicholas Lourie, and Jonathan Berant.
\newblock {C}ommonsense{QA}: A question answering challenge targeting commonsense knowledge.
\newblock In \emph{Proceedings of the 2019 Conference of the North {A}merican Chapter of the Association for Computational Linguistics: Human Language Technologies, Volume 1 (Long and Short Papers)}, pp.\  4149--4158, Minneapolis, Minnesota, June 2019. Association for Computational Linguistics.
\newblock \doi{10.18653/v1/N19-1421}.
\newblock URL \url{https://aclanthology.org/N19-1421}.

\bibitem[Team et~al.(2024)Team, Riviere, Pathak, Sessa, Hardin, Bhupatiraju, Hussenot, Mesnard, Shahriari, Ram{\'e}, et~al.]{team2024gemma}
Gemma Team, Morgane Riviere, Shreya Pathak, Pier~Giuseppe Sessa, Cassidy Hardin, Surya Bhupatiraju, L{\'e}onard Hussenot, Thomas Mesnard, Bobak Shahriari, Alexandre Ram{\'e}, et~al.
\newblock Gemma 2: Improving open language models at a practical size.
\newblock \emph{arXiv preprint arXiv:2408.00118}, 2024.

\bibitem[Touvron et~al.(2023)Touvron, Martin, Stone, Albert, Almahairi, Babaei, Bashlykov, Batra, Bhargava, Bhosale, et~al.]{touvron2023llama}
Hugo Touvron, Louis Martin, Kevin Stone, Peter Albert, Amjad Almahairi, Yasmine Babaei, Nikolay Bashlykov, Soumya Batra, Prajjwal Bhargava, Shruti Bhosale, et~al.
\newblock Llama 2: Open foundation and fine-tuned chat models.
\newblock \emph{arXiv preprint arXiv:2307.09288}, 2023.

\bibitem[Yao et~al.(2023)Yao, Wu, Li, Youn, and He]{yao2023zeroquant}
Zhewei Yao, Xiaoxia Wu, Cheng Li, Stephen Youn, and Yuxiong He.
\newblock Zeroquant-v2: Exploring post-training quantization in llms from comprehensive study to low rank compensation.
\newblock \emph{arXiv preprint arXiv:2303.08302}, 2023.

\bibitem[Ye et~al.(2019)Ye, Chen, Wang, and Ling]{ye2019glue}
Zhi-Xiu Ye, Qian Chen, Wen Wang, and Zhen-Hua Ling.
\newblock Align, mask and select: A simple method for incorporating commonsense knowledge into language representation models.
\newblock \emph{arXiv preprint arXiv:1908.06725}, 2019.

\bibitem[Zhang et~al.(2024{\natexlab{a}})Zhang, Cheng, Constantinides, and Zhao]{zhang2024lqer}
Cheng Zhang, Jianyi Cheng, George~A Constantinides, and Yiren Zhao.
\newblock Lqer: Low-rank quantization error reconstruction for llms.
\newblock \emph{arXiv preprint arXiv:2402.02446}, 2024{\natexlab{a}}.

\bibitem[Zhang et~al.(2024{\natexlab{b}})Zhang, Zeng, Wang, and Lu]{zhang2024tinyllama}
Peiyuan Zhang, Guangtao Zeng, Tianduo Wang, and Wei Lu.
\newblock Tinyllama: An open-source small language model, 2024{\natexlab{b}}.

\bibitem[Zhang et~al.(2023)Zhang, Chen, Bukharin, Karampatziakis, He, Cheng, Chen, and Zhao]{zhang2023adalora}
Qingru Zhang, Minshuo Chen, Alexander Bukharin, Nikos Karampatziakis, Pengcheng He, Yu~Cheng, Weizhu Chen, and Tuo Zhao.
\newblock Adalora: Adaptive budget allocation for parameter-efficient fine-tuning.
\newblock \emph{arXiv preprint arXiv:2303.10512}, 2023.

\bibitem[Zheng et~al.(2023)Zheng, Chiang, Sheng, Zhuang, Wu, Zhuang, Lin, Li, Li, Xing, et~al.]{zheng2023vicuna}
Lianmin Zheng, Wei-Lin Chiang, Ying Sheng, Siyuan Zhuang, Zhanghao Wu, Yonghao Zhuang, Zi~Lin, Zhuohan Li, Dacheng Li, Eric Xing, et~al.
\newblock Judging llm-as-a-judge with mt-bench and chatbot arena.
\newblock \emph{Advances in Neural Information Processing Systems}, 36:\penalty0 46595--46623, 2023.

\end{thebibliography}
